%% file: _main.tex
\input{_constants}
\arxiv 

\pdfoutput=1
\documentclass[10pt,twocolumn,letterpaper]{article}
\input{cvpr_header}

\unless\ifarxiv \myexternaldocument{_supplementary} \fi

\newcommand\copyrighttext{%
  \footnotesize \textcopyright 2025 IEEE. Personal use of this material is permitted.
  Permission from IEEE must be obtained for all other uses, in any current or future
  media, including reprinting/republishing this material for advertising or promotional
  purposes, creating new collective works, for resale or redistribution to servers or
  lists, or reuse of any copyrighted component of this work in other works.}
\newcommand\copyrightnotice{%
\begin{tikzpicture}[remember picture,overlay]
\node[anchor=south,yshift=10pt] at (current page.south) {\fbox{\parbox{\dimexpr\textwidth-\fboxsep-\fboxrule\relax}{\copyrighttext}}};
\end{tikzpicture}%
}

\begin{document}
\title{\paperTitle}
\author{\authorBlock}
\maketitle

\copyrightnotice

\input{00_abstract}
\input{01_intro}
\input{02_related}
\input{03_method}
\input{04_experimental_setup}
\input{05_results}
\input{10_conclusion}

{\small
\bibliographystyle{ieeenat_fullname}
\bibliography{11_references}
}

\ifarxiv \clearpage \appendix \input{12_appendix} \fi

\end{document}

%% file: _constants.tex
\def\paperID{3}
\def\confName{ICCV}
\def\confYear{2025}

\def\paperTitle{Featuremetric Optimization of Geometric Primitives}
\def\paperTitle{PixCuboid: Room Layout Estimation from Multiple Views}
\def\paperTitle{PixCuboid: Room Layout Estimation from Multiple Views via Dense Featuremetric Alignment}

\def\paperTitle{PixCuboid: Multi-view Room Layout Estimation \\from Dense Featuremetric Alignment}

\def\paperTitle{PixCuboid:  Room Layout Estimation from Multi-view Featuremetric Alignment}

\def\methodName{PixCuboid}

\def\authorBlock{
    Gustav Hanning \qquad
    Kalle Åström \qquad
    Viktor Larsson \\
    Lund University \\
}

\newif\ifreview 
\newif\ifarxiv \newcommand{\arxiv}{\arxivtrue}
\newif\ifcamera 
\newif\ifrebuttal 

%% file: cvpr_header.tex
\ifreview \usepackage[review]{cvpr} \fi
\ifarxiv \usepackage[pagenumbers]{cvpr} \fi
\ifrebuttal \usepackage[rebuttal]{cvpr} \fi
\ifcamera \usepackage{cvpr} \fi

\input{_macros}  

\usepackage{xr-hyper}

\makeatletter
\newcommand*{\addFileDependency}[1]{
  \typeout{(#1)}
  \@addtofilelist{#1}
  \IfFileExists{#1}{}{\typeout{No file #1.}}
}

\makeatother
\newcommand*{\myexternaldocument}[1]{
    \externaldocument{#1}
    \addFileDependency{#1.tex}
    \addFileDependency{#1.aux}
}

\definecolor{cvprblue}{rgb}{0.21,0.49,0.74}
\usepackage[pagebackref,breaklinks,colorlinks,allcolors=cvprblue]{hyperref}
\usepackage[capitalize]{cleveref}
\crefname{section}{Sec.}{Secs.}
\crefname{table}{Table}{Tables}
\crefname{figure}{Fig.}{Figs.}
\usepackage{gensymb}
\usepackage{pgfplots}
\pgfplotsset{compat=1.18}
\usepackage{tikz}
\usepackage{bm}
\usepackage{threeparttable}
\usepackage{overpic}

\ifarxiv \crefname{appendix}{App.}{Apps.}
\else \crefname{appendix}{Suppl.}{Suppls.} \fi

\iftrue
\fi

%% file: _macros.tex
\usepackage[accsupp]{axessibility}  
\usepackage{graphicx}	
\usepackage{amsmath}	
\usepackage{amssymb}	
\usepackage{booktabs}
\usepackage{times}
\usepackage{microtype}
\usepackage{epsfig}
\usepackage{caption}
\usepackage{float}
\usepackage{placeins}
\usepackage{color, colortbl}
\usepackage{stfloats}
\usepackage{enumitem}
\usepackage{tabularx}
\usepackage{xstring}
\usepackage{multirow}
\usepackage{xspace}
\usepackage{url}
\usepackage{subcaption}
\usepackage{xcolor}
\usepackage[hang,flushmargin]{footmisc}

\ifcamera \usepackage[accsupp]{axessibility} \fi

% ...

\ifarxiv  \fi

\newcommand{\R}[1]{{%
    \textbf{%
        \ifstrequal{#1}{1}{\textcolor{red}{R#1}}{%
        \ifstrequal{#1}{2}{\textcolor{blue}{R#1}}{%
        \ifstrequal{#1}{3}{\textcolor{magenta}{R#1}}{%
        \ifstrequal{#1}{4}{\textcolor{teal}{R#1}}{%
                           \textcolor{cyan}{R#1}%
        }}}}%
    }%
}}

\newcommand{\xx}{\boldsymbol{x}}
\newcommand{\II}{\mathrm{\bf I}}
\newcommand{\FF}{\mathrm{\bf F}}
\newcommand{\CF}{\mathrm{\bf C_F}}
\newcommand{\EE}{\mathrm{\bf E}}
\newcommand{\CE}{\mathrm{\bf C_E}}

\definecolor{PredColor}{RGB}{14,154,167}
\definecolor{GTColor}{RGB}{246,205,97}
\definecolor{InitColor}{RGB}{254,138,113}
\definecolor{InitVPColor}{RGB}{74,78,77}

%% file: 00_abstract.tex
\begin{abstract}
Coarse room layout estimation provides important geometric cues for many downstream tasks.
Current state-of-the-art methods are predominantly based on single views and often assume panoramic images. 
We introduce \methodName{}, an optimization-based approach for cuboid-shaped room layout estimation, which is based on multi-view alignment of dense 
deep features.
By training  with the optimization end-to-end, we learn feature maps that yield large convergence basins and smooth loss landscapes in the alignment.
This allows us to initialize the room layout using simple heuristics. 
For the evaluation we propose two new benchmarks based on ScanNet++ and 2D-3D-Semantics, with manually verified ground truth 3D cuboids.
In thorough experiments we validate our approach and significantly outperform the competition.
Finally, while our network is trained with single cuboids, the flexibility of the optimization-based approach allow us to easily extend to multi-room estimation, e.g.~larger apartments or offices.
%
%
%
Code and model weights are available at \url{https://github.com/ghanning/PixCuboid}.

\end{abstract}

%% file: 01_intro.tex
\vspace{-5mm}

\section{Introduction}
\label{sec:intro}

For indoor scenes, knowing the room layout (i.e.~position of the walls, ceiling, and floor) can provide important geometric cues for downstream applications.
For example, 
for anchoring virtual content to walls in AR applications.
Room layouts can also serve as a map for localization problems with low-fidelity sensors (e.g.~radar, ultrasonics or 1D lidars).
In such cases, the layout provides an abstract representation of the scene, excluding clutter and retaining only key structural elements.
%
Recent methods have focused on the monocular case, predicting the full room layout
from a single image or panorama.
In this paper, we argue that only considering single images makes the task unnecessarily hard, since in many application contexts multi-view imagery is readily available.
Instead, we consider solving the room layout estimation problem from a \textbf{collection of posed images}, e.g.~obtained via SLAM or Structure-from-Motion.
Knowing the camera geometry greatly simplifies the problem as it resolves the global scale and the parallax between views allows for proper geometric reasoning about the extent of the room.
\input{figs/teaser-v2}
Most methods currently estimate the room layout in a feed-forward manner via regression directly from the images.
 In contrast, we propose an optimization-based approach which can naturally integrate image and pose information from an arbitrary number of images.
Inspired by PixLoc~\cite{sarlin2021back}, the optimization is based on direct alignment of learned dense feature maps.
 This is performed in a coarse-to-fine manner, allowing our method to obtain high accuracy results even from poor initial estimates.
The feature extraction network is trained end-to-end with unrolled optimization to ensure a smooth loss landscape for the alignment, leading to large convergence basins.
 As our method operates directly on the RGB images, it does not require performing costly dense 3D reconstruction, instead fitting directly to the pixels.
 In the paper, we focus on room shapes consisting of a single cuboid.
However, neither the learned features or optimization framework is specific for cuboids, but can in principle be applied to any parametric room representation consisting of flat surfaces that allow warping between images, e.g.~3D polygons or compositions of multiple cuboids.
 
In the paper, we make the following contributions:
 \begin{itemize}
\item We propose a featuremetric 
approach for  room layout estimation 
 based on multi-view alignment of deep features.
\item We propose a simple cuboid initialization heuristic which only relies on the orientation and position of the cameras. 
\item 
We provide 
new benchmarks based on ScanNet++ v2 \cite{yeshwanth2023scannet++} and 2D-3D-Semantics \cite{armeni2017joint} where we provide manually verified ground truth 3D cuboids and code for evaluation.
\end{itemize}

\input{figs/overview}


%% file: figs/teaser-v2.tex
\begin{figure}[t]
    \centering
    \begin{overpic}[width=0.47\textwidth]{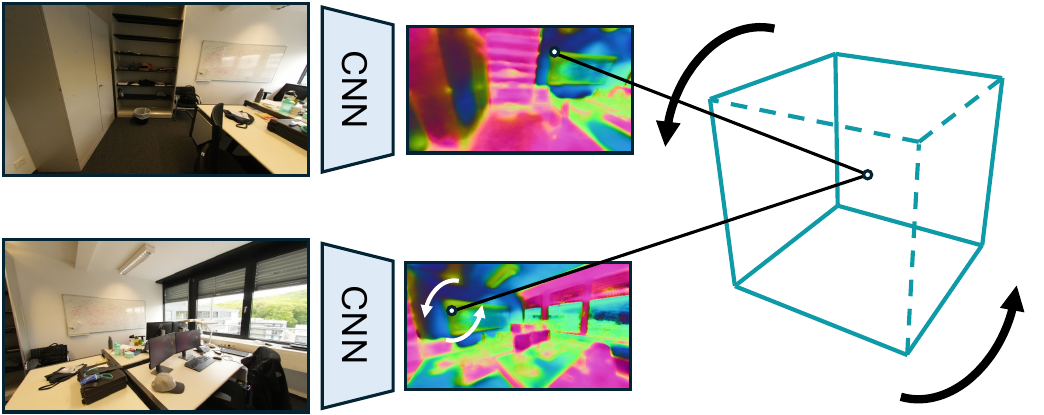}

        \put(70,3) {\scriptsize Cuboid $\mathcal{C}$}
        
        \put(76,14){\scriptsize $\mathcal{W}_{i \to j}$}

        \put(40,38.5) {\scriptsize Image point $\bm{x}_{ik}$}
        
        \put(10,20) {\scriptsize Image $\II_i$}
        \put(41,22) {\scriptsize Feature map $\FF_i$}

        \put(10,-3) {\scriptsize Image $\II_j$}
        \put(41,-1) {\scriptsize Feature map $\FF_j$}

    \end{overpic}
    \caption{\textbf{Featuremetric alignment with \methodName{}.}
    From the posed input images $\{ \II_i \}$ (two or more) we extract feature maps $\{ \FF_i \}$. Points $\{ \bm{x}_{ik} \}$ are sampled in each feature map and warped ($\mathcal{W}_{i \to j}$) via the cuboid $\mathcal{C}$ to the other views. We find the optimal cuboid 
    by minimizing the featuremetric error (\cref{eq:featuremetric-cost}).}
\end{figure}

%% file: figs/overview.tex
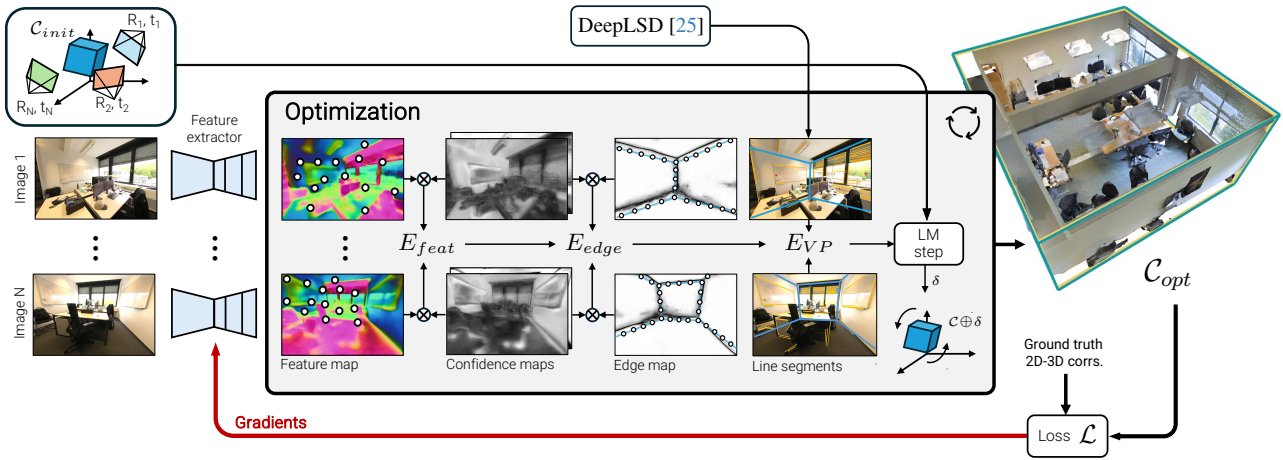
\begin{figure*}[tp]
    \centering
    \begin{overpic}[width=\linewidth]{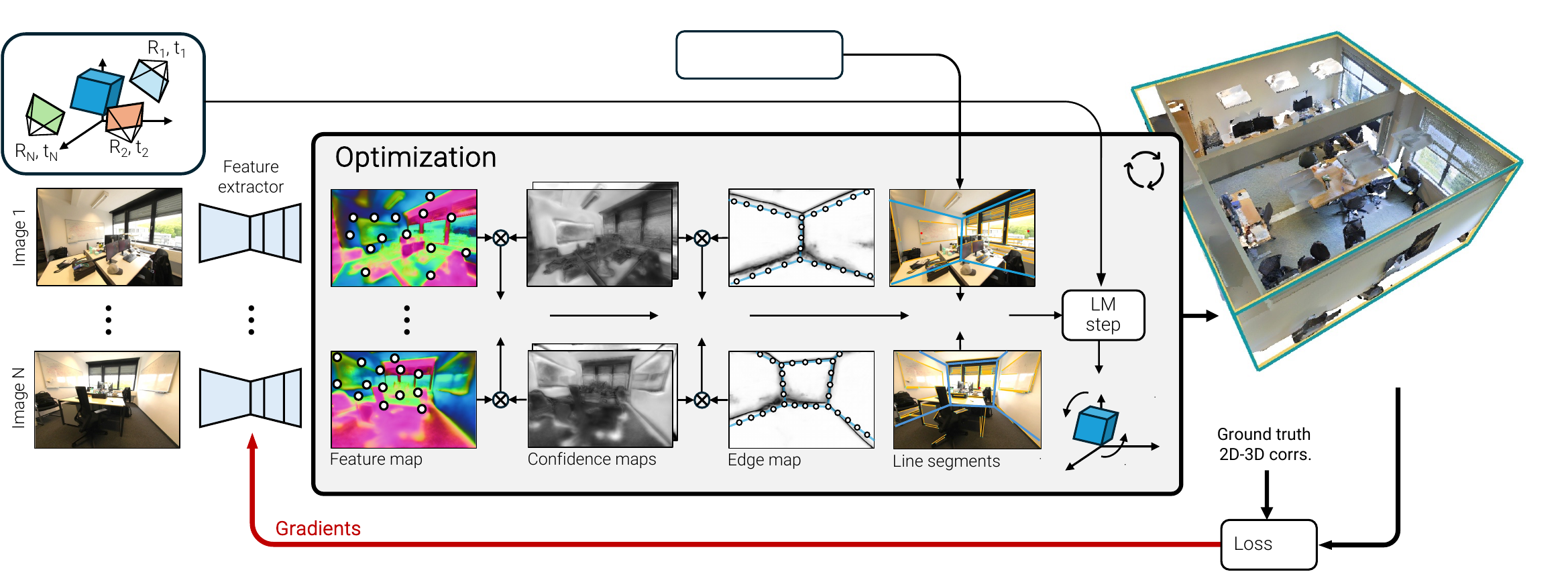}
    \put(2.0,32.3){\scriptsize $\mathcal{C}_{{init}}$}

    \put(43.3,32.5){\resizebox{0.1\textwidth}{!}{ DeepLSD~\cite{pautrat2023deeplsd}}}

    \put(30,15.9){\small $E_{feat}$}
    \put(42.7,15.9){\small $E_{edge}$}
    \put(59.4,15.9){\small $E_{VP}$}
    \put(86.8,13.5){\large $\mathcal{C}_{{opt}}$}

    \put(70.6,13.2){\tiny $\delta$}
    \put(71.9,10.1){\tiny $\mathcal{C}$}
    \put(72.7,10.2){\tiny $\oplus$}
    \put(73.8,10.1){\tiny $\delta$}

    \put(81.8,1.2){$\mathcal{L}$}

    \end{overpic}

    \caption{\textbf{Room Geometry from \methodName}. 
    Taking only posed images as input, \methodName{} estimates the room cuboid $\mathcal{C}_{opt}$ by optimizing from a coarse initial estimate $\mathcal{C}_{init}$ (see Figure~\ref{fig:cuboid-init} for examples).
    First, a deep network predicts dense feature-, confidence- and edge maps for each image.
    The optimization then minimizes a combination of three terms:
    \textbf{a)} a multi-view featuremetric alignment which warps features between images using the cuboid faces, \textbf{b)} a monocular edge cost which tries to align the projected cuboid with the learned edge map, and \textbf{c)} a vanishing point-based cost that aligns the orientation of the cuboid with lines detected in the image. The network is trained end-to-end by supervising on the result of the optimization, propagating gradients back to the network through the optimization steps. 
    }
    \label{fig:overview}
\end{figure*}

%% file: 02_related.tex
\section{Related Work}
\label{sec:related}

\textbf{Featuremetric alignment}, minimizing the distance between extracted features, has been used to tackle a wide variety of problems, for example monocular depth and egomotion estimation \cite{shu2020feature}, camera tracking \cite{xu2020deep}, structure-from-motion \cite{lindenberger2021pixel} and visual localization \cite{sarlin2021back}. These methods typically extract features with a convolutional neural network to improve accuracy and robustness as compared to photometric alignment. Similar to our approach, \cite{xu2020deep} and \cite{sarlin2021back} apply featuremetric alignment at multiple scales, sequentially refining the camera pose and their networks are trained end-to-end via differentiable optimization steps. 
In contrast, we have fixed poses and optimize the scene representation (cuboid).


\textbf{Room layout estimation} from a single view, inferring the location of the floor, ceiling and walls from an image captured indoors, is a well-studied problem. Early algorithms \cite{hedau2009recovering,lee2009geometric} predicted the layout from one perspective image and relied on geometric reasoning with e.g.~vanishing points and line segments. With the emergence of deep learning they were superseded by methods \cite{mallya2015learning,lee2017roomnet,nie2020total3dunderstanding,zhang2021holistic} trained on annotated datasets \cite{song2015sun}. 
Due to the limited field-of-view of the perspective image a lot of focus has recently been given to room layout estimation from a 360\degree{} equirectangular panorama, which contains a more complete view of the surrounding environment. A wide range of architectures have been suggested, for example recurrent \cite{sun2019horizonnet,wang2021led2}, graph convolutional \cite{pintore2021deep3dlayout} and transformer-based \cite{wang2022psmnet,su2023gpr,dong2024panocontext} neural networks. A common approach \cite{sun2019horizonnet,zou2018layoutnet,wang2021led2,su2023gpr} is to identify boundaries between floor, wall and ceiling in the image, from which the layout is derived.
Monocular methods, however, suffer from scale ambiguity and often assume knowledge about the camera height relative to the floor to fix the scale. Another approach is to use multiple views, which only a few prior works have considered. PSMNet \cite{wang2022psmnet} predicts the layout from a pair of panoramic images but requires an approximate relative pose as input, a limitation that the later GPR-Net \cite{su2023gpr} gets rid of by direct regression with a pose transformer. MVLayoutNet \cite{hu2022mvlayoutnet} and the multi-view method of \citet{pintore20183d} can leverage more than two panoramas, but as no public implementations exist we do not compare against these networks.
In this work we formulate layout estimation as an optimization problem that combines geometric cues with learned deep features, extracted from a simple U-Net style network architecture. Our proposed method \methodName{} uses multiple posed perspective views for a larger context and to resolve the global scale. 

Many layout estimation methods make assumptions about the room shape. Due to its simplicity a box or cuboid assumption is commonly employed \cite{hedau2009recovering,nie2020total3dunderstanding,zhang2021holistic}. A less restrictive model is the "Manhattan world" where walls meet at right angles \cite{zou2018layoutnet,sun2019horizonnet,wang2021led2}. \citet{lee2009geometric} proposed the "indoor world" model which extends the Manhattan world with a single-floor, single-ceiling constraint. 
In this work we focus on single cuboid-shaped rooms, but show that our optimization-based approach is flexible and can be extended to other setups, e.g.~apartments consisting of multiple cuboids.

\textbf{Layout from 3D point clouds and RGB-D.}
Several prior works have used laser scans or RGB-D sensors to reconstruct indoor scenes. \citet{xiao2014reconstructing} present a system to produce 3D models from laser points, textured with ground-level photographs to create interactive maps of museums. Later, DNN-based methods \cite{liu2018floornet,chen2019floor,yue2023connecting} extract floor plans from point clouds captured with RGB-D sensors. In contrast, \methodName{} operates on posed RGB images and does not rely on additional information from lasers or depth sensors.

\textbf{Plane reconstruction} is the task of detecting planar surfaces in images and differs from room layout estimation in that not only the floor, ceiling and walls should be reconstructed. Both single view \cite{liu2019planercnn,yu2019single} and multi-view \cite{jin2021planar,agarwala2022planeformers,watson2024airplanes} methods exist. PlanarRecon \cite{xie2022planarrecon} and UniPlane \cite{huang2024uniplane} are two recent methods that focus on plane reconstruction from posed monocular videos by the use of 3D feature volumes.

For an in-depth review of reconstruction methods for indoor environments we refer the interested reader to the survey paper of \citet{pintore2020state}.

%% file: 03_method.tex
\section{Method}
\label{sec:method}


We assume that the room shape can be represented by a cuboid $\mathcal{C}$, where the six faces correspond to the walls, floor and ceiling.
As input, our method takes a collection of images $\II_1, \II_2, \dots, \II_n$ captured inside the room, together with their camera poses $(\bm{R}_i,\bm{t}_i)$ and intrinsics $\bm{K}_i$.

To estimate the room cuboid $\mathcal{C}$ we propose an optimization-based approach where an initial cuboid is refined using both multi-view consistency and monocular cues.
Each image is passed independently through a CNN that produce dense feature maps, which are used to define the cost function in the optimization.
Similar to \cite{sarlin2021back}, the network is trained end-to-end by only supervising on the result of the cuboid optimization, ensuring that the features we learn provide useful cues for the cuboid fitting. 
See \cref{fig:overview} for an overview of our method which we call \methodName.

In the next section we detail the cuboid parameterization, followed by definition of the cost functions (\cref{subsec:geometric-optimization-of-cuboids}). 
\cref{subsec:learning-to-optimize-room-layouts} describes how the feature extractor is trained and finally in \cref{subsec:initializing-the-optimization} we propose a simple heuristic for finding initial cuboid estimates.

\subsection{Cuboid Parameterization}

Let $\bm{R} \in SO(3)$ be the rotation that rotates the coordinate system such that the cuboid is axis aligned.
We then parameterize the offsets $\bm{d} \in \mathbb{R}^6$ for each of the six faces of the cuboid along either the x-, y- or z-axis. So the first face is defined by the plane $(1,~0,~0)\bm{X} = d_1$, and so on.
Together, the rotation $\bm{R}$ and vector $\bm{d}$ then minimally parameterizes the 9 degrees of freedom of the cuboid $\mathcal{C}$.
The translation of the cuboid is encoded in the offsets $\bm{d}$.
Note that the parameterization is not unique, as the axes can be switched by changing the rotation and switching corresponding elements of $\bm{d}$.
%
%
An important aspect of this parameterization (compared to e.g.~explicitly parameterizing the translation) is that it decouples the parameters for each face.
This allow us to easily perform optimization over a subset of faces in cases where the cuboid is only partially observable.

%
%
%



\subsection{Geometric Optimization of Cuboids}
\label{subsec:geometric-optimization-of-cuboids}

To refine cuboids we define a cost function depending on three terms:
\textbf{a)} a featuremetric cost $E_{feat}$, measuring multi-view consistency by performing alignment between feature maps,
    \textbf{b)} a monocular edge cost $E_{edge}$, forcing the projected edges of the cuboid to align with a predicted edge map,
    and \textbf{c)}  a VP-based cost $E_{VP}$, where the vanishing points defined by the cuboid is compared with extracted line segments.
The full cost function is then given by
\begin{equation}
    E(\mathcal{C}) = E_{feat}(\mathcal{C}) + 
    \alpha E_{edge}(\mathcal{C}) + 
    \beta E_{VP}(\mathcal{C}).
\end{equation}
If 2D line segments are not available we set $\beta = 0$.
The three terms are further detailed in the following paragraphs.

\noindent\textbf{Featuremetric cost}: Inspired by the success of PixLoc \cite{sarlin2021back}, we leverage featuremetric alignment. From each image $\II_i  \in \mathbb{R}^{W \times H \times 3}$ a dense feature map $\FF_i \in \mathbb{R}^{W \times H \times D}$ is extracted with a CNN. 
The featuremetric cost function ${E}_{feat}$ is then defined by measuring the consistency of the warped features using the faces of the cuboid, i.e. ${E}_{feat}(\mathcal{C}) =$
\begin{equation}
\label{eq:featuremetric-cost}
    \sum_{i,j} \sum_k w_{ijk} \rho \left( \left\| \FF_i[\xx_{ik}] - \FF_j\left[\mathcal{W}_{i\to j}(\xx_{ik}, \mathcal{C})\right] \right\|^2 \right).
\end{equation}
Here $\{ \xx_{ik} \}$ is a set of sampled image points in image $\II_i$ and $\mathcal{W}_{i\to j}$ represents the warping of these points to image $\II_j$ via the cuboid, i.e.~first projecting them onto the planes of $\mathcal{C}$ and then into $\II_j$.
This warp is a differentiable function of the cuboid parameters allowing us to optimize over $\mathcal{C}$.
$[\cdot]$ denotes lookup with sub-pixel interpolation and $\rho$ is a robust loss function.
The residual weights $ w_{ijk} = \CF_i[\xx_{ik}] \CF_j[\mathcal{W}_{i\to j}(\xx_{ik}, \mathcal{C})]$ are interpolated from the confidence images $\CF_i, \CF_j \in \mathbb{R}^{W \times H}$ which are predicted by the network. We also utilize $\CF_i$ for point sampling: the image points $\{ \xx_{ik} \}$ are drawn, without replacement, from the probability map $\CF_i^\gamma$, where $\gamma \in \mathbb{R}$.


\noindent\textbf{Cuboid edge cost}: In addition to the feature map $\FF_i$ we also let the CNN predict a dense edge map $\EE_i \in \mathbb{R}^{W \times H}$ which aims to delineate the room edges. We sample 3D points $\{ \bm{X}_j \}$ uniformly on the twelve edges of the cuboid $\mathcal{C}$.
The edge cost $E_{edge}$ is then defined by evaluating the edge map $\EE_i$ at the projections of these points, i.e.
\begin{equation}
    E_{edge}(\mathcal{C}) =
    \sum_i \sum_j w_{ij} \EE_i[\Pi_i(\bm{R}_i \bm{X}_j + \bm{t}_i)]^2,
\end{equation}
where $\Pi_i : \mathbb{R}^3 \to \mathbb{R}^2$ is the projection into the image using the known intrinsics $\bm{K}_i$. 
As with $E_{feat}$, we predict a confidence map $\CE_i$ from which the weights $w_{ij}$ are interpolated.


\noindent\textbf{Vanishing point cost}: 
Finally, the VP cost $E_{VP}$ measures consistency between line segments in the images and the orientation of the cuboid.
Projected into image $\II_i$, the cuboid defines three vanishing points, given by the columns of $\bm{R}_i \bm{R}^T = \begin{bmatrix} \bm{v}_{i,1} & \bm{v}_{i,2} & \bm{v}_{i,3} \end{bmatrix}$. 
We extract line segments $\{ \bm{l}_{ij} \}$ from the image using DeepLSD \cite{pautrat2023deeplsd} and use the consistency measure of \cite{tardif2009non} between line segments and vanishing points to form the vanishing point cost function
\begin{equation}
    E_{VP}(\mathcal{C}) =
    \sum_i \sum_j \min \left( \min_{k\in\{1,2,3\}} D_{VP}(\bm{l}_{ij}, \bm{v}_{ik}), \tau \right)^2 ,
\end{equation}
where each line is softly assigned to one of the three VPs by only considering the minimum residual.
Here
%
\begin{equation}
    D_{VP}(\bm{l},\bm{v}) = |\hat{\bm{l}}^T \bm{l}_1| / {\sqrt{\hat{l}_1^2+\hat{l}_2^2}}
\end{equation}
denotes the distance between the line segment endpoint $\bm{l}_1$ and a line $\hat{\bm{l}} = \bar{\bm{l}} \times \bm{v}$ passing through its midpoint $\bar{\bm{l}}$ and the vanishing point $\bm{v}$. We cap the distance at $\tau$ to account for line segments that are not aligned with any of the cuboid sides.
Note that the lines do not need to coincide with the outline of the room to be consistent with the vanishing points.

\noindent\textbf{Coarse-to-fine optimization}:
The feature network follows a U-Net style~\cite{ronneberger2015u} architecture with a ResNet-101 \cite{he2016deep} encoder and our multi-scale decoder predicts the feature-, confidence- and edge maps at three different resolutions (see \cref{fig:network}). 
Similar to \cite{sarlin2021back}, the optimization is then performed in a coarse-to-fine manner. 
For each scale level $s \in \left\{ 
    \text{coarse}, \text{medium}, \text{fine} \right\}$
the cost is minimized using the corresponding network outputs,
\begin{equation}
    \mathcal{C}^s = \arg\min_{\mathcal{C}} E(\mathcal{C}, \FF^s, \CF^s, \EE^s, \CE^s),
\end{equation}
each optimization initialized from the output of the previous scale, i.e.~medium scale is starting from $\mathcal{C}^{\text{coarse}}$ and so on.
The minimization is performed using standard Levenberg-Marquardt~\cite{levenberg1944method,marquardt1963algorithm} optimization, iterated until convergence.


\input{figs/feature_extractor}

\subsection{Learning to Optimize Room Layouts}
\label{subsec:learning-to-optimize-room-layouts}
For training we assume that we have posed images together with a ground truth 3D mesh that has semantic labels for walls, ceiling and floor. 
We train the network end-to-end, propagating gradients through the optimization process.
During training, the same coarse-to-fine approach is used, but for each scale $s$ we compute $\mathcal{C}^s$ using a fixed number of unrolled Levenberg-Marquardt iterations minimizing $E(\mathcal{C})$.
Supervision for the network is only applied on resulting optimized cuboids $\mathcal{C}^s$.
This forces the network to learn features and confidence maps that are useful for optimization.

As rooms are not perfectly cuboid shaped, we propose to supervise using points sampled from the ground truth mesh.
We sample points from the mesh corresponding to the wall, ceiling and floor semantic labels and project into the images, filtering for occlusion using the ground truth depth maps.
This yields a set of 2D-3D correspondences $\{ (\bm{x}_{ik}^{GT}, \bm{X}_{ik}^{GT}) \}$ for each image.
The loss for a cuboid is then computed as
\begin{equation}
\label{eq:loss}
    \mathcal{L}(\mathcal{C}) =
    \frac{1}{N}
    \sum_{i,j} \sum_k \| \mathcal{W}_{i\to j}(\bm{x}_{ik}^{GT}, \mathcal{C}) - \Pi_j (\bm{R}_j \bm{X}_{ik}^{GT} + \bm{t}_j) \|^2,
\end{equation}
where $N$ is the total number of points successfully warped between images.
The loss is applied after each scale level and our network is trained to minimize
\begin{equation}
    \mathcal{L}_{total} = \mathcal{L}(\mathcal{C}^{\text{coarse}}) + \gamma_m\mathcal{L}(\mathcal{C}^{\text{medium}}) + \gamma_f\mathcal{L}(\mathcal{C}^{\text{fine}}) ,
\end{equation}
where $\gamma_m$ and $\gamma_f$ are the relative weightings of the scales.
After each scale we check whether the optimization failed (loss above some threshold), and if so stop gradients to the next scale (setting $\gamma_m$ or $\gamma_f$ to zero for this instance).
This prevents oversmoothing the fine and medium feature maps.

For the coarse optimization the cuboid parameters are initialized from the ground truth layout (see \cref{subsec:datasets} for details), perturbed by random rotation, translation and scaling, so that the network is exposed to examples of varying difficulty throughout the training.
In addition to the network parameters we learn a separate LM dampening factor for each of the nine cuboid parameters similar to \cite{sarlin2021back}.

\noindent\textbf{Pre-training}:
As the training procedure requires the network to at least be somewhat successful to get meaningful gradients from the optimization, we perform some simple pre-training.
While featuremetric alignment can work reasonably with off-the-shelf ImageNet pre-trained features~\cite{lindenberger2021pixel} , this is not the case for the edge loss.
%
Therefore we pre-train only the edge map, supervising with a weighted MSE loss against a line-drawing of the projected ground truth cuboid. 

\subsection{Initializing the Optimization}
\label{subsec:initializing-the-optimization}

At inference time the cuboid is initialized based on the camera poses. We set the cuboid z axis to the mean of the camera negative y axes (assumed to be pointing down on average). The x and y axes of the cuboid are selected by randomly sampling two basis vectors in the plane orthogonal to the z axis. We then run a few iterations of optimization with $E_{VP}$ to further enhance the orientation of the initial cuboid. Plane offsets $\bm{d}$ are set so that the cuboid contains all cameras with some margin. In \cref{subsec:ablation-experiments} we compare this initialization strategy to using uniformly sampled rotation matrices $\bm{R}$ and also validate the gain of vanishing point optimization.

%% file: figs/feature_extractor.tex
\begin{figure}
    \centering
    \includegraphics[width=\columnwidth]{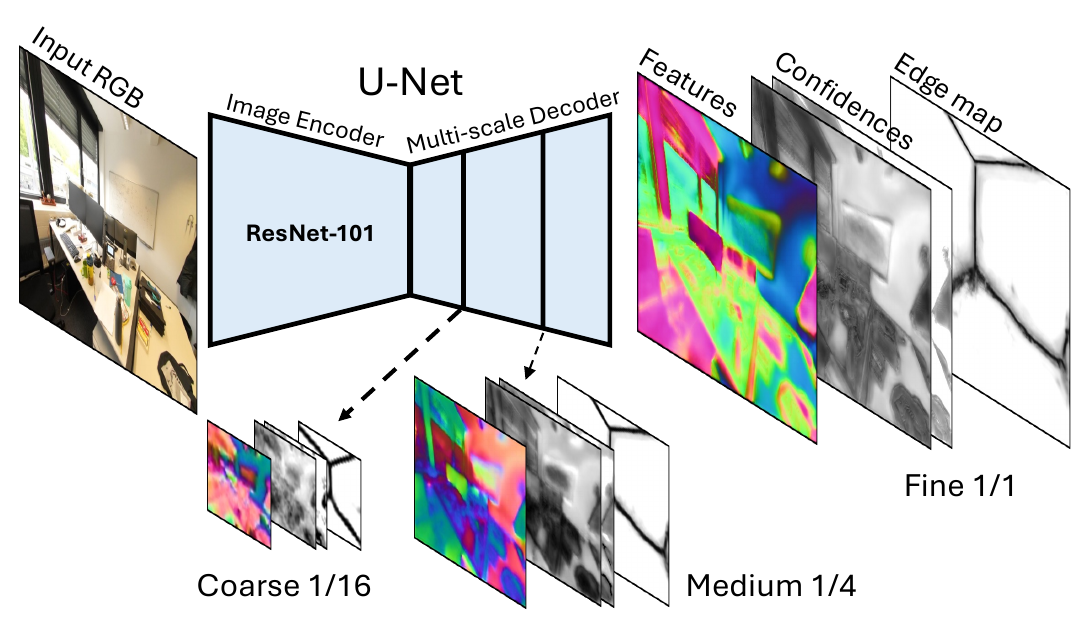}
    
    \caption{\textbf{Network Architecture}.
    Our feature extractor uses a U-Net~\cite{ronneberger2015u} architecture consisting of a ResNet-101 image encoder followed by a multi-scale decoder extracting dense feature-, confidence- and edge maps.
    }
    \label{fig:network}
\end{figure}

%% file: 04_experimental_setup.tex
\section{Experimental Setup}
\label{sec:experimental-setup}
For the experiments we use the following settings:
$\alpha = 0.05$, $\beta = 40$ and $\tau = 0.05$. The number of LM iterations on each scale level is 15.
Cuboids are initialized as described in \cref{subsec:initializing-the-optimization}, using a margin of 2.5 m between camera centers and cuboid faces.
We additionally expand the cuboid after each optimization step if needed so that the cameras are at least 0.1 m from the faces.

\subsection{Datasets}
\label{subsec:datasets}

\textbf{ScanNet++}:
From ScanNet++ v2 \cite{yeshwanth2023scannet++}, an indoor dataset comprised of 1006 scenes of varying types, we create a new benchmark specifically to evaluate multi-view cuboid room layout estimation. For each scene we use the ground truth semantic mesh to fit a cuboid to the vertices classified as either "floor", "wall" or "ceiling" by minimizing the distance between the cuboid and the vertices using L-BFGS \cite{liu1989limited}. Since not all scenes are cuboid-shaped, and because the L-BFGS optimization occasionally fails, we manually inspect the results to determine which scenes should be included. This way 391 out of 856 scenes in the existing ScanNet++ v2 training set were selected as a new training set. From the existing validation set we pick 28 out of 50 scenes. As the test scenes of ScanNet++ do not include any semantic mesh we split the 28 validation scenes into new validation and test sets with 10 and 18 scenes, respectively.

For every scene in this new dataset we sample image tuples, consisting of ten random DSLR images, to be used as the input to multi-view room layout estimation methods. For each training scene, 250 tuples are sampled and 20 tuples for each of the validation and test scenes. In total there are hence $391 \times 250 = 97750$ image tuples for training, $10 \times 20 = 200$ for validation and $18 \times 20 = 360$ for testing. 
For experiments with $k < 10$ images, we select the first $k$ from each tuple.


\textbf{2D-3D-Semantics}:
To enable comparison with panorama-based room layout estimation methods, and to assess the generalization capabilities of \methodName{}, we use 2D-3D-Semantics \cite{armeni2017joint}. This is an indoor dataset divided into six areas in three different buildings. Each area is further split into spaces, for example offices or hallways. Similar to ScanNet++ v2 we fit a cuboid to every space by minimizing its distance to points labeled "floor", "wall" or "ceiling" in the supplied point cloud. Again we inspect the results to find cuboid-shaped spaces. In addition, only spaces with at least two panorama images captured within the fitted cuboid are considered. For spaces with more than two such panoramas we randomly select two of them. The result is a set of 160 spaces that can be used for evaluation. As we do not train or tune our model on 2D-3D-Semantics no training/validation/test split is performed. 
The panorama images for all spaces are divided into four perspective views with 90\degree{} horizontal field-of-view to allow comparison between methods that work on the two different types of images.

We will make the ground truth cuboids, image tuples and evaluation code publicly available for both datasets.

\subsection{Metrics}
\label{subsec:metrics}

We evaluate the estimated layouts with the commonly used 3D Intersection over Union (IoU) and the Chamfer distance between the predicted layout and the ground truth cuboid. For methods that predict a cuboid-shaped layout we compute the rotation error between the two cuboids, taking rotational symmetries into account. We also report the area under the recall curve for the rotation error, using one coarse (20\degree) and one fine (1\degree) threshold. These metrics are averaged over image tuples (ScanNet++) or spaces (2D-3D-Semantics). For single-view methods we only consider the prediction with the highest IoU for each tuple/space.

The predicted and ground truth room layouts are rendered into the perspective views and we compute the depth RMSE and the ratio of pixels for which the normal angle error is less than 10\degree, averaged over all images. For single-view panorama methods the prediction corresponding to the perspective view is used.
Finally we measure the mean time in seconds to predict the room layout for one tuple or space. 

\subsection{Training Details}
\label{subsec:training-details}

We train our model on the ScanNet++ training set and tune its hyperparameters on the validation set (using our data split as explained in \cref{subsec:datasets}). The five first images in each tuple, which are undistorted beforehand, are used as input. We utilize a two-stage training process as described in \cref{subsec:learning-to-optimize-room-layouts}. First the edge maps are pre-trained with a weighted MSE loss, followed by training of the full network with the loss in \cref{eq:loss}.
See our supplementary material for a more detailed description of the training process.

%% file: 05_results.tex
\section{Results}
\label{sec:results}

\subsection{Room Layout Estimation}
\label{subsec:room-layout-estimation}

\input{tables/methods}

\textbf{Baselines}: On ScanNet++ \methodName{} is compared against the single-view layout estimation methods Total3DUnderstanding \cite{nie2020total3dunderstanding} and Implicit3DUnderstanding \cite{zhang2021holistic}. They both take a single perspective image as input and simultaneously predict the room layout, camera pose and 3D object bounding boxes and meshes.
On 2D-3D-Semantics we also compare with three recent panorama-based methods: Deep3DLayout \cite{pintore2021deep3dlayout}, LED\textsuperscript{2}-Net \cite{wang2021led2} and PSMNet \cite{wang2022psmnet}. Deep3DLayout and LED\textsuperscript{2}-Net predict the room layout from a single panoramic view while PSMNet uses two panorama images.
To the networks LED\textsuperscript{2}-Net and PSMNet we input the average camera height over the dataset from the ground truth. PSMNet does not estimate the room height so we give it the height of the corresponding ground truth cuboid as well, along with the relative pose between the two panoramas (without added noise). For all methods we use the authors' official implementations and their provided pre-trained network weights. For Deep3DLayout we utilize the weights fine-tuned on Pano3DLayout \cite{pintore2021deep3dlayout}. For a summary of all methods, see \cref{tab:methods}.

The two-view GPR-Net \cite{su2023gpr}, the multi-view MVLayoutNet \cite{hu2022mvlayoutnet} and the work of \citet{pintore20183d} are also relevant competing methods, but as no public implementations are available we do not include them in our benchmark.

\input{tables/room_layout}

\textbf{Results}: For ScanNet++ we run Total3DUnderstanding, Implicit3DUnderstanding and \methodName{} on the five first images of each image tuple in the test set. As described in \cref{subsec:metrics} we use the best prediction (highest IoU) out of five for the two single-view methods. The results are presented in \cref{tab:room-layout-results} (top section). \methodName{} outperforms the two competing methods by a large margin on all metrics (except runtime), however it should be noted that our method is the only one trained on ScanNet++ v2.

On 2D-3D-Semantics the perspective methods (Total3DUnderstanding, Implicit3DUnderstanding) make eight predictions (two panoramas, split into four perspective views) per space. We also run Deep3DLayout and LED\textsuperscript{2}-Net which predict two room layouts for each space (one per panorama image). Again we emphasize that for all these single-view methods only the top prediction is included in the metrics. PSMNet and \methodName{} outputs a single room layout prediction from the two panoramas and the eight perspective images, respectively (thus taking the same image data as input). We report the results in \cref{tab:room-layout-results} (bottom section). The panorama methods generally predict better layouts than the single-view perspective-based ones, but are outperformed by \methodName{} on all metrics.

\input{figs/example_layout}

In \cref{fig:example-layout} we show qualitative examples of predicted room layouts in 2D-3D-Semantics. For single-view methods the room layout with the highest IoU metric is visualized. Some failure cases of \methodName{} are visualized in \cref{fig:failure-cases}. Due to the vanishing point cost $E_{VP}$ our method is often able to estimate the orientation of the cuboid accurately, even when the plane offsets $\bm{d}$ cannot be properly determined.

\input{figs/failure_cases}

\subsection{Ablation Experiments}
\label{subsec:ablation-experiments}

\input{tables/ablations_small}

We validate the design of \methodName{} in a series of ablation experiments on our ScanNet++ v2 test set. In addition to previous metrics we also report the success rate on the finest scale 
(proportion of image tuples with average warp error \eqref{eq:loss} smaller than 3 pixels).

First, the learned feature maps are compared to those of PixLoc (trained on Extended CMU Seasons \cite{Sattler2018CVPR,Badino2011}) and to using the images directly (i.e. photometric alignment). Here only the featuremetric cost $E_{feat}$ is considered in the LM optimization ($\alpha = \beta = 0$). Results are shown in the top section of \cref{tab:ablation-study}, from which it is clear that learning features specifically for room layout estimation is beneficial.

Next, we try different combinations of the cost functions $E_{feat}$, $E_{edge}$ and $E_{VP}$ (\cref{tab:ablation-study}, second section). Using $E_{edge}$ on its own (second row) gives better layout predictions than with only featuremetric cost $E_{feat}$ (first row), showcasing the significance of the learned edge map. The two costs complements each other as seen on row 3, where the success rate increases by almost 18 percentage points compared to optimization with just $E_{edge}$. Adding the vanishing point cost $E_{VP}$ results in better metrics across the board (rows 4-6). This cost is particularly helpful in determining the orientation of the cuboid, with large reductions in the rotation error and better estimation of the surface normals.

To study the impact of point sampling we compare the guided sampling outlined in \cref{subsec:geometric-optimization-of-cuboids} to random sampling and to sampling from the ground truth points $\bm{x}_{ik}^{GT}$ that lie on the floor, wall or ceiling. We train models with these different point sampling approaches and give the results in the third section of \cref{tab:ablation-study}. Guided sampling is used during evaluation. The results show that training with guided sampling performs the best, although the differences are minor.

We run \methodName{} on both lower and higher resolution input images (resized to 256 and 768 pixels in height, respectively) and compare to our standard size 512 px, which is what the network sees during training (\cref{tab:ablation-study}, section four). The success threshold is adjusted to account for the difference in image dimensions. Our model performs only slightly worse with the low resolution images. There is no gain in using a higher resolution - the best performance is reached with the medium sized 512 px input.

\input{figs/cuboid_init}

The importance of cuboid initialization is examined by employing four different strategies. As a baseline we sample the orientation $\bm{R}$ uniformly and then select plane offsets $\bm{d}$ so that all cameras are inside the cuboid, with a margin of 2.5 m on each side (\cref{tab:ablation-study}, section five, first row). The proposed initialization scheme in \cref{subsec:initializing-the-optimization}, where $\bm{R}$ is set based on the mean camera y axis (assumed to be pointing down), performs significantly better (third row). For both of these initialization procedures we also try running five iterations of cuboid optimization with $E_{VP}$, after which $\bm{d}$ is re-computed so that all cameras are again contained by the cuboid with the same 2.5 m margin (rows two \& four). We see that this extra step results in improved layouts, especially when using random initialization. \cref{fig:cuboid-init} shows examples of cuboids initialized with our proposed method.

Lastly the coarse-to-fine optimization is ablated in the last section of \cref{tab:ablation-study}. Here we stop the optimization after the first ("coarse", first row) and second ("medium", second row) scale level and compare to using all three scales (third row). A reasonable room layout is generally found already after the coarse optimization, but it is not so accurate. At the second scale the cuboid is refined, resulting in an increase in the success rate by almost 30 pp. At the last scale level we see only a marginal improvement to the metrics.

\input{figs/num_images}

In \cref{fig:num-images} we also look at how the quality of \methodName{}'s predictions vary with the number of input views. As expected the IoU increases with the number of views, with the largest gain seen when going from two to three views.

\subsection{Multi-room Layout Estimation}
\label{subsec:multi-room-layout-estimation}

\input{figs/replica}

As a final experiment we apply \methodName{} to a multi-room scenario: a scene
from the Replica \cite{replica19arxiv} dataset with four interconnected rooms. We use the trajectory and pre-rendered frames from \cite{bruns2024neural}. A simple algorithm is devised to handle multiple rooms. We start by subsampling the list of frames by a factor of 60, and then loop over the remaining frames in order. When 8 frames have been accumulated, the cuboid optimization is run.
In this experiment we optimize the new cuboid together with previously added ones, with shared orientation, floor/ceiling height and wall locations.
Then, we skip over subsequent frames until the camera has moved outside the last cuboid at which point we start accumulating new frames and repeat the process. A new cuboid is accepted only if it does not overlap (IoU $> 0.01$) with existing ones. Results are presented in \cref{fig:replica}.

%% file: tables/methods.tex
\begin{table}
\centering
\def\imageicon{\raisebox{-0.05cm}{\includegraphics[width=0.4cm]{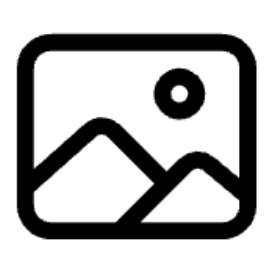}}}
\def\panoicon{\raisebox{-0.05cm}{\includegraphics[width=0.4cm]{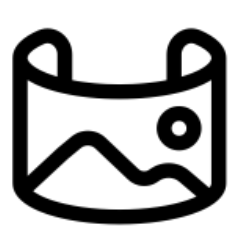}}}

\resizebox{\columnwidth}{!}{
\begin{tabular}{l r r r}
  \toprule
  Method & Input & Output & Trained on \\
  \midrule
  Total3D \cite{nie2020total3dunderstanding} & 1$\times$\imageicon & Cuboid & \footnotesize SUN RGB-D \cite{song2015sun}, Pix3D \cite{sun2018pix3d} \\
  Implicit3D \cite{zhang2021holistic} & 1$\times$\imageicon & Cuboid & \footnotesize SUN RGB-D \cite{song2015sun}, Pix3D \cite{sun2018pix3d} \\
  Deep3DLayout \cite{pintore2021deep3dlayout} &  1$\times $\panoicon& Mesh & \footnotesize MatterportL.~\cite{zou2021manhattan}, Pano3DL.~\cite{pintore2021deep3dlayout} \\
  LED\textsuperscript{2}-Net \cite{wang2021led2} & 1$\times$\panoicon & Polygon & \footnotesize Realtor360 \cite{yang2019dula} \\
  PSMNet \cite{wang2022psmnet} & 2$\times$\panoicon & Polygon  & \footnotesize ZInD \cite{cruz2021zillow} \\
  \methodName{} (ours) & N$\times$\imageicon & Cuboid & \footnotesize ScanNet++ v2 \cite{yeshwanth2023scannet++} \\
  \bottomrule
\end{tabular}
}
\caption{An overview of the room layout estimation methods included in our experimental evaluation. The methods take a varying number of perspective (\imageicon) or panoramic (\panoicon) images as input.}

\label{tab:methods}
\end{table}

%% file: tables/room_layout.tex
\begin{table*}[b]
\centering
\begin{threeparttable}
    \begin{tabular}{c l cc ccc cc c}
        \toprule
        && \multicolumn{2}{c}{3D} & \multicolumn{3}{c}{Rotation} & \multicolumn{2}{c}{Pixel-wise} \\
        \cmidrule(lr){3-4} \cmidrule(lr){5-7} \cmidrule(lr){8-9}
        & Method & IoU $\uparrow$ & Chamfer $\downarrow$ & Mean $\downarrow$ & \multicolumn{2}{c}{AUC@\{1,20\}\degree $\uparrow$} & Depth $\downarrow$ & Normal $\uparrow$ & Time $\downarrow$ \\
        \midrule
        \multirow{3}{*}{\rotatebox{90}{\small ScanNet}} & Total3D \cite{nie2020total3dunderstanding} {\tiny CVPR20} & 34.5 & 1.44 m & 13.7\degree & 0.0 & 37.6 & 0.86 m & 37.2 & \textbf{0.16 s} \\
        & Implicit3D \cite{zhang2021holistic} {\tiny CVPR21} & 30.8 & 1.60 m & 16.4\degree & 0.0 & 31.0 & 1.02 m & 33.1 & \textbf{0.16 s} \\
        & \methodName{} (ours) & \textbf{87.2} & \textbf{0.22 m} & \textbf{1.3\degree} & \textbf{45.7} & \textbf{95.3} & \textbf{0.09 m} & \textbf{96.1} & 0.31 s \\
        \midrule
        \multirow{6}{*}{\rotatebox{90}{\small 2D-3D-S}} & Total3D \cite{nie2020total3dunderstanding} {\tiny CVPR20} & 31.8 & 1.57 m & 8.9\degree & 0.0 & 56.9 & 1.41 m & 44.1 & 0.32 s \\
        & Implicit3D \cite{zhang2021holistic} {\tiny CVPR21} & 31.6 & 1.59 m & 10.4\degree & 0.0 & 51.5 & 1.51 m & 37.6 & \textbf{0.15 s} \\
        & Deep3DLayout \cite{pintore2021deep3dlayout} {\hfill \tiny TOG21} & 58.8 & 0.68 m & N/A & N/A & N/A & 0.44 m & 52.6 & 1.30 s \\
        & LED\textsuperscript{2}-Net\tnote{\textdagger}~ \cite{wang2021led2} {\tiny CVPR21} & 68.7 & 0.45 m & N/A & N/A & N/A & 0.40 m\tnote{*} & 34.6\tnote{*} & 0.71 s \\
        & PSMNet\tnote{\textdagger}~\tnote{$\ddagger$}~ \cite{wang2022psmnet} {\tiny CVPR22} & 43.5 & 1.04 m & N/A & N/A & N/A & 0.63 m\tnote{*} & 51.4\tnote{*} & 2.32 s \\
        & \methodName{} (ours) & \textbf{89.0} & \textbf{0.18 m} & \textbf{0.7\degree} & \textbf{48.8} & \textbf{97.0} & \textbf{0.10 m} & \textbf{96.1} & 0.42 s \\
        \bottomrule
    \end{tabular}
    \caption{Room layout estimation results on our ScanNet++ v2 test set and the cuboid-shaped spaces of 2D-3D-Semantics.
    For single-view methods the metrics are computed using the best prediction for each image tuple or space.}
    \begin{tablenotes}
        \small
        \item [\textdagger] Uses the ground truth camera height.
        \item [$\ddagger$] Uses the ground truth room height and \textit{vanishing angle} (see code of \cite{wang2022psmnet} for details).
        \item [*] Excludes views (10 for LED\textsuperscript{2}-Net and 104 for PSMNet, out of 1280) where the layout does not contain the camera.
    \end{tablenotes}
    \label{tab:room-layout-results}
\end{threeparttable}
\end{table*}

%% file: figs/example_layout.tex
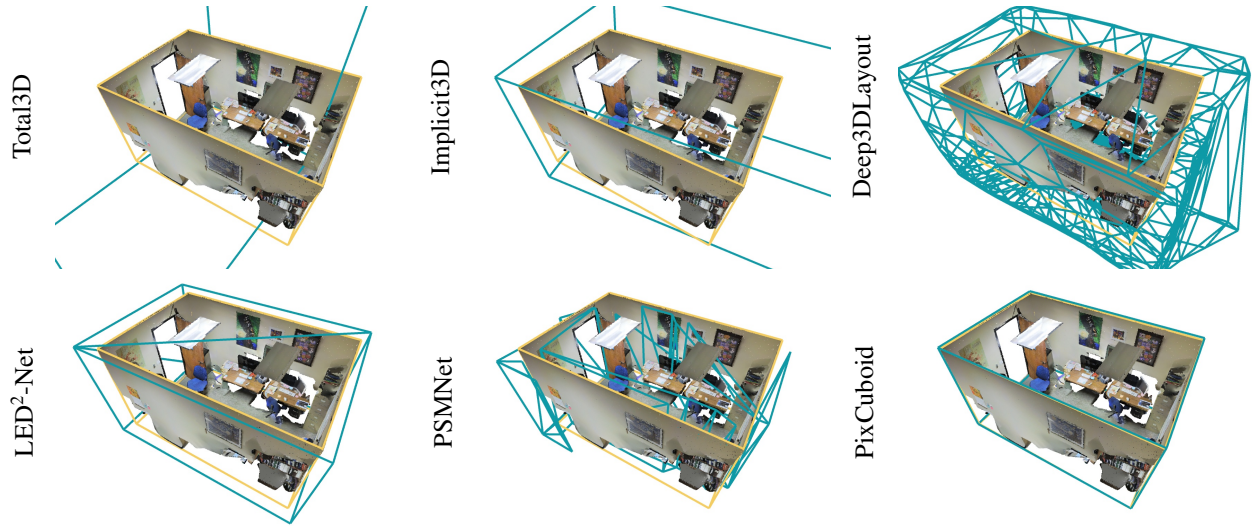
\begin{figure*}[tp]
    \centering
    \begin{tikzpicture}[scale=1.0]
        \node at (0.0,3.5) {\includegraphics[scale=0.13]{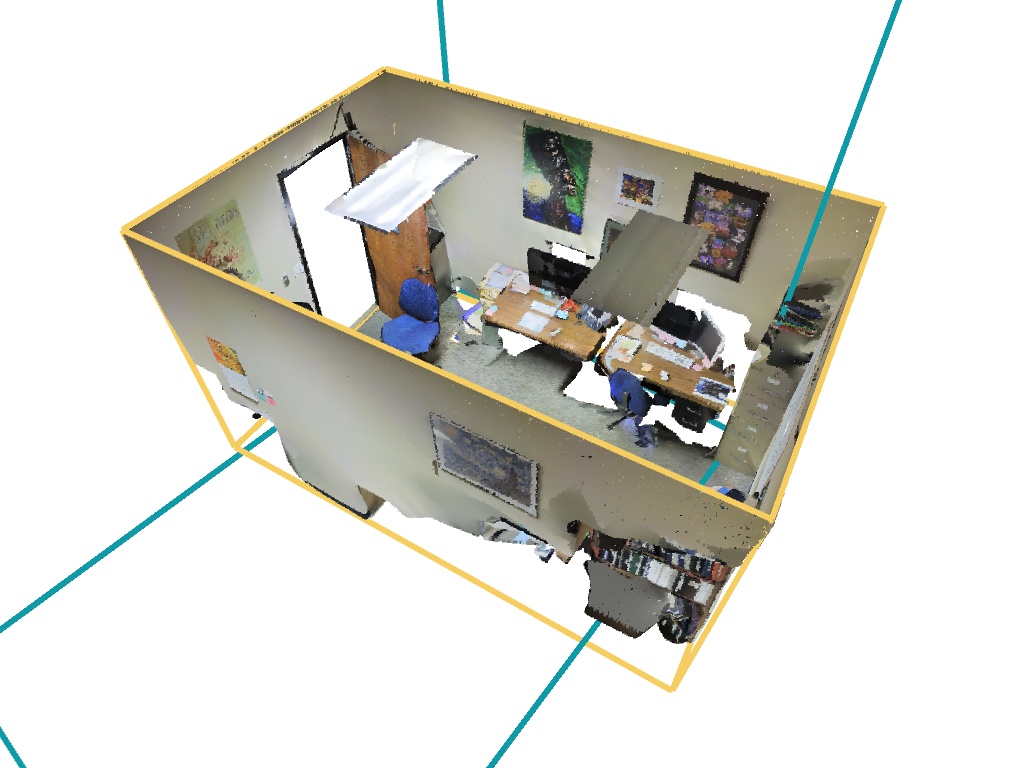}};
        \node at (5.6,3.5) {\includegraphics[scale=0.13]{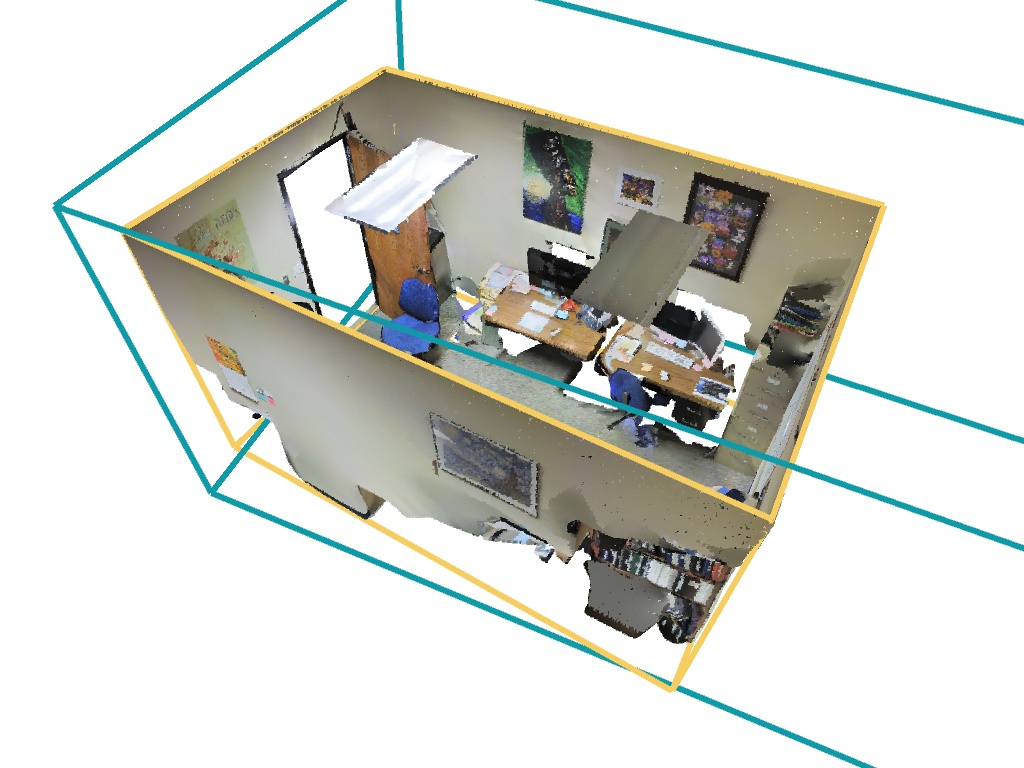}};
        \node at (11.2,3.5) {\includegraphics[scale=0.13]{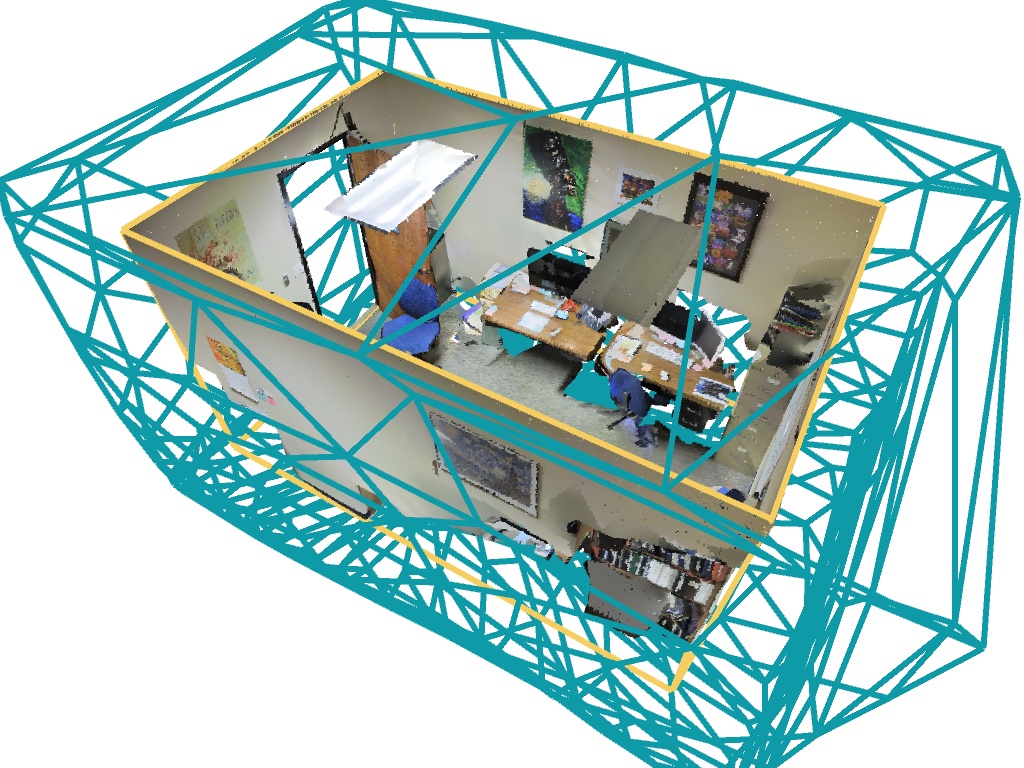}};
        \node at (0.0,0.0) {\includegraphics[scale=0.13]{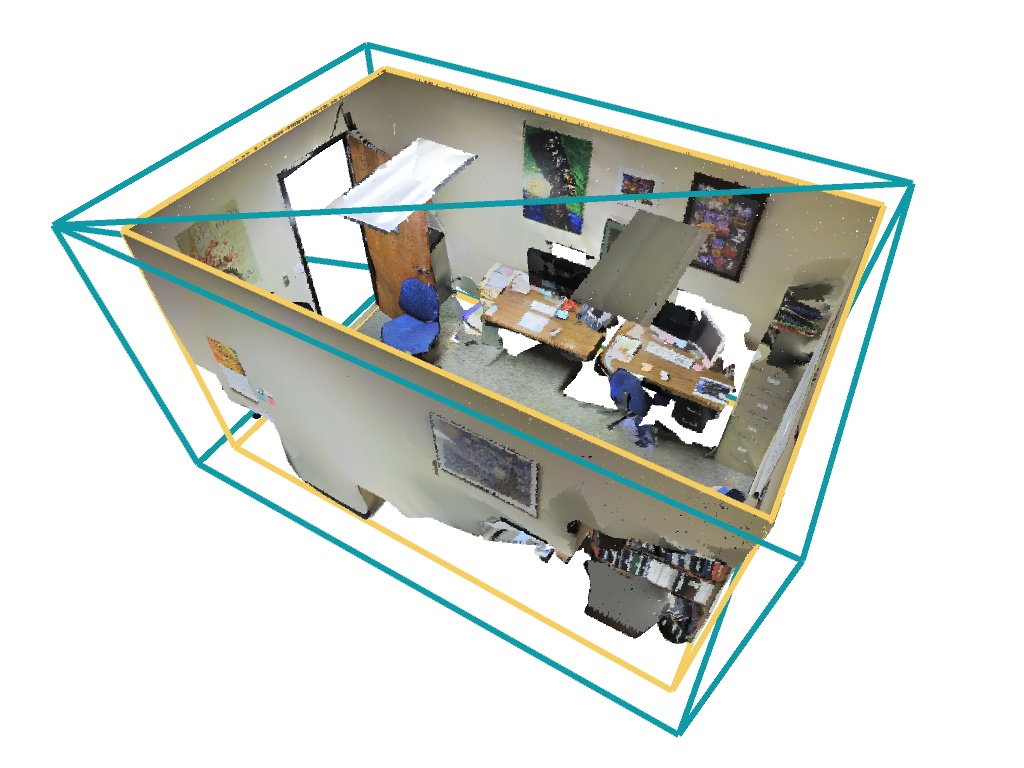}};
        \node at (5.6,0.0) {\includegraphics[scale=0.13]{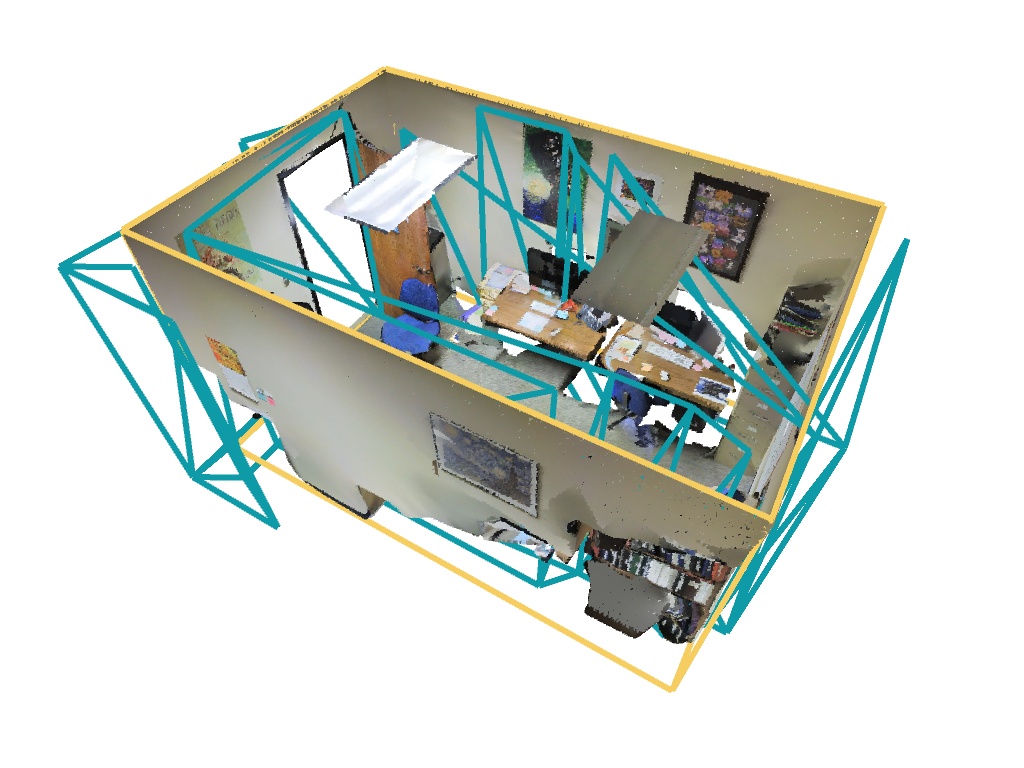}};
        \node at (11.2,0.0) {\includegraphics[scale=0.13]{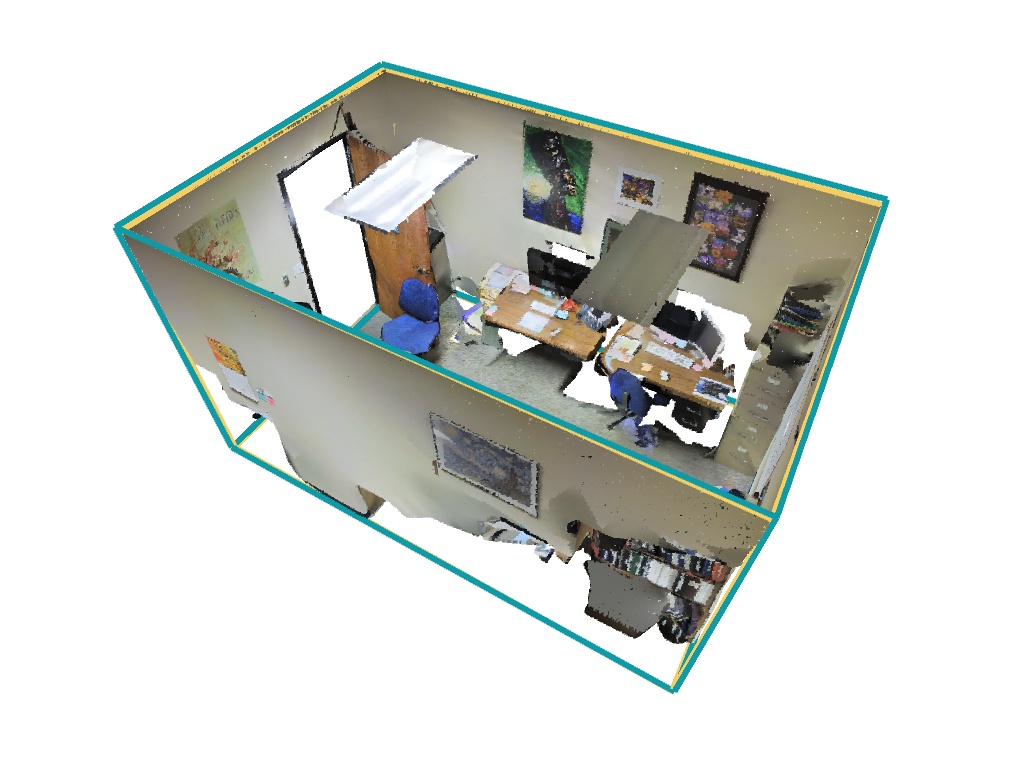}};

        \node[align=center,rotate=90] at (-2.8,3.8) {Total3D};
        \node[align=center,rotate=90] at (2.8,3.8) {Implicit3D};
        \node[align=center,rotate=90] at (8.4,3.8) {Deep3DLayout};
        \node[align=center,rotate=90] at (-2.8,0.0) {LED\textsuperscript{2}-Net};
        \node[align=center,rotate=90] at (2.8,0.0) {PSMNet};
        \node[align=center,rotate=90] at (8.4,0.0) {\methodName{}};
    \end{tikzpicture}
    \caption{Qualitative comparisons of predicted room layouts for one space in 2D-3D-Semantics. The ground truth point cloud is shown for visualization purposes, but \textbf{the methods only have the RGB images as input}. Predictions are shown in \textcolor{PredColor}{\bf blue} and the ground truth cuboids in \textcolor{GTColor}{\bf yellow}. None of the methods are trained on this dataset. For more examples see our supplementary material.
    }
    \label{fig:example-layout}
\end{figure*}

%% file: figs/failure_cases.tex
\begin{figure}
    \centering
    \begin{tikzpicture}[scale=1.0]
        \node at (0.0,0.0) {\includegraphics[scale=0.1]{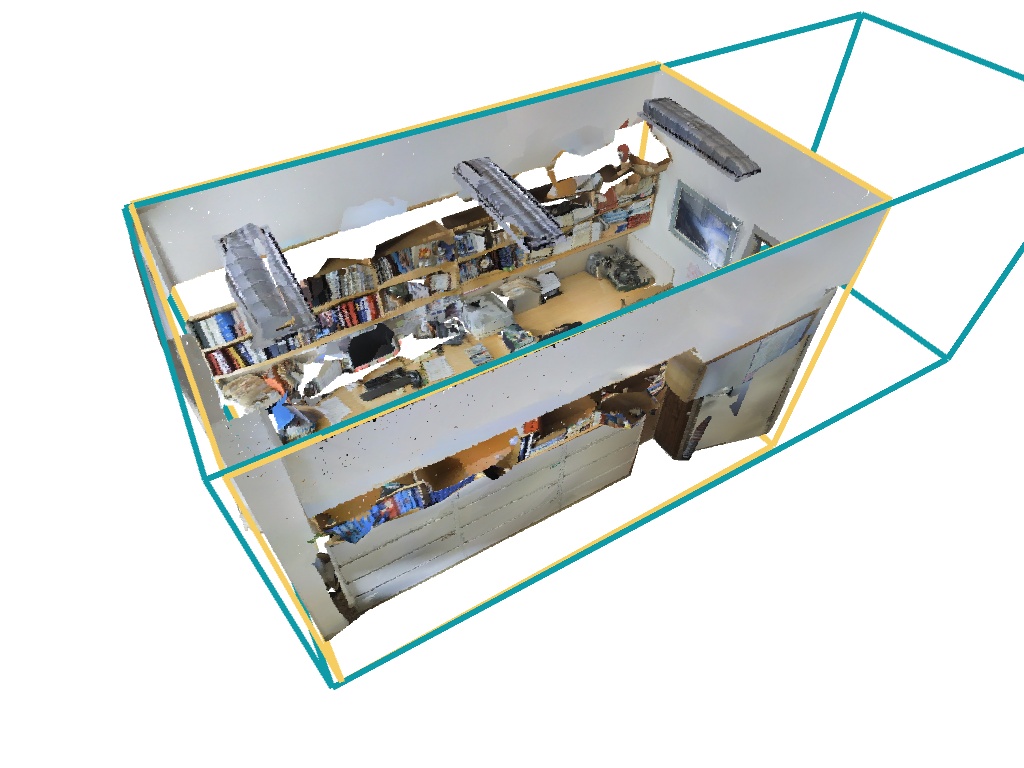}};
        \node at (3.75,0.0) {\includegraphics[scale=0.1]{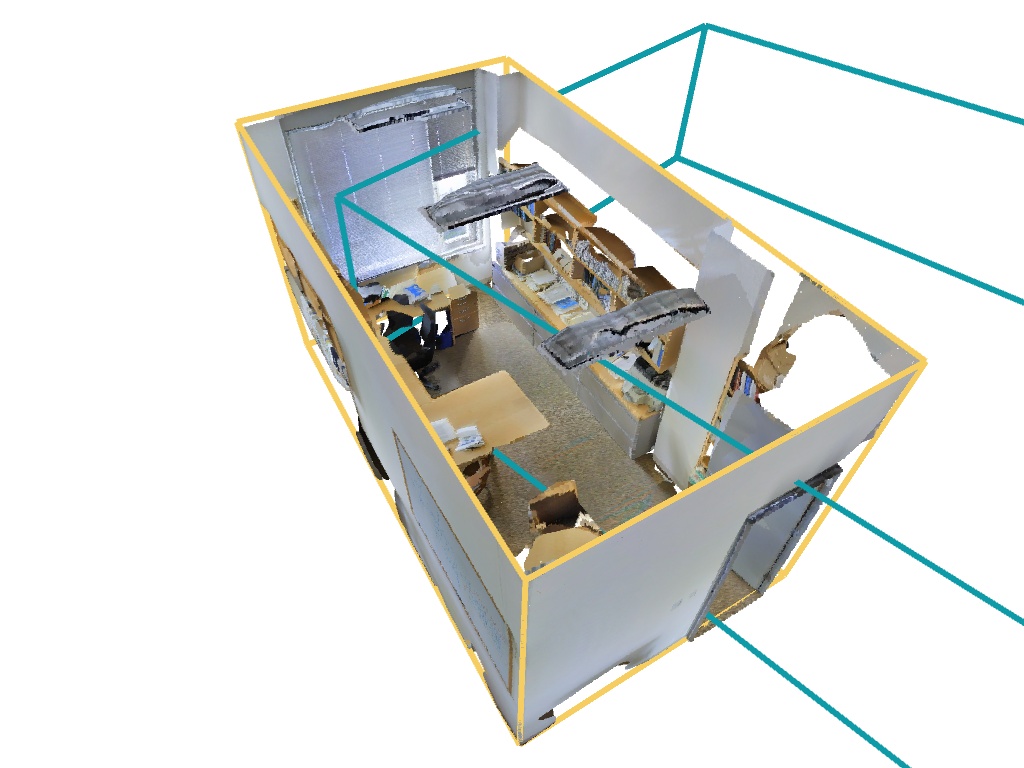}};
    \end{tikzpicture}
    \caption{Failure cases of \methodName{} on 2D-3D-Semantics. Predicted room layouts are shown in \textcolor{PredColor}{\bf blue} and ground truth cuboids in \textcolor{GTColor}{\bf yellow}. See our supplementary material for more examples.
    }
    \label{fig:failure-cases}
\end{figure}

%% file: tables/ablations_small.tex
\begin{table}
\centering
\resizebox{\columnwidth}{!}{
\begin{tabular}{c l c c c}
  \toprule
  && IoU $\uparrow$ & Rot. $\downarrow$ & Success $\uparrow$ \\
  \midrule
  \multirow{3}{*}{\rotatebox{90}{Feat.}}
  & RGB & 24.2 & 36.4\degree & 1.4\% \\
  & PixLoc (CMU) & 25.3 & 33.4\degree & 1.1\% \\
  & \methodName{} {\footnotesize ($E_{feat}$ only)} & \textbf{35.2} & \textbf{23.0\degree} & \textbf{5.8\%} \\
  \midrule
  \multirow{6}{*}{\rotatebox{90}{Cost}}
  & $E_{feat}$ & 35.2 & 23.0\degree & 5.8\% \\
  & $E_{edge}$ & 76.1 & 4.7\degree & 43.6\% \\
  & $E_{feat}$ + $E_{edge}$ & 81.9 & 3.8\degree & 61.4\% \\
  & $E_{feat}$ + $E_{VP}$ & 44.6 & 1.5\degree & 21.7\% \\
  & $E_{edge}$ + $E_{VP}$ & 83.1 & \textbf{1.3\degree} & 51.9\% \\
  & $E_{feat}$ + $E_{edge}$ + $E_{VP}$ & \textbf{87.2} & \textbf{1.3\degree} & \textbf{67.2\%} \\
  \midrule
  \multirow{3}{*}{\rotatebox{90}{Sampl.}}
  & Random & 86.9 & 1.4\degree & 65.8\% \\
  & Floor/wall/ceiling & 86.6 & 1.4\degree & 66.4\% \\
  & Guided & \textbf{87.2} & \textbf{1.3\degree} & \textbf{67.2\%} \\
  \midrule
  \multirow{3}{*}{\rotatebox{90}{Res.}}
  & Low (256 px) & 84.2 & \textbf{1.3\degree} & 63.3\% \\
  & Medium (512 px) & \textbf{87.2} & \textbf{1.3\degree} & \textbf{67.2\%} \\
  & High (768 px) & 85.7 & \textbf{1.3\degree} & 66.1\% \\
  \midrule
  \multirow{4}{*}{\rotatebox{90}{Init.}}
  & Random & 60.8 & 18.0\degree & 40.0\% \\
  & Random + VP & 72.2 & 10.9\degree & 51.9\% \\
  & Y down & 84.9 & 2.1\degree & 64.7\% \\
  & Y down + VP & \textbf{87.2} & \textbf{1.3\degree} & \textbf{67.2\%} \\
  \midrule
  \multirow{3}{*}{\rotatebox{90}{Scales}}
  & Coarse & 81.0 & 1.4\degree & 36.7\% \\
  & Coarse + medium & 86.5 & 1.4\degree & 65.6\% \\
  & Coarse + medium + fine & \textbf{87.2} & \textbf{1.3\degree} & \textbf{67.2\%} \\
  \bottomrule
\end{tabular}
}
\caption{Ablation experiments on our ScanNet++ v2 test set. The full set of metrics is available in the supplementary material.
}
\label{tab:ablation-study}
\end{table}

%% file: figs/cuboid_init.tex
\begin{figure}
    \centering
    \begin{tikzpicture}[scale=1.0]
        \node at (0.0,0.0) {\includegraphics[scale=0.1]{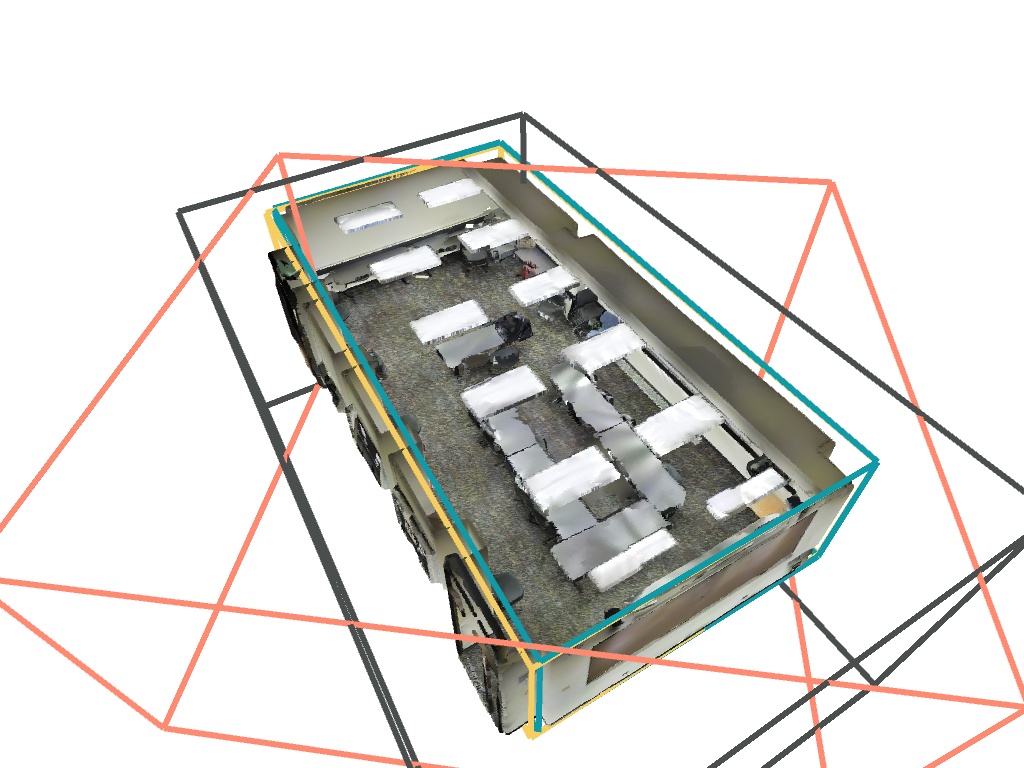}};
        \node at (3.75,0.0) {\includegraphics[scale=0.1]{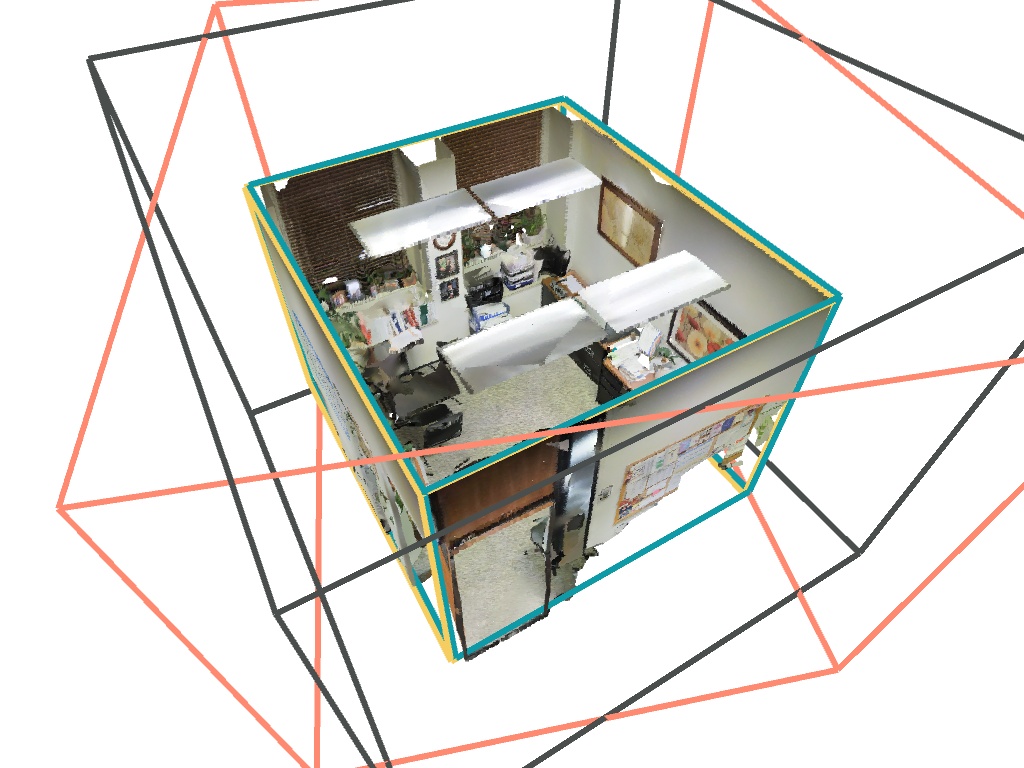}};
    \end{tikzpicture}
    \caption{Cuboid initialization examples on 2D-3D-Semantics. Even from poor initial estimates (\textcolor{InitColor}{\bf red}), \methodName{} can align the cuboids with the help of vanishing points (\textcolor{InitVPColor}{\bf gray}) and is able to converge to accurate layouts (\textcolor{PredColor}{\bf blue}). The ground truth is shown in \textcolor{GTColor}{\bf yellow}. See our supplementary material for more examples.
    }
    \label{fig:cuboid-init}
\end{figure}

%% file: figs/num_images.tex
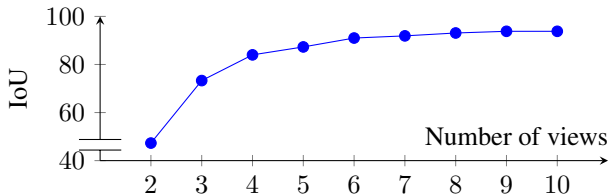
\begin{figure}
    \centering
    \begin{tikzpicture}
        \begin{axis}[
            width=\columnwidth,
            height=0.42\columnwidth,
            xlabel=Number of views,
            ylabel=IoU,
            xmin=1, xmax=11,
            ymin=40, ymax=100,
            xtick={2,3,...,10},
            axis lines = left,
            ytick={0,20,...,100},
            axis y discontinuity=parallel,
            x label style={
                at={(axis description cs:0.82,0.3)},
                anchor=north,
            },
        ]
            \addplot[mark=*,blue] plot coordinates {
            (2,47.3)
            (3,73.3)
            (4,84.0)
            (5,87.3)
            (6,91.0)
            (7,91.9)
            (8,93.1)
            (9,93.8)
            (10,93.8)
            };
        \end{axis}
    \end{tikzpicture}
    \caption{IoU as a function of the number of input views. Results on our ScanNet++ v2 test set.}
    \label{fig:num-images}
\end{figure}

%% file: figs/replica.tex
\begin{figure}[tp]
    \centering
    \includegraphics[width=\linewidth]{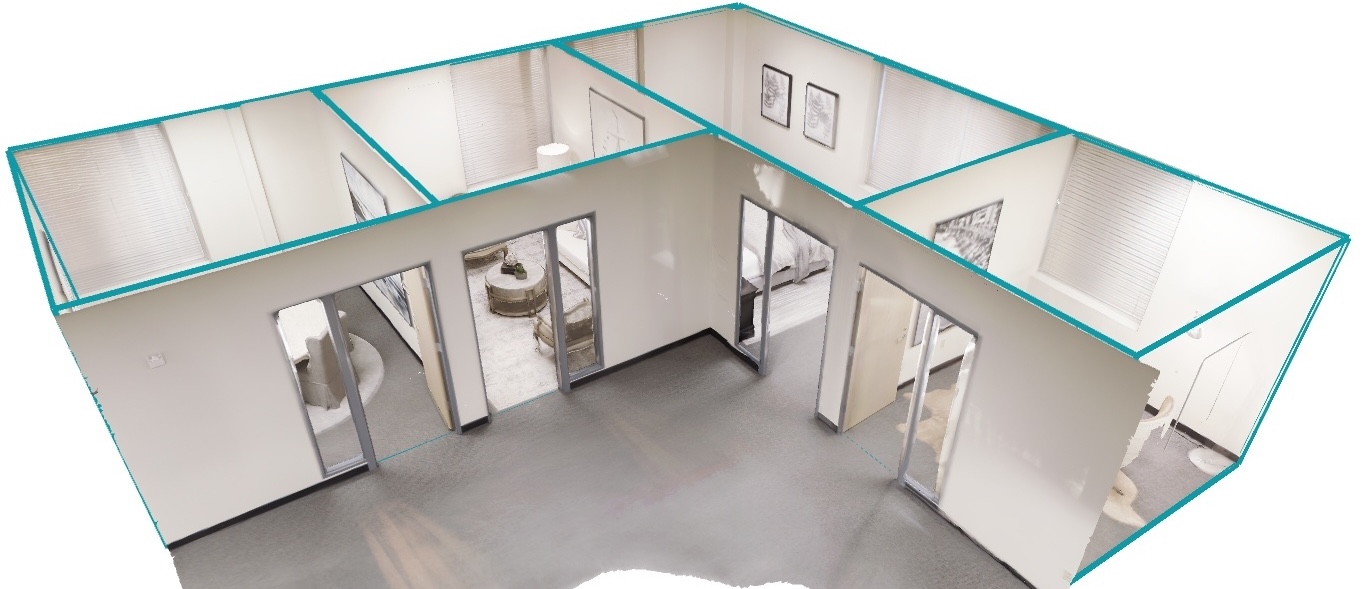}
    \caption{Multi-room layout estimation on Replica.} 
    \label{fig:replica}
\end{figure}

%% file: 10_conclusion.tex
\section{Conclusion}
\label{sec:conclusion}

We have presented \methodName{} which formulates the room layout estimation task as an optimization problem by aligning deep features across images.
As the network learns relatively low-level features (useful for pixel-wise direct alignment), our method is able to generalize to new datasets without retraining, in contrast to competing methods.
The flexibility of the optimization-based approach allow us to include an arbitrary number of images.
Further, other geometric constraints can easily be integrated, e.g.~optimizing multiple cuboids with the same orientation
or shared room height.
We have focused on indoor room estimation, but we believe the method can be generalized to outdoor images, i.e.~fitting cuboids or other geometric primitives to buildings.

{\small 
\noindent\textbf{Acknowledgments}
The work was supported by ELLIIT, 
the Swedish Research Council (Grant No. 2023-05424), and the Wallenberg AI, Autonomous Systems and
Software Program (WASP) funded by the Knut and Alice Wallenberg Foundation.
Compute was provided by the supercomputing resource Berzelius provided by National Supercomputer Centre at Linköping University and the Knut and Alice Wallenberg foundation.
}

%% file: 12_appendix.tex
\section{Training Details}

In this section we provide a more thorough description of how \methodName{}'s neural network is trained.
We employ a two-stage training procedure where first the edge maps $\EE_i$ are pre-trained with a weighted MSE loss (summed over all scale levels), using line renderings of the ground truth cuboids as target images. Pixels that lie on the cuboid edges are up-weighted by a factor of five. The input images are resized to 512 pixels in height while maintaining the aspect ratio. The ResNet-101 \cite{he2016deep} encoder is initialized with weights trained on ImageNet-1K \cite{deng2009imagenet} and is not frozen during this first training stage. In the first training stage a batch size of 10 is used and 150 examples are randomly sampled from each training scene at every epoch.

Next, the full network is trained with the loss in Eq. \eqref{eq:loss}, applied at each scale and summed. We define a success threshold for the loss at each scale level (48, 12 and 3 pixels, respectively) and if the optimization is not successful the loss on the subsequent level is zeroed out (by setting $\gamma_m=0$ or $\gamma_f=0$, otherwise $\gamma_m = \gamma_f = 1$), to prevent learning from examples that are too difficult. Random horizontal cropping is performed after resizing, resulting in images of size 512x512.
The 2D-3D correspondences $\{ (\bm{x}_{ik}^{GT}, \bm{X}_{ik}^{GT}) \}$ are found by taking the vertices of the semantic mesh labeled "floor", "wall" or "ceiling" that are visible in each particular view. From these we select 256 points randomly during training to compute the loss. As described in Sec. \ref{subsec:learning-to-optimize-room-layouts} we use guided point sampling for the image points $\{ \xx_{ik} \}$. 256 points are sampled, individually on each scale level with $\gamma = 4$.
We set $\beta = 0$ during training since the vanishing point cost does not include any learned component, and let $\alpha = 0.1$. 40 points are sampled on each cuboid edge to compute the edge cost. We run three iterations of LM optimization on each scale level. Cuboids are initialized from the ground truth by applying a random rotation in the $[0\degree, 15\degree]$ range, followed by translation of up to 0.5 m in each direction and a resizing of the sides between $[-1, 1.5]$ m.
The learned damping parameters of the LM optimization are handled like in \cite{sarlin2021back}. Feature maps $\FF_i$ have dimension 128, 128 and 32 on the coarse, medium and fine scale levels, respectively. We use a batch size of 4 and sample 50 examples per training scene at each epoch in the second training stage. The network weights (30 million parameters) are saved at each epoch and we pick the ones that minimize the warp loss Eq. \eqref{eq:loss} on the validation set.

In both stages the Adam \cite{kingma2014adam} optimizer is used to train the network, with a learning rate of $5 \times 10^{-6}$. Gradients are clipped to the $[-1,1]$ range. The training is run for 10 epochs. The pre-training takes 13 h and the training of the full network finishes in 27 h, using a NVIDIA TITAN V GPU.

\section{Results}



In \cref{fig:example-layouts} we display additional examples of predicted room layouts in 2D-3D-Semantics. \cref{fig:failure-cases-extra,fig:cuboid-init-extra} present extra failure cases and cuboid initializations, respectively. Table \cref{tab:ablation-study-full} contains the full set of results for the ablation study in Sec. \ref{subsec:ablation-experiments}.

We also visualize the feature-, edge- and confidence maps (on the finest scale level) for the first five images of an image tuple in our ScanNet++ v2 test in \cref{fig:example-maps}. \methodName{} learns features that are consistent between views (second column) and ignores clutter by assigning high confidence to image points that lie on the floor, walls or ceiling (third column). It can often predict the cuboid edges accurately despite the presence of occluding objects (fourth column). 


\input{figs/example_layouts}

\input{figs/failure_cases_extra}

\input{figs/cuboid_init_extra}

\input{tables/ablations}

\input{figs/example_feat_edge_maps}

%% file: figs/example_layouts.tex
\begin{figure*}[tp]
    \centering
    \begin{tikzpicture}[scale=1.0]
        \node at (2.3,14.0) {\includegraphics[scale=0.1]{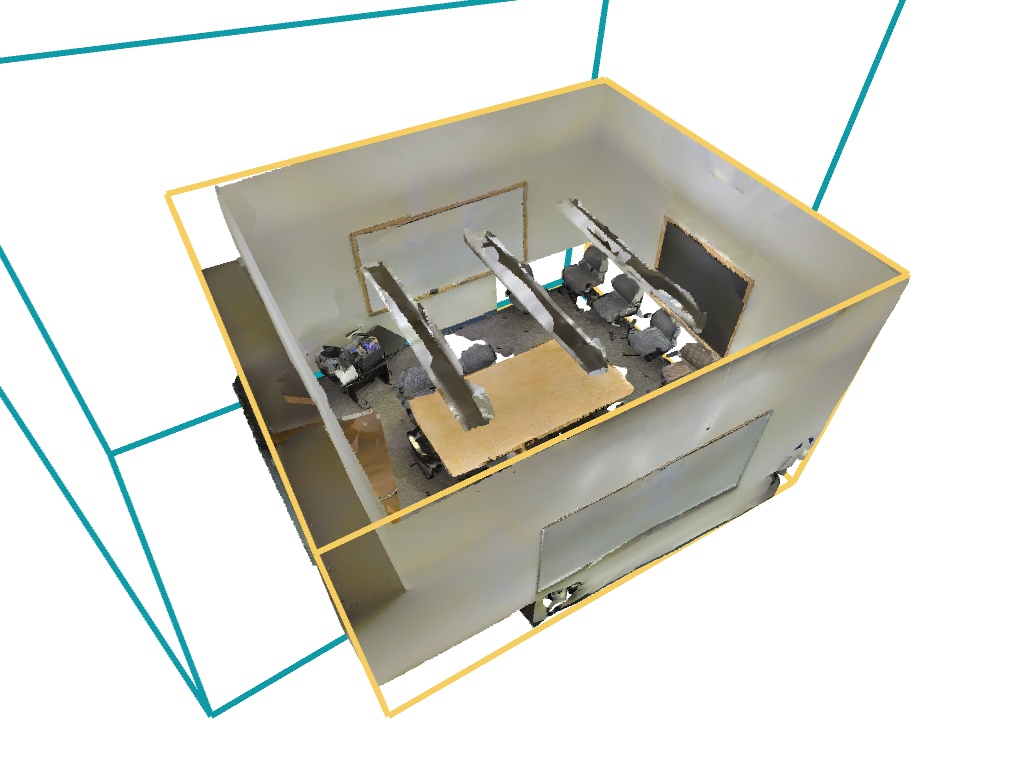}};
        \node at (2.3,11.2) {\includegraphics[scale=0.1]{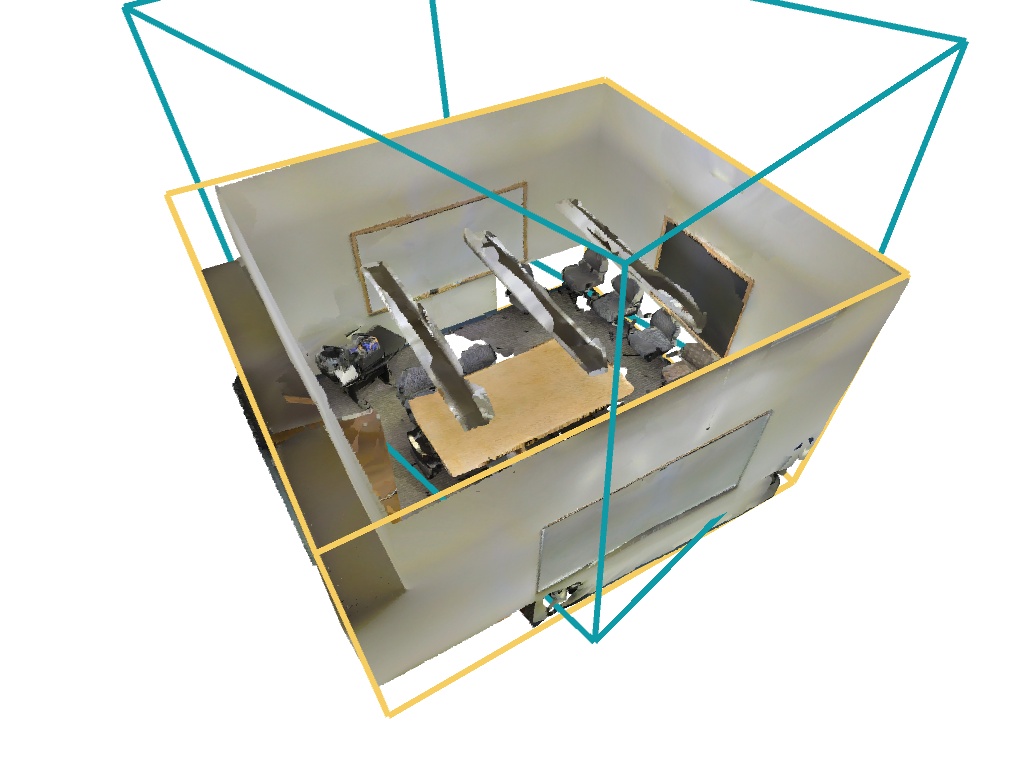}};
        \node at (2.3,8.4) {\includegraphics[scale=0.1]{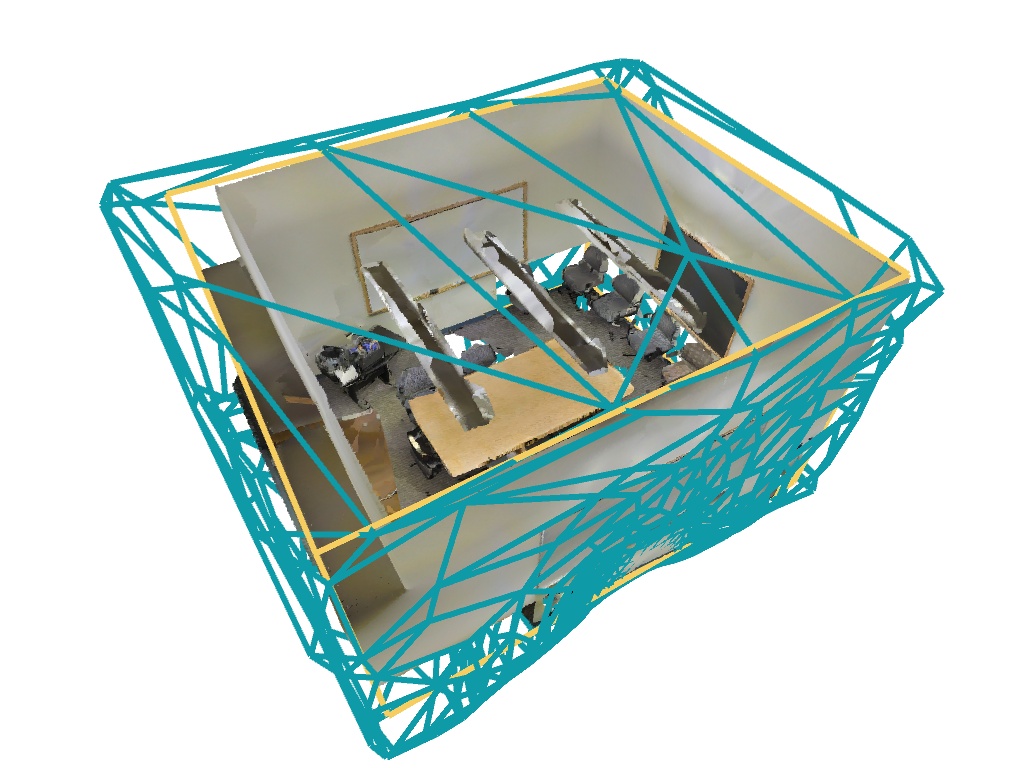}};
        \node at (2.3,5.6) {\includegraphics[scale=0.1]{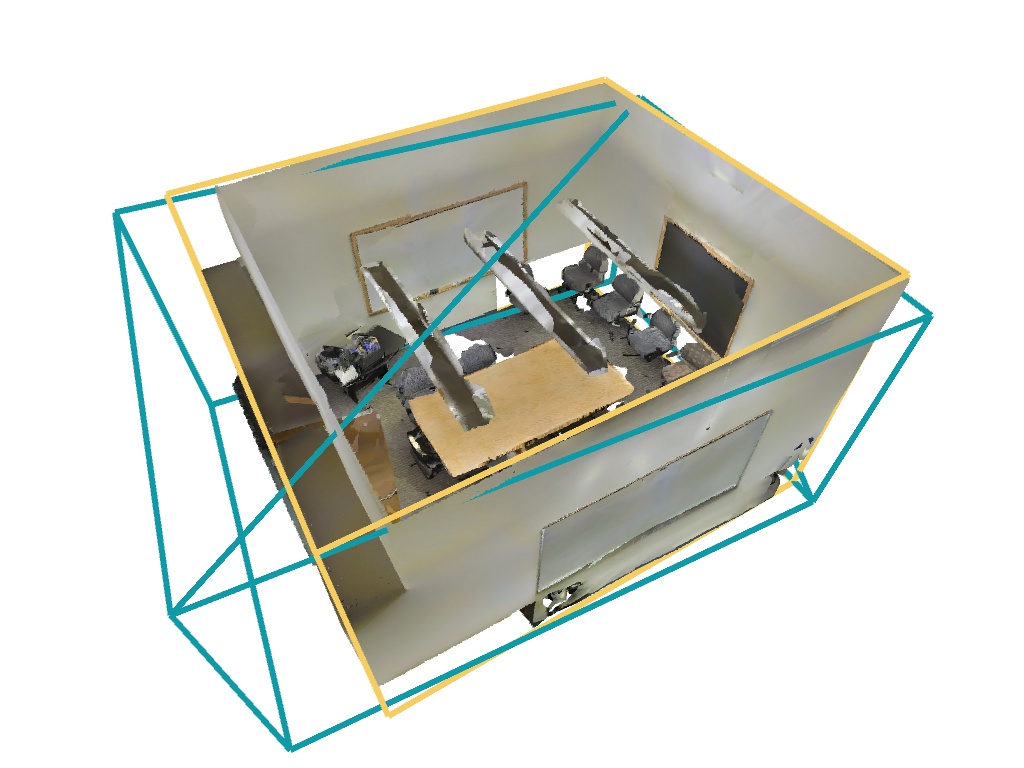}};
        \node at (2.3,2.8) {\includegraphics[scale=0.1]{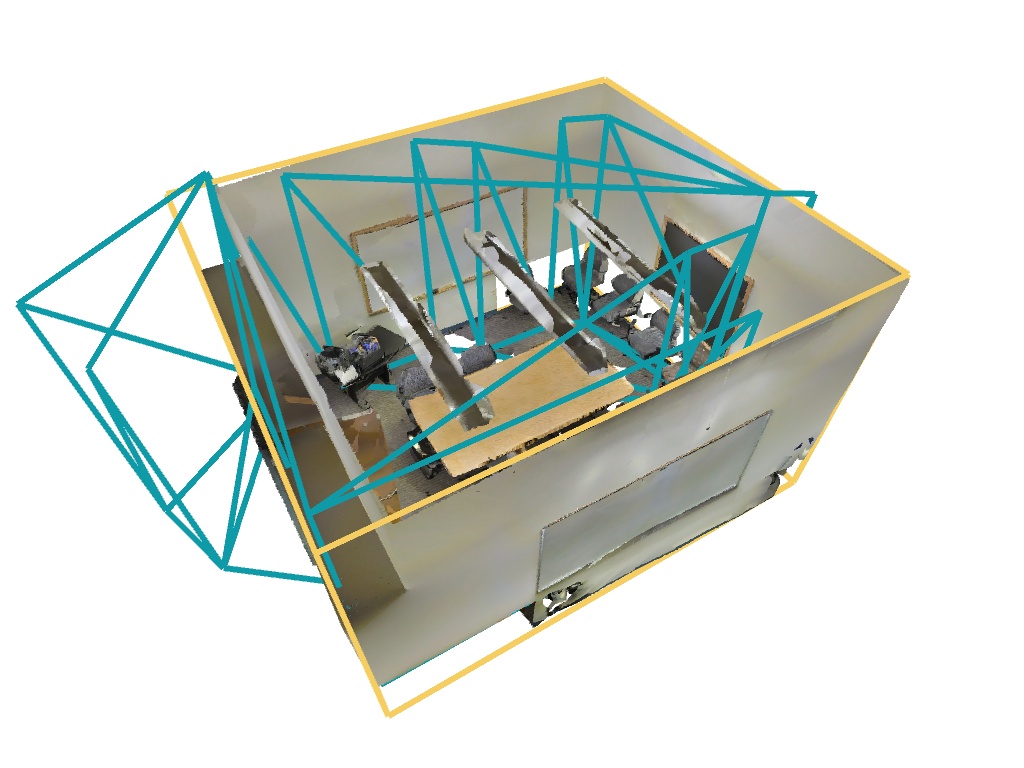}};
        \node at (2.3,0.0) {\includegraphics[scale=0.1]{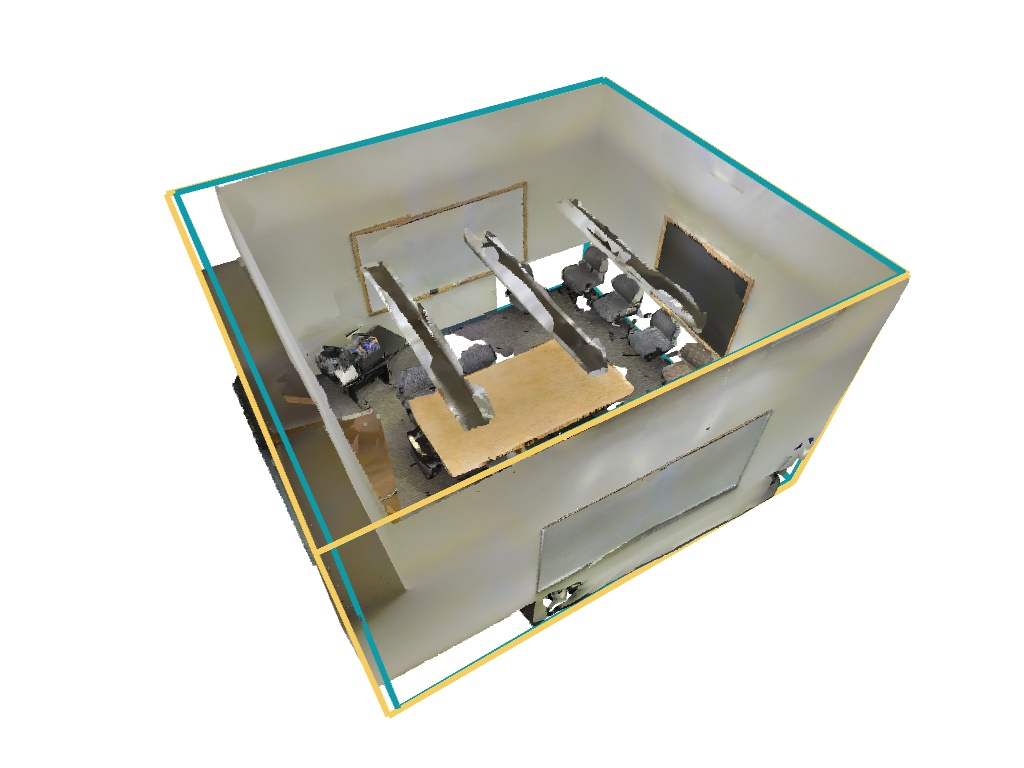}};

        \node at (6.0,14.0) {\includegraphics[scale=0.1]{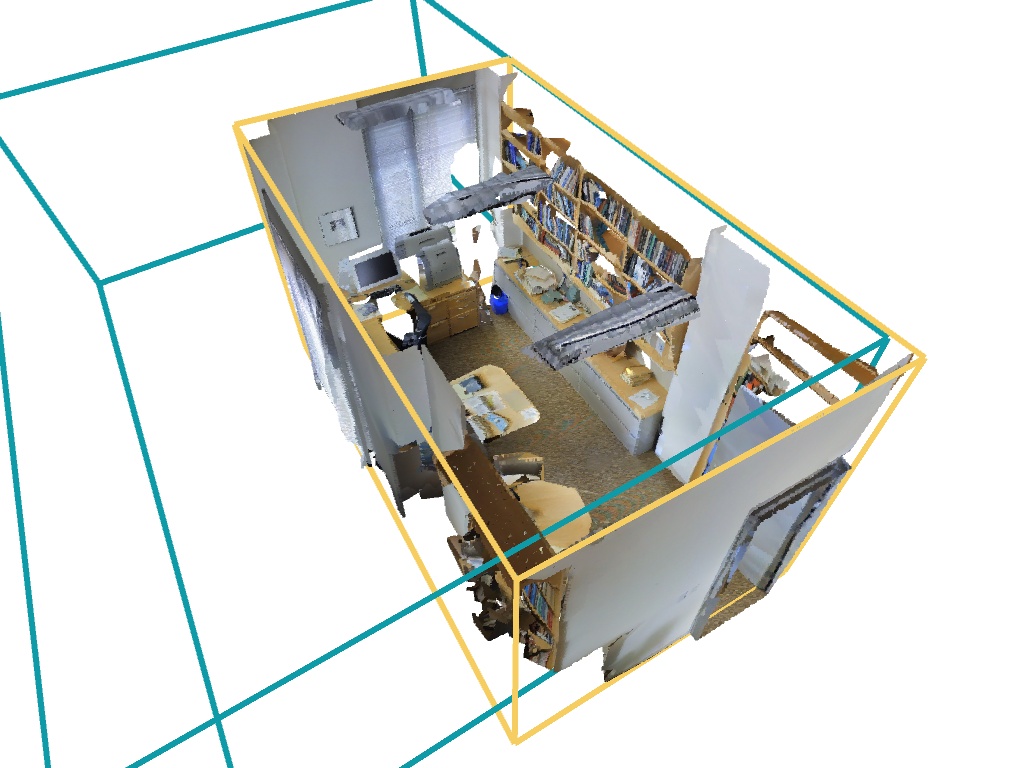}};
        \node at (6.0,11.2) {\includegraphics[scale=0.1]{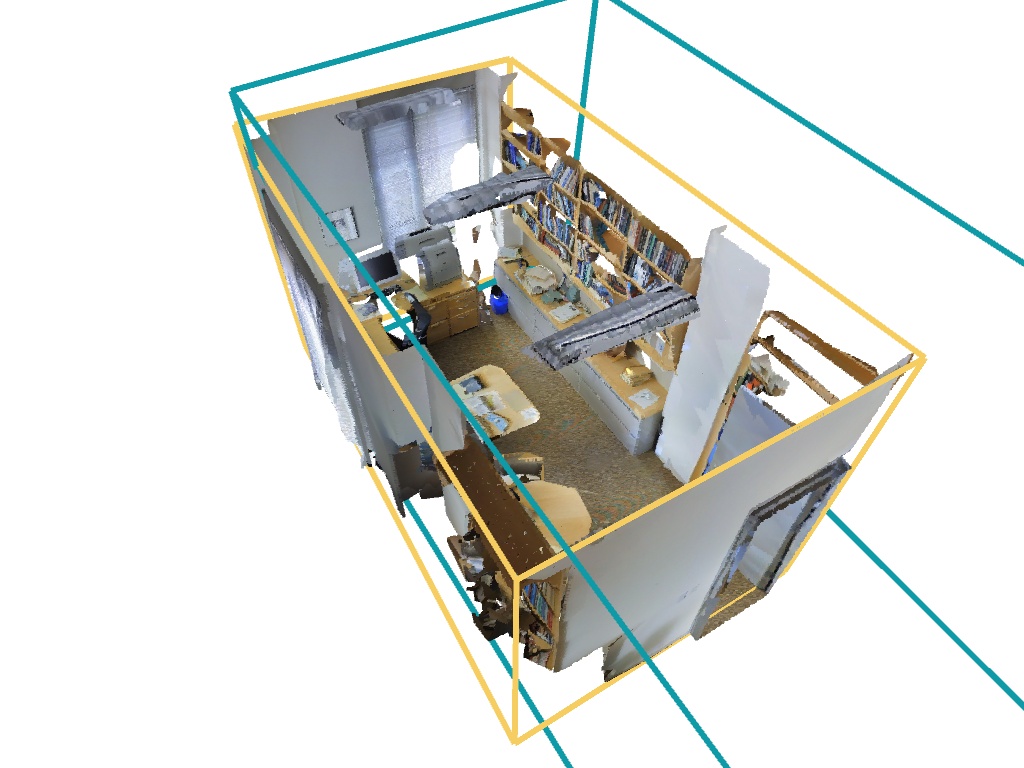}};
        \node at (6.0,8.4) {\includegraphics[scale=0.1]{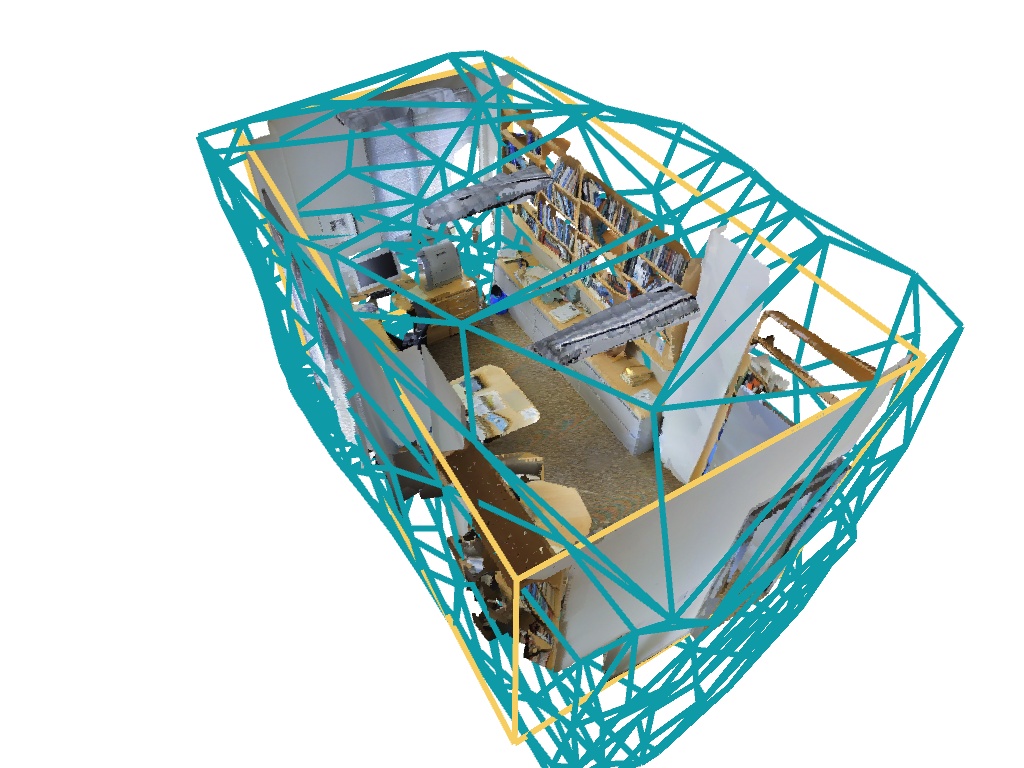}};
        \node at (6.0,5.6) {\includegraphics[scale=0.1]{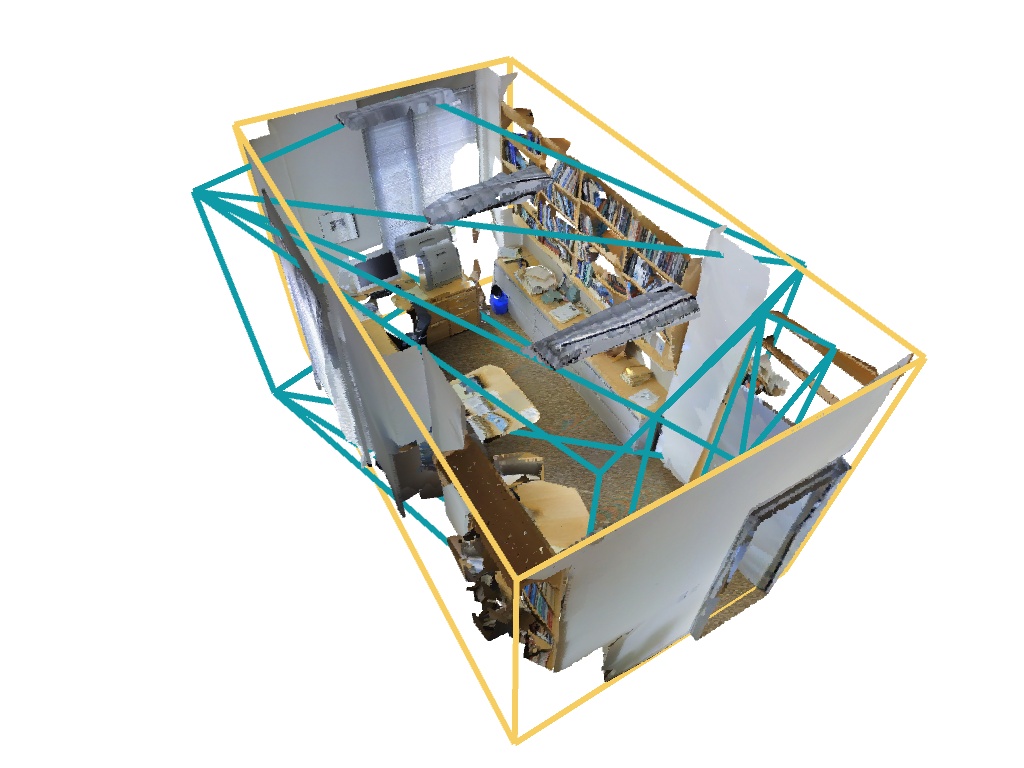}};
        \node at (6.0,2.8) {\includegraphics[scale=0.1]{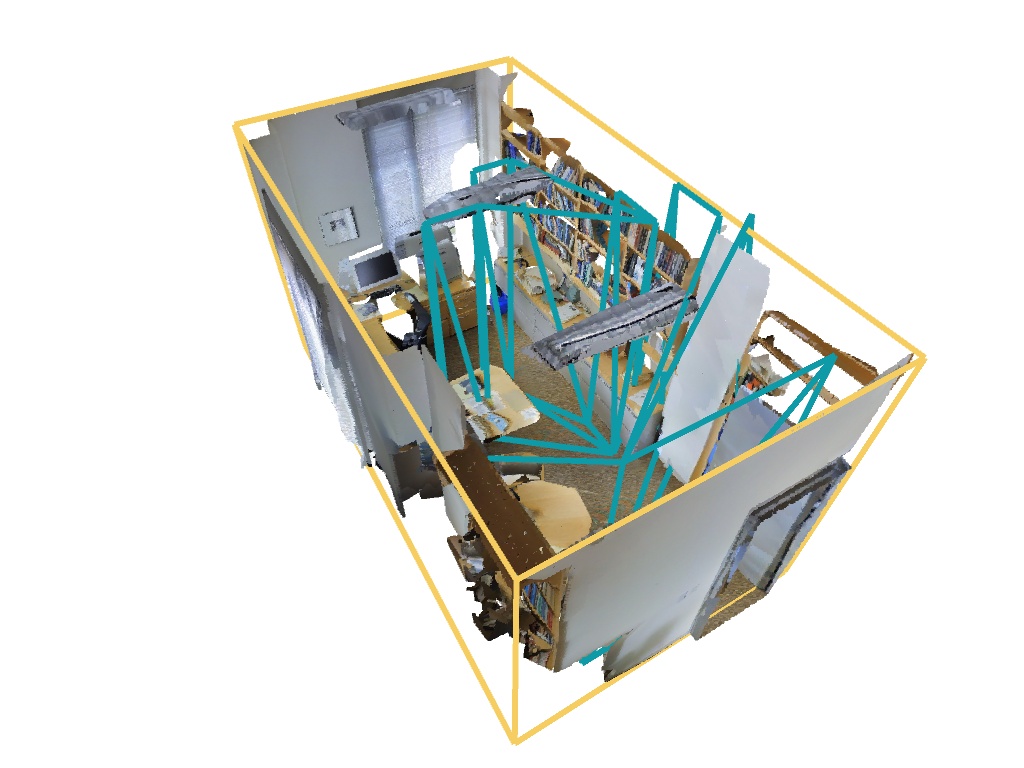}};
        \node at (6.0,0.0) {\includegraphics[scale=0.1]{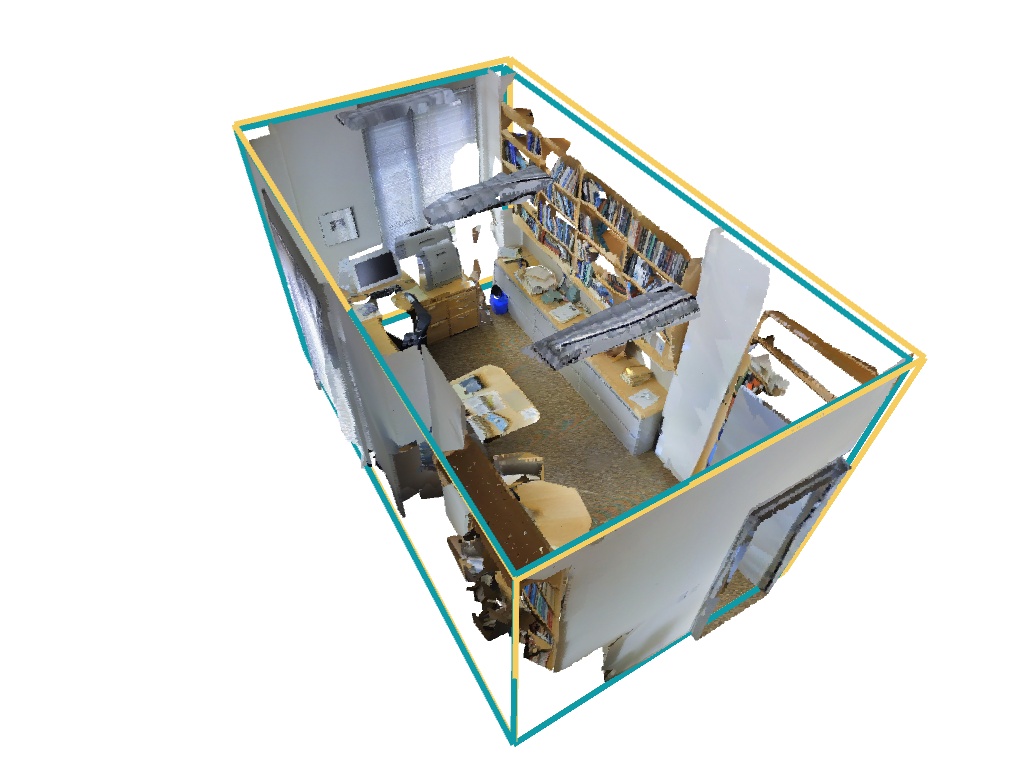}};

        \node at (9.7,14.0) {\includegraphics[scale=0.1]{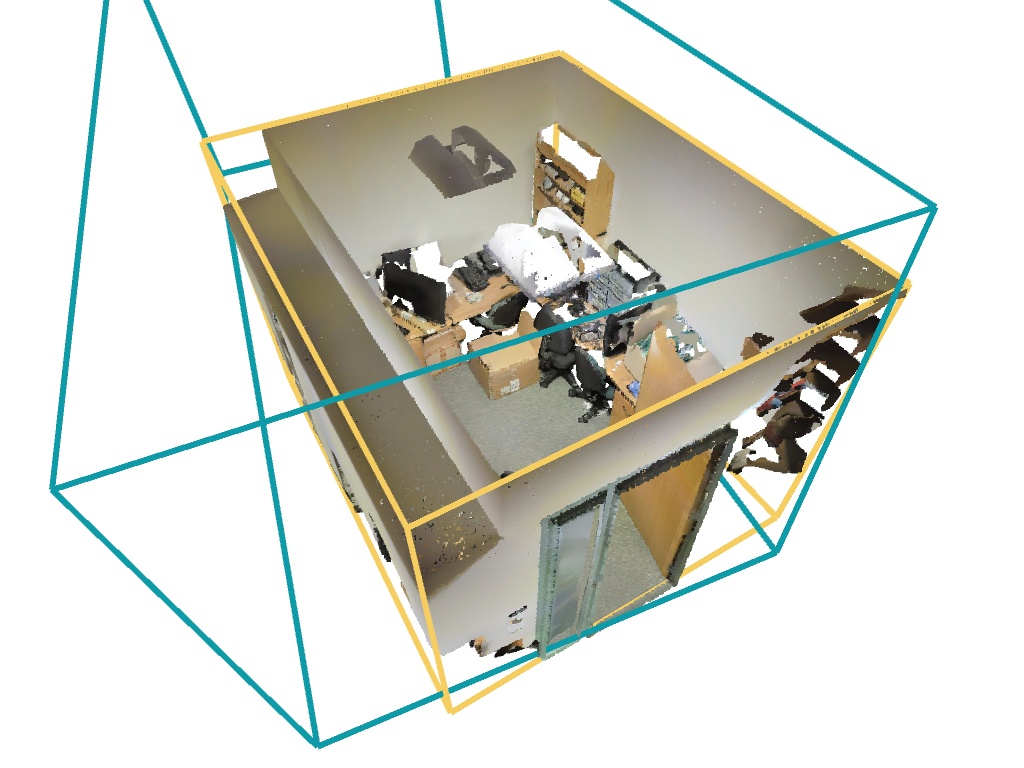}};
        \node at (9.7,11.2) {\includegraphics[scale=0.1]{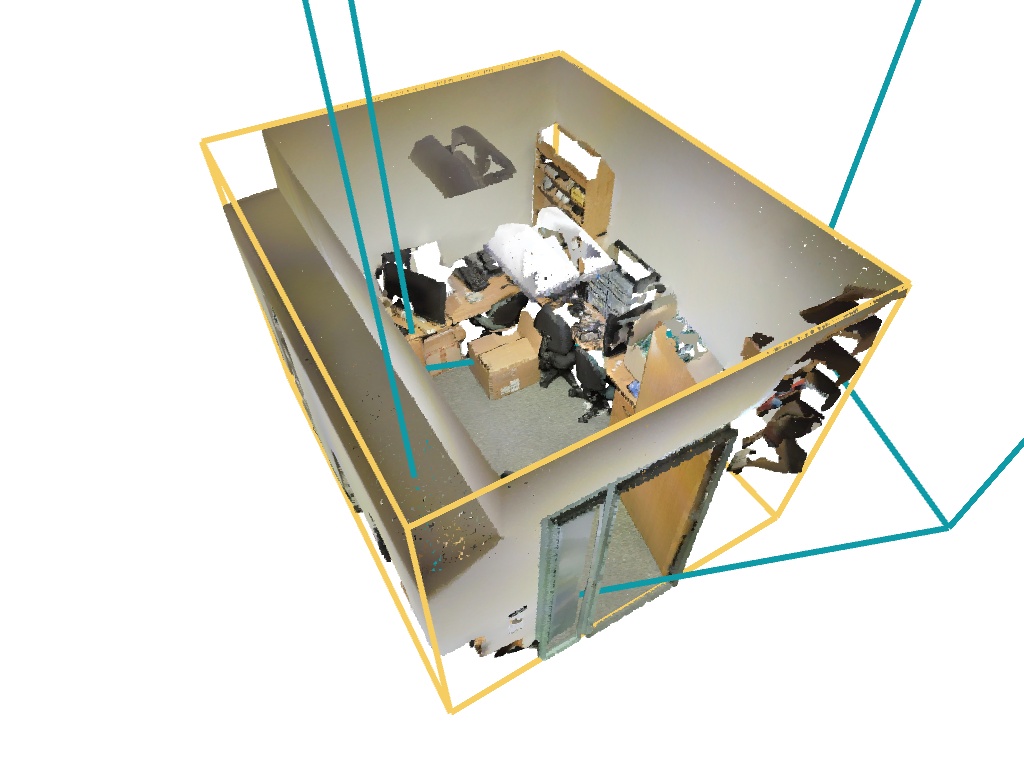}};
        \node at (9.7,8.4) {\includegraphics[scale=0.1]{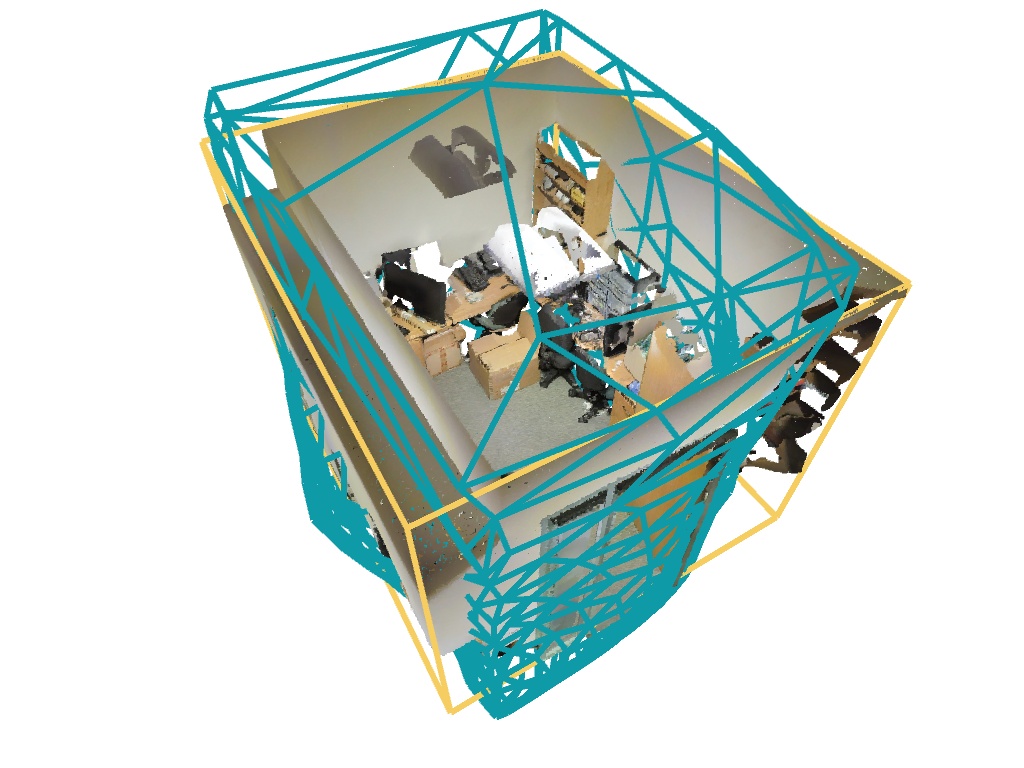}};
        \node at (9.7,5.6) {\includegraphics[scale=0.1]{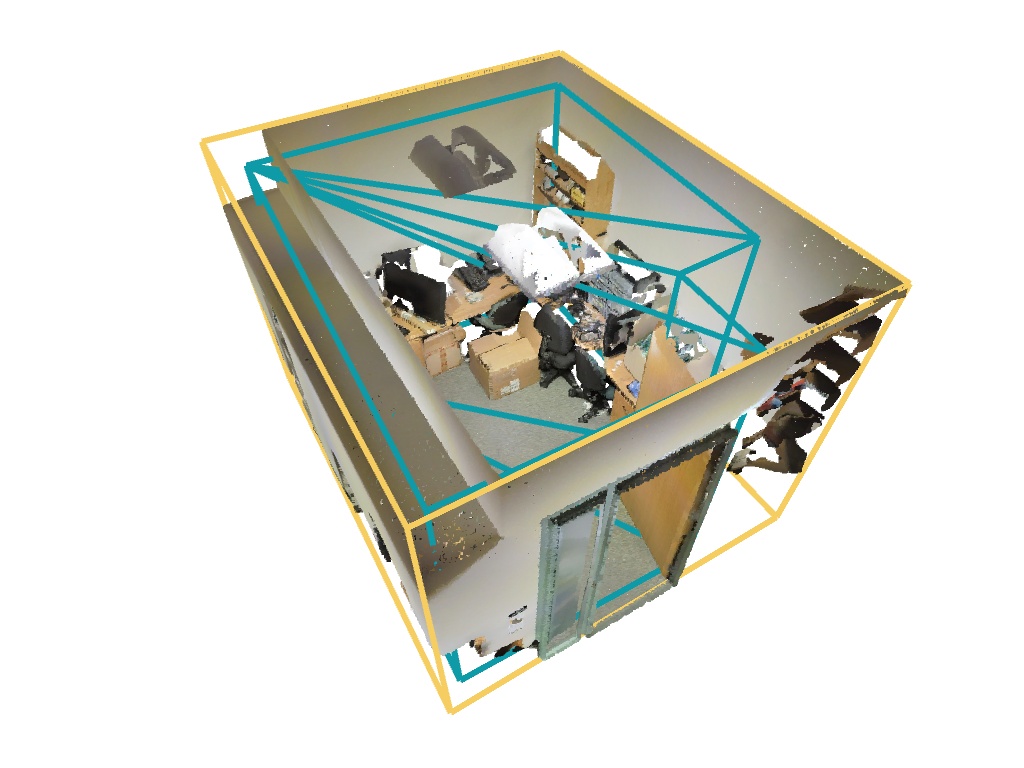}};
        \node at (9.7,2.8) {\includegraphics[scale=0.1]{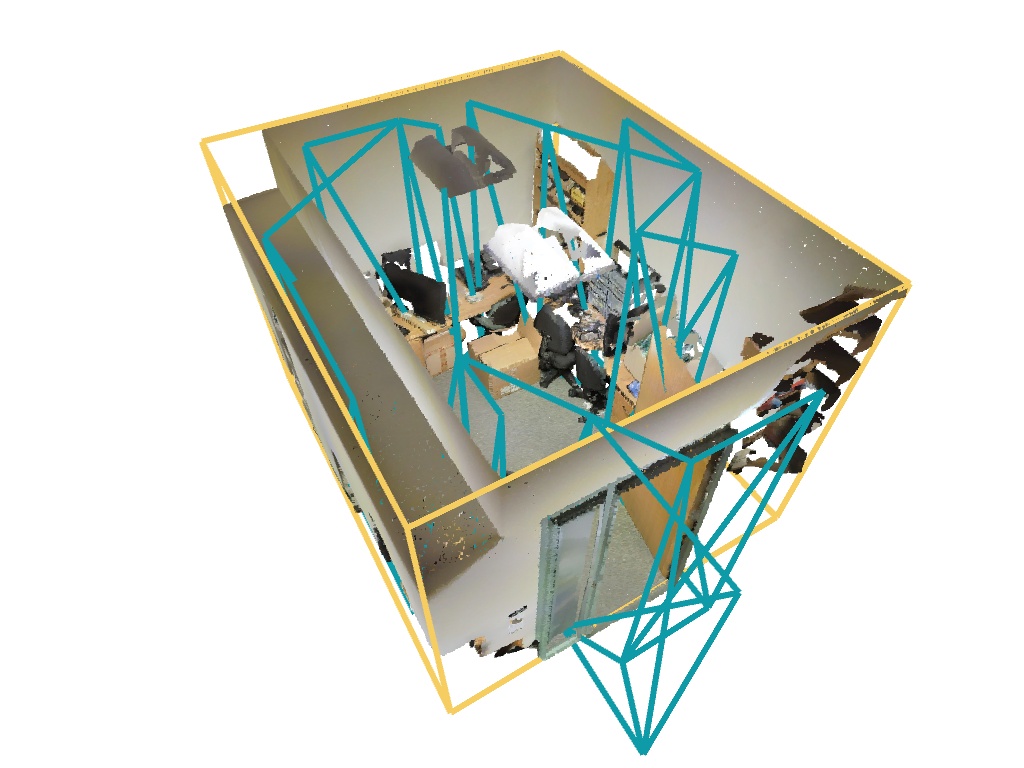}};
        \node at (9.7,0.0) {\includegraphics[scale=0.1]{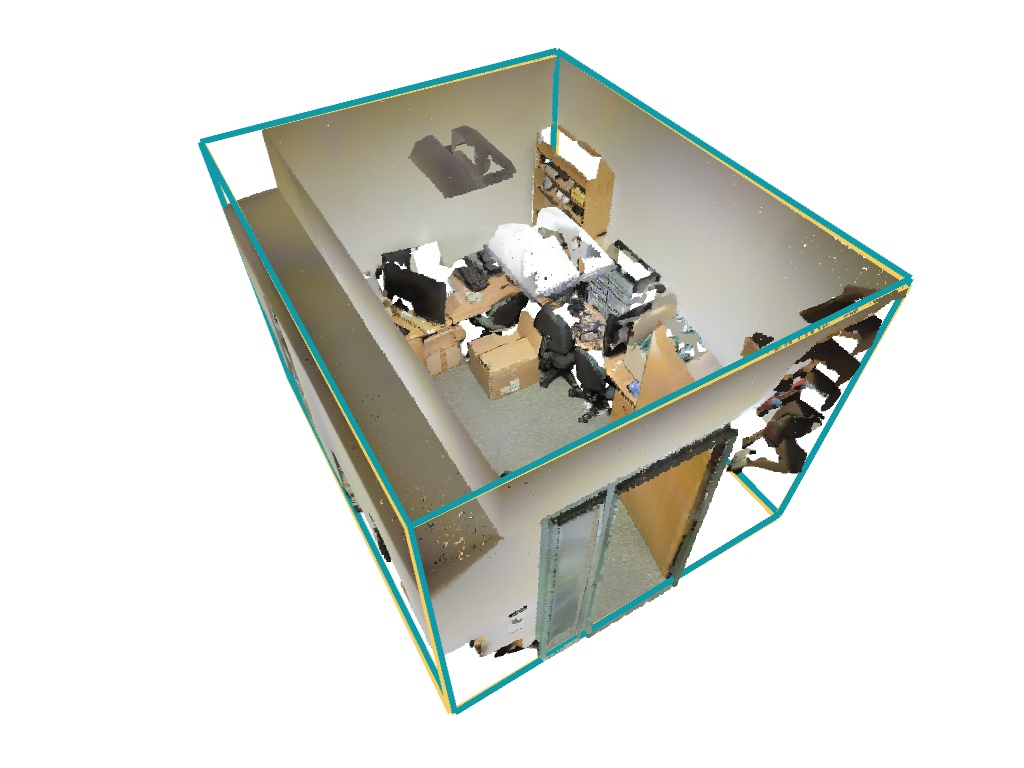}};

        \node at (13.4,14.0) {\includegraphics[scale=0.1]{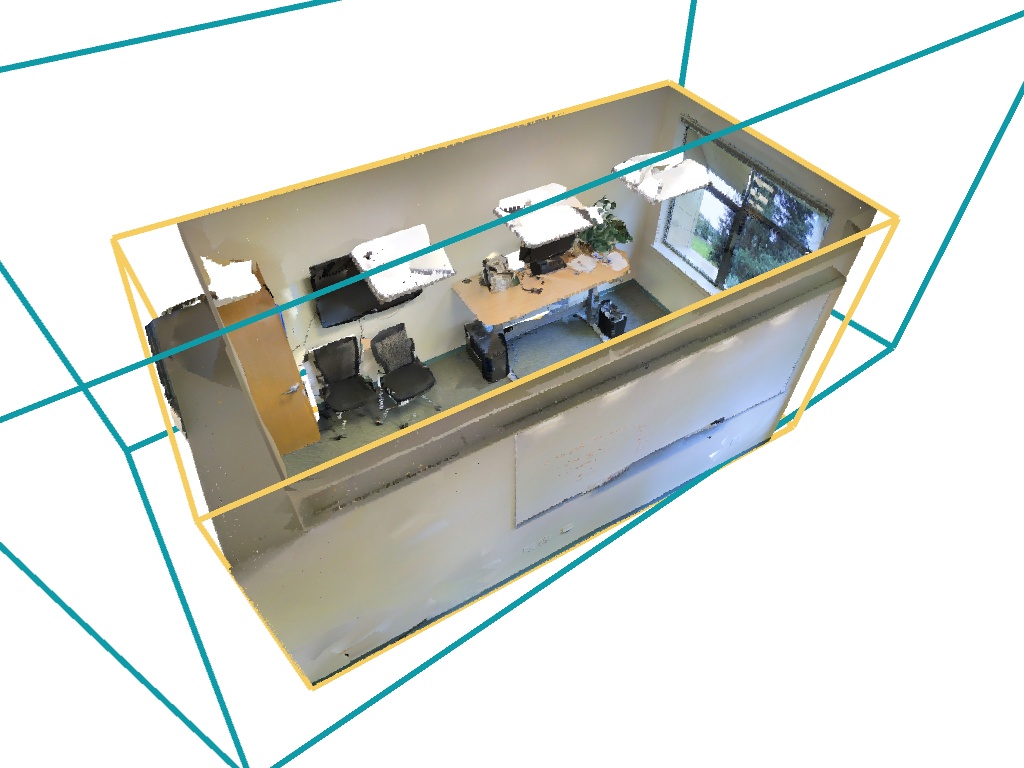}};
        \node at (13.4,11.2) {\includegraphics[scale=0.1]{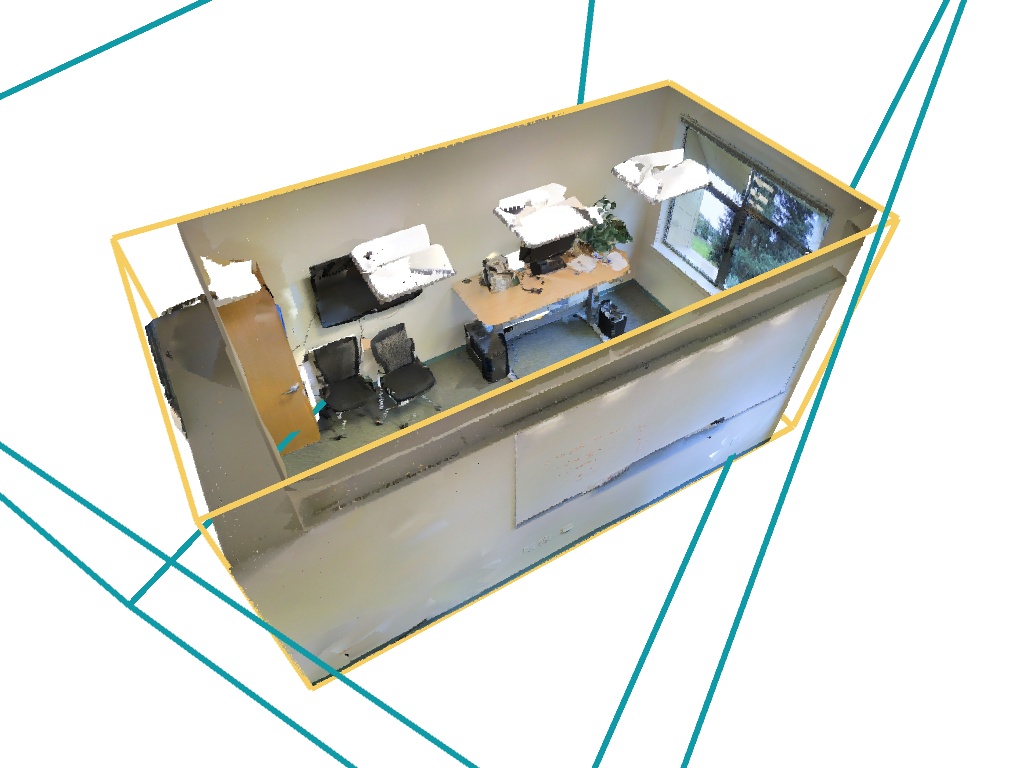}};
        \node at (13.4,8.4) {\includegraphics[scale=0.1]{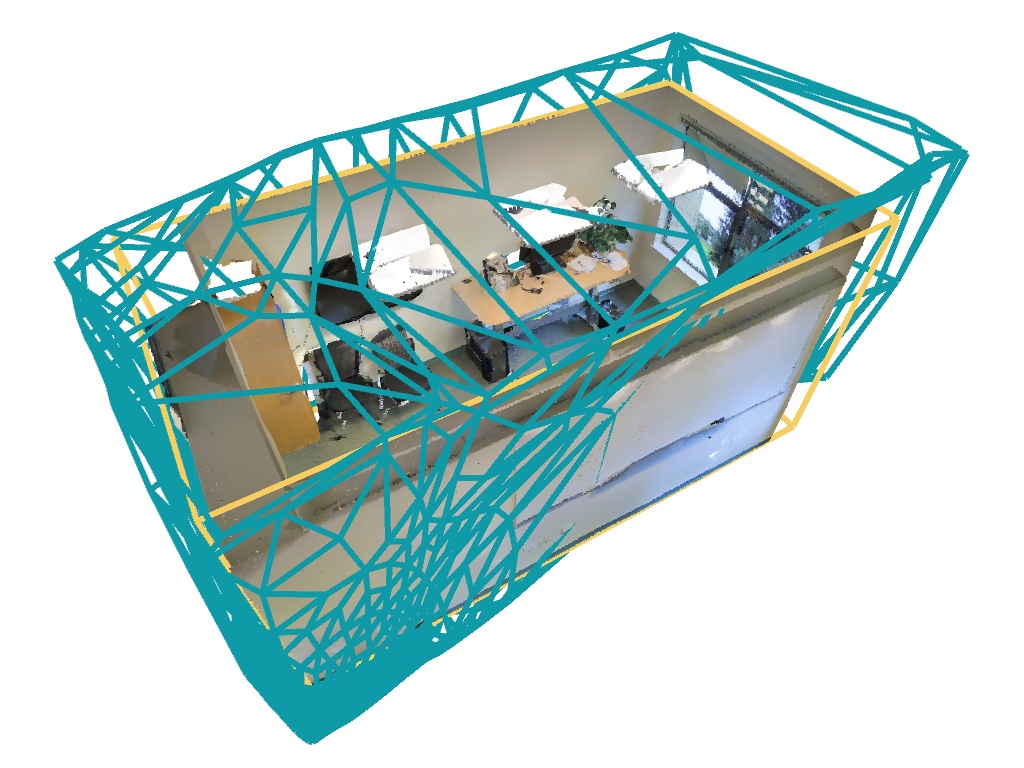}};
        \node at (13.4,5.6) {\includegraphics[scale=0.1]{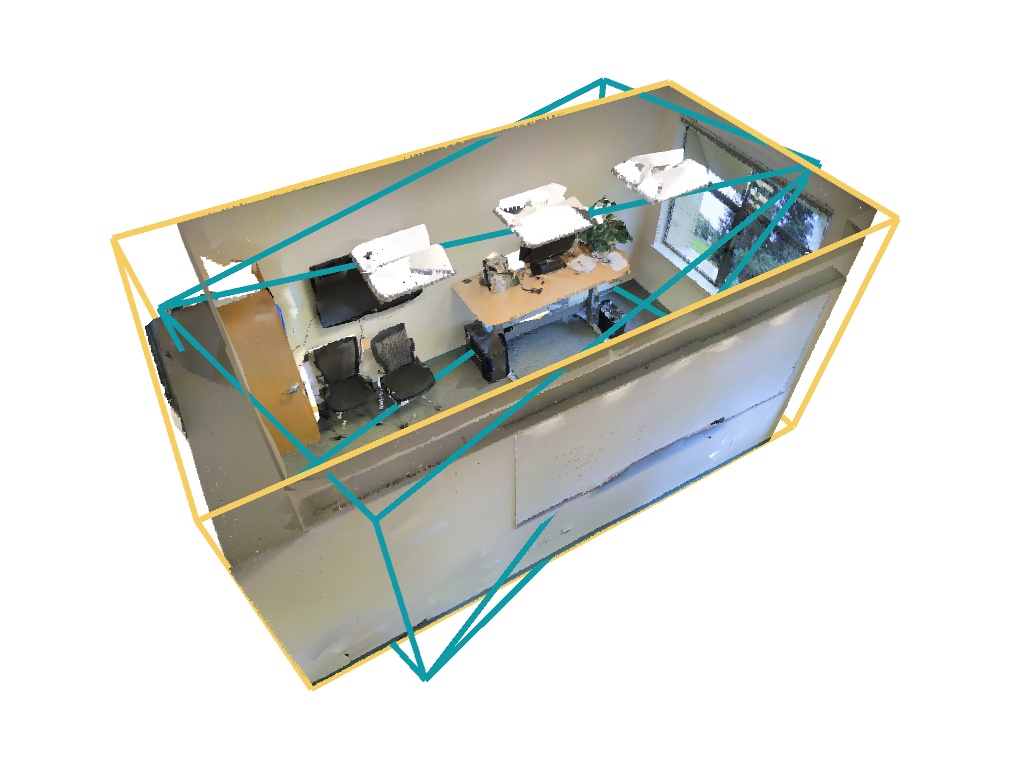}};
        \node at (13.4,2.8) {\includegraphics[scale=0.1]{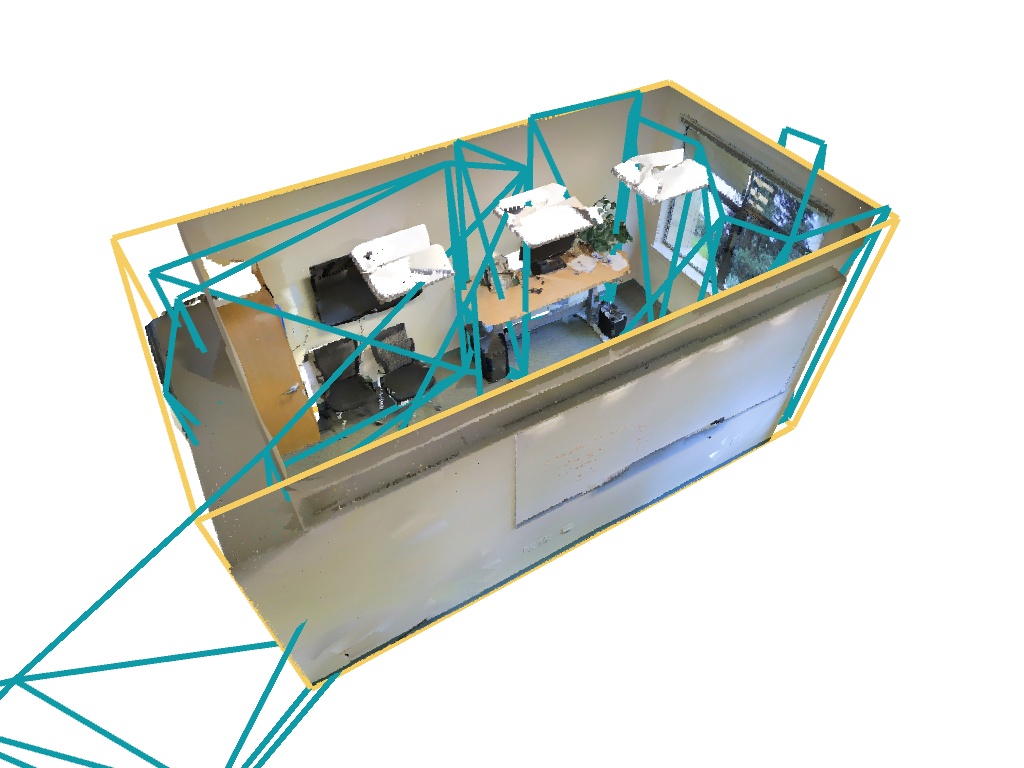}};
        \node at (13.4,0.0) {\includegraphics[scale=0.1]{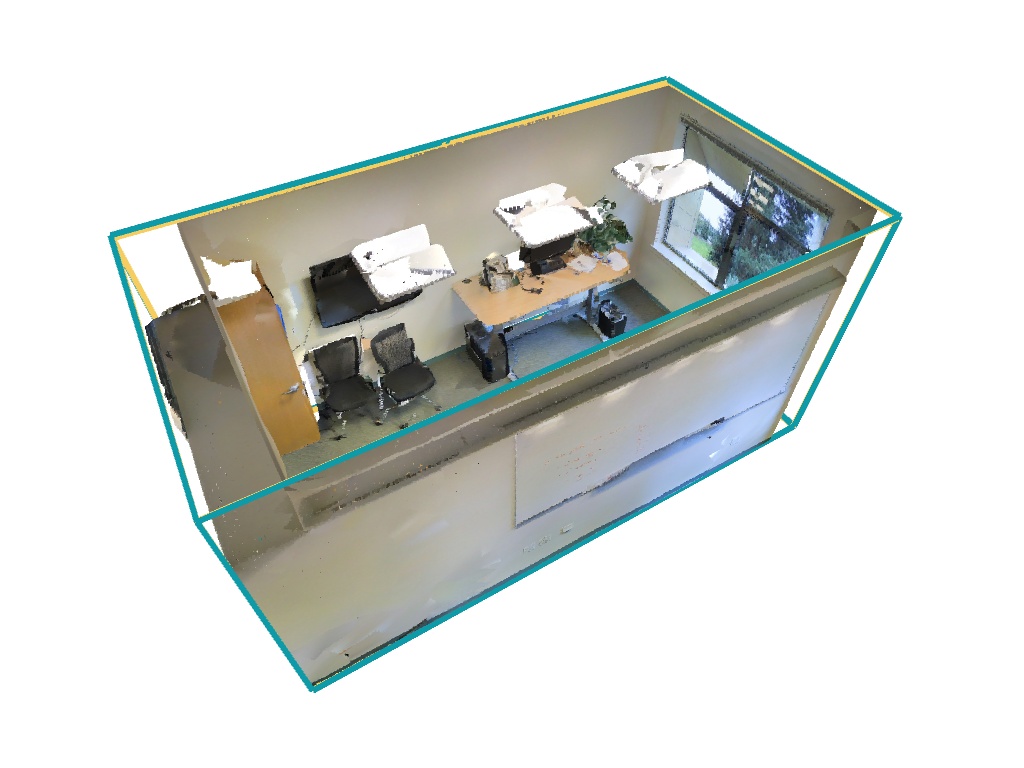}};

        \node[align=center,rotate=90] at (0.0,14.0) {Total3D};
        \node[align=center,rotate=90] at (0.0,11.2) {Implicit3D};
        \node[align=center,rotate=90] at (0.0,8.4) {Deep3DLayout};
        \node[align=center,rotate=90] at (0.0,5.6) {LED\textsuperscript{2}-Net};
        \node[align=center,rotate=90] at (0.0,2.8) {PSMNet};
        \node[align=center,rotate=90] at (0.0,0.0) {\methodName{}};
    \end{tikzpicture}
    \caption{Qualitative comparisons of predicted room layouts for spaces in 2D-3D-Semantics. Predictions are shown in \textcolor{PredColor}{blue} and the ground truth cuboids in \textcolor{GTColor}{yellow}. None of the methods are trained on this dataset.}
    \label{fig:example-layouts}
\end{figure*}
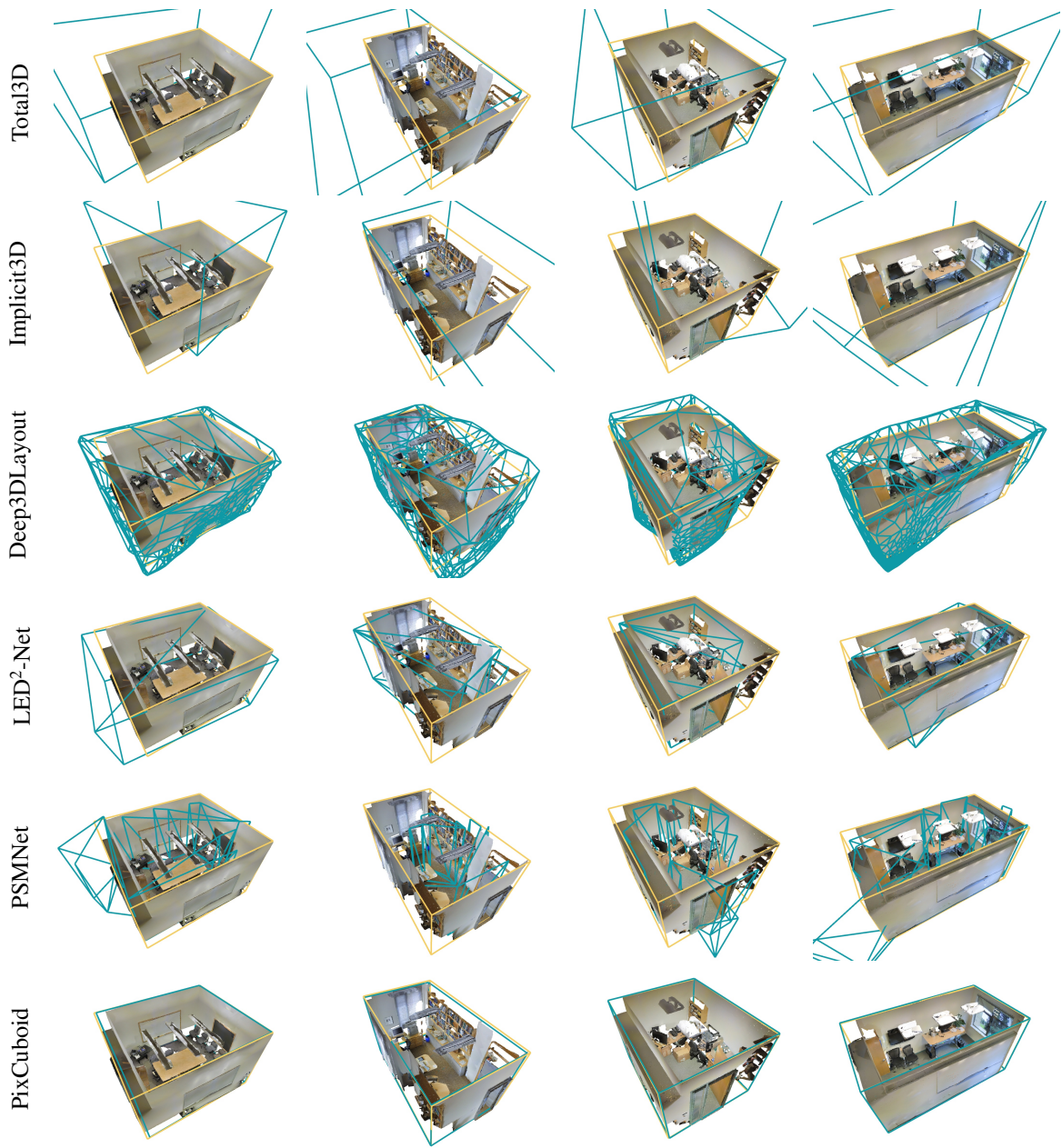

%% file: figs/failure_cases_extra.tex
\begin{figure*}
    \centering
    \begin{tikzpicture}[scale=1.0]
        \node at (0.0,0.0) {\includegraphics[scale=0.2]{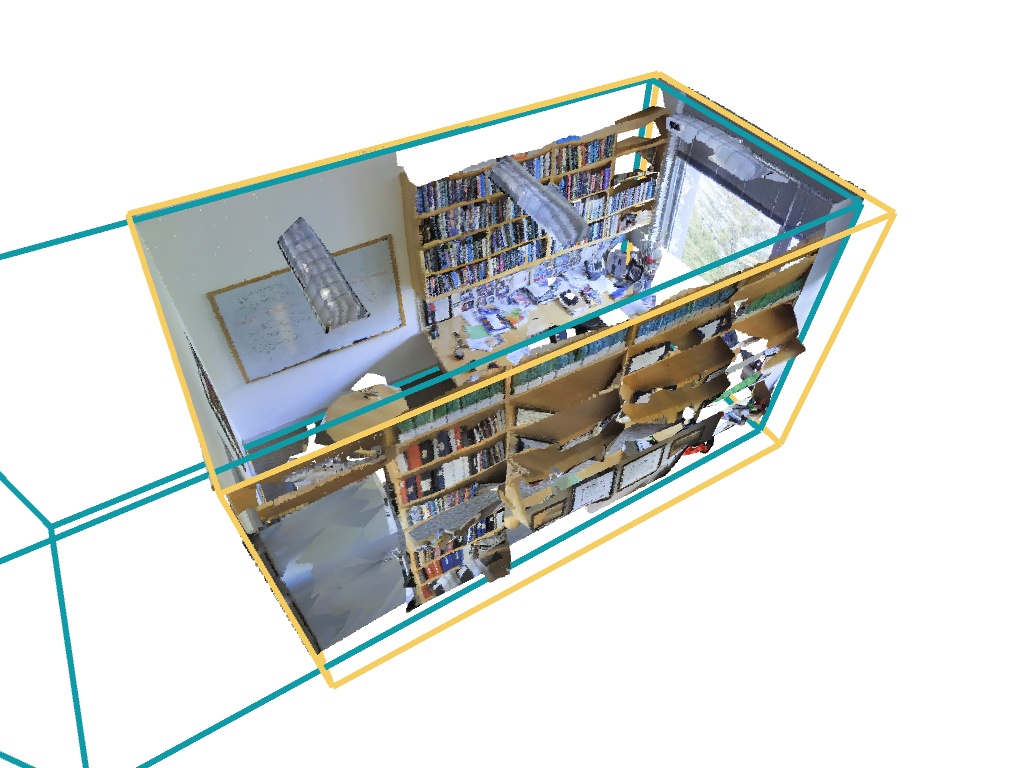}};
        \node at (0.0,5.7) {\includegraphics[scale=0.2]{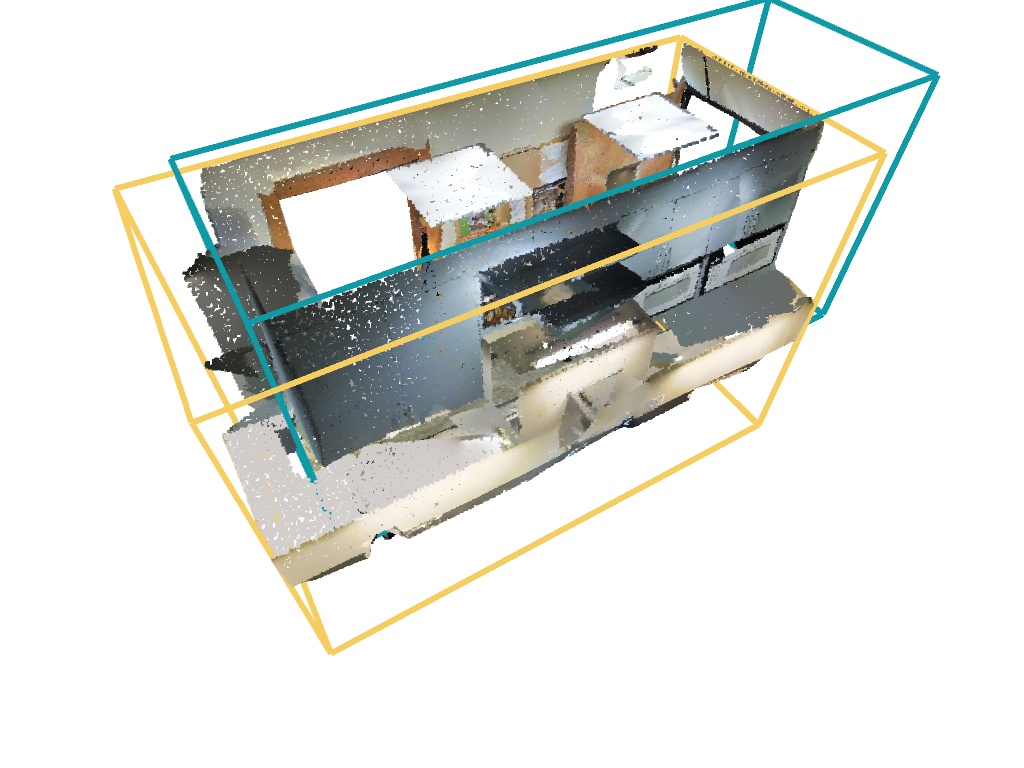}};
        \node at (7.5,0.0) {\includegraphics[scale=0.2]{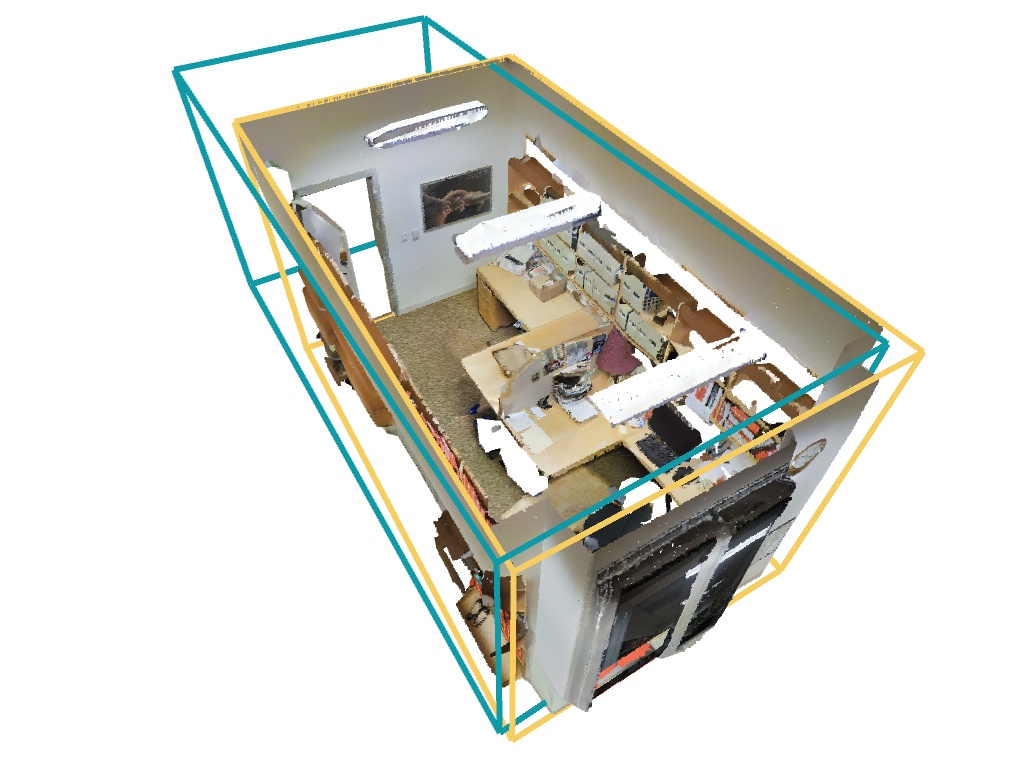}};
        \node at (7.5,5.7) {\includegraphics[scale=0.2]{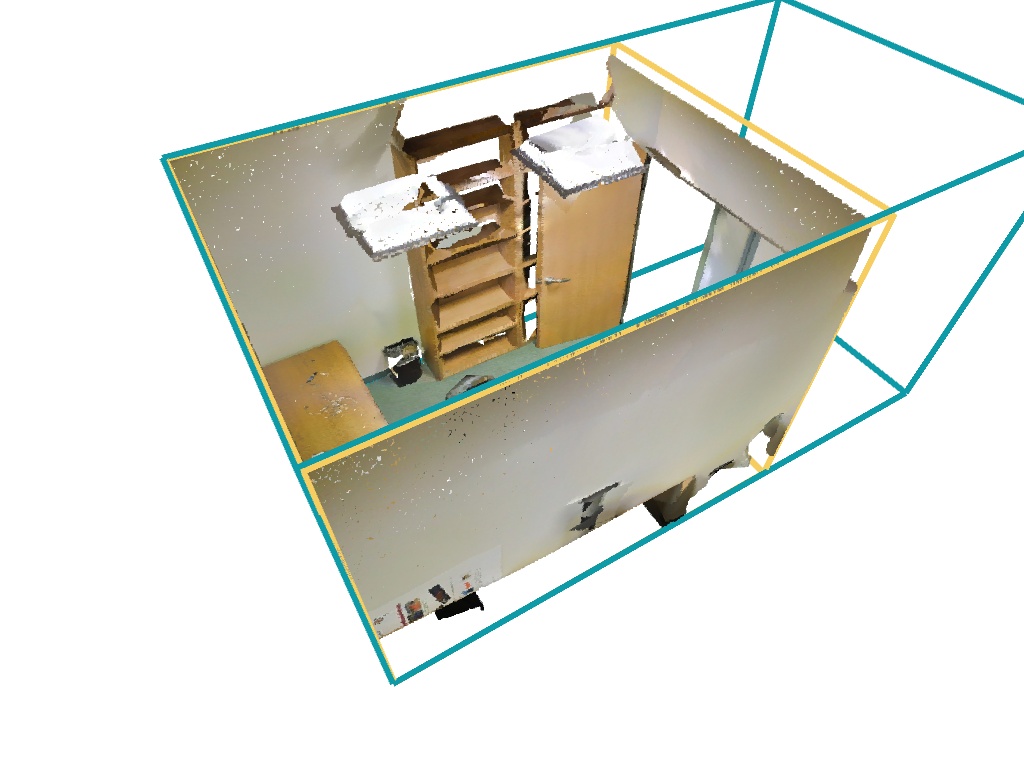}};
    \end{tikzpicture}
    \caption{Failure cases of \methodName{} on 2D-3D-Semantics. Predicted room layouts are shown in \textcolor{PredColor}{\bf blue} and ground truth cuboids in \textcolor{GTColor}{\bf yellow}.}
    \label{fig:failure-cases-extra}
\end{figure*}

%% file: figs/cuboid_init_extra.tex
\begin{figure*}
    \centering
    \begin{tikzpicture}[scale=1.0]
        \node at (0.0,0.0) {\includegraphics[scale=0.2]{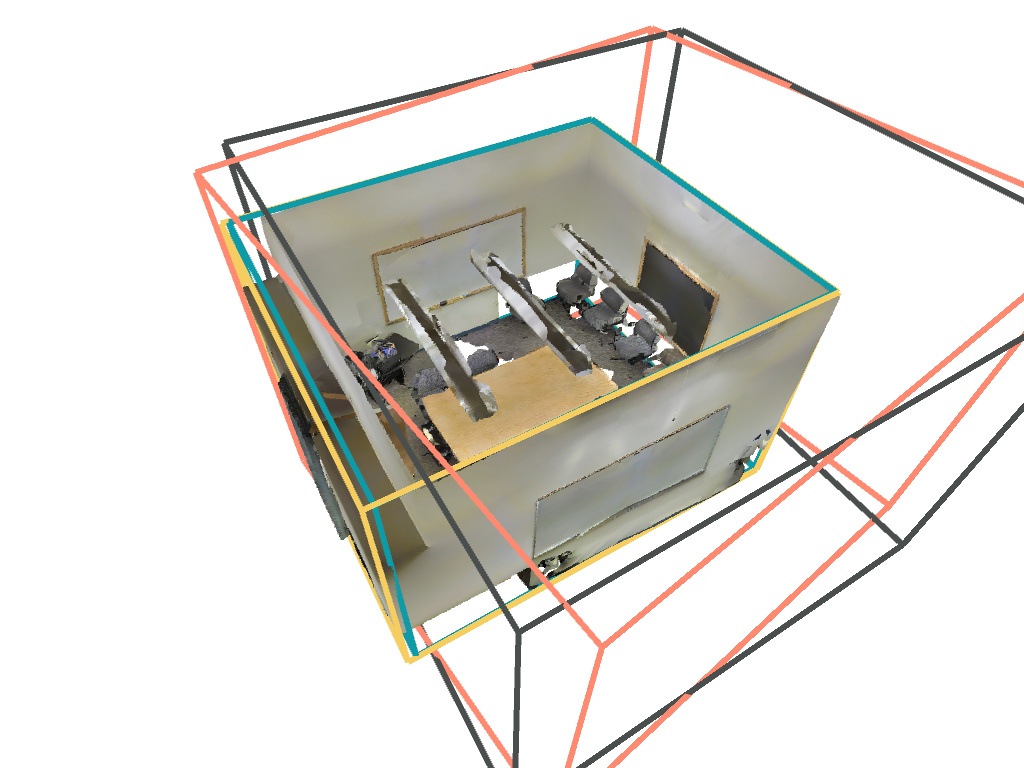}};
        \node at (0.0,5.7) {\includegraphics[scale=0.2]{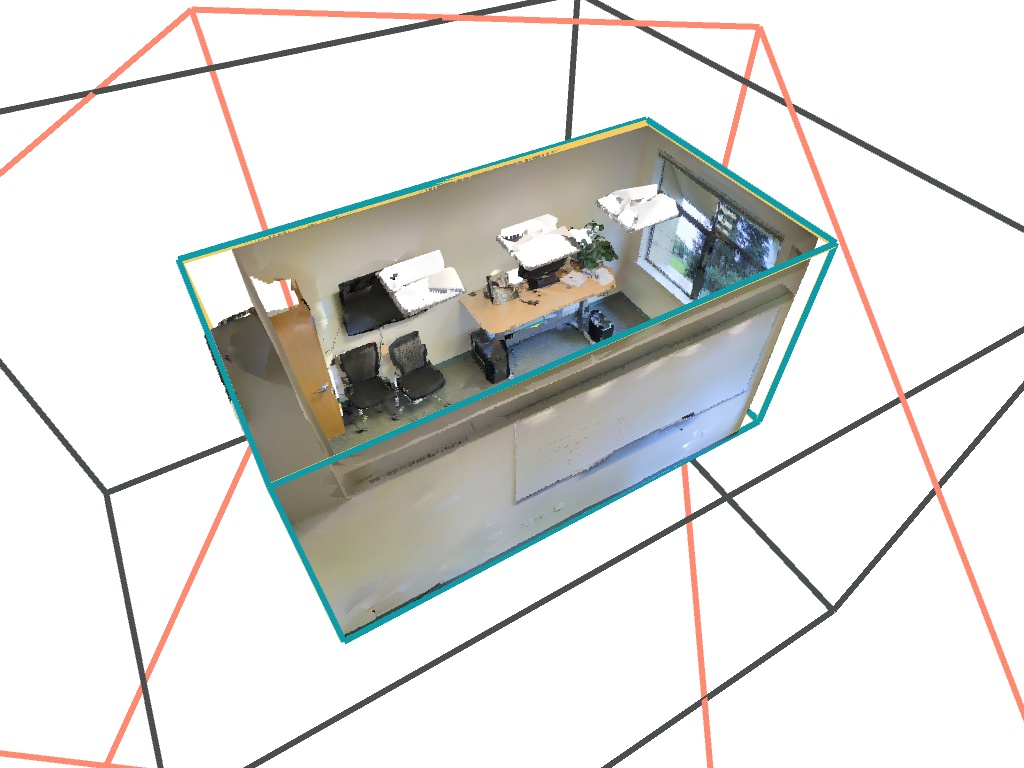}};
        \node at (7.5,0.0) {\includegraphics[scale=0.2]{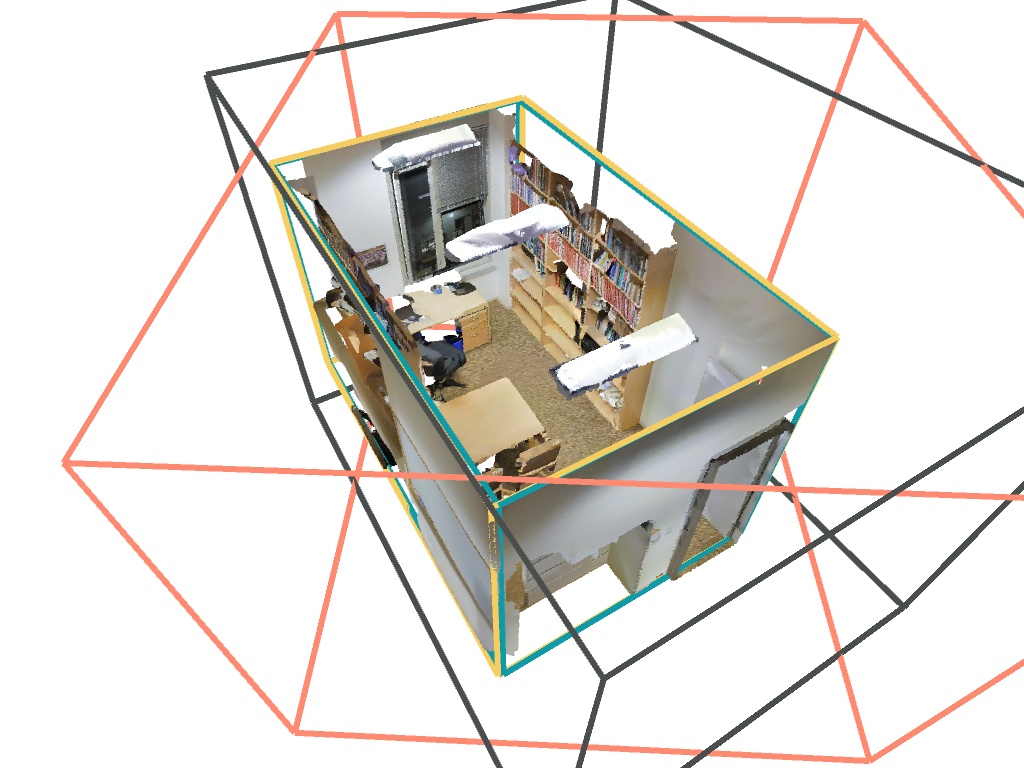}};
        \node at (7.5,5.7) {\includegraphics[scale=0.2]{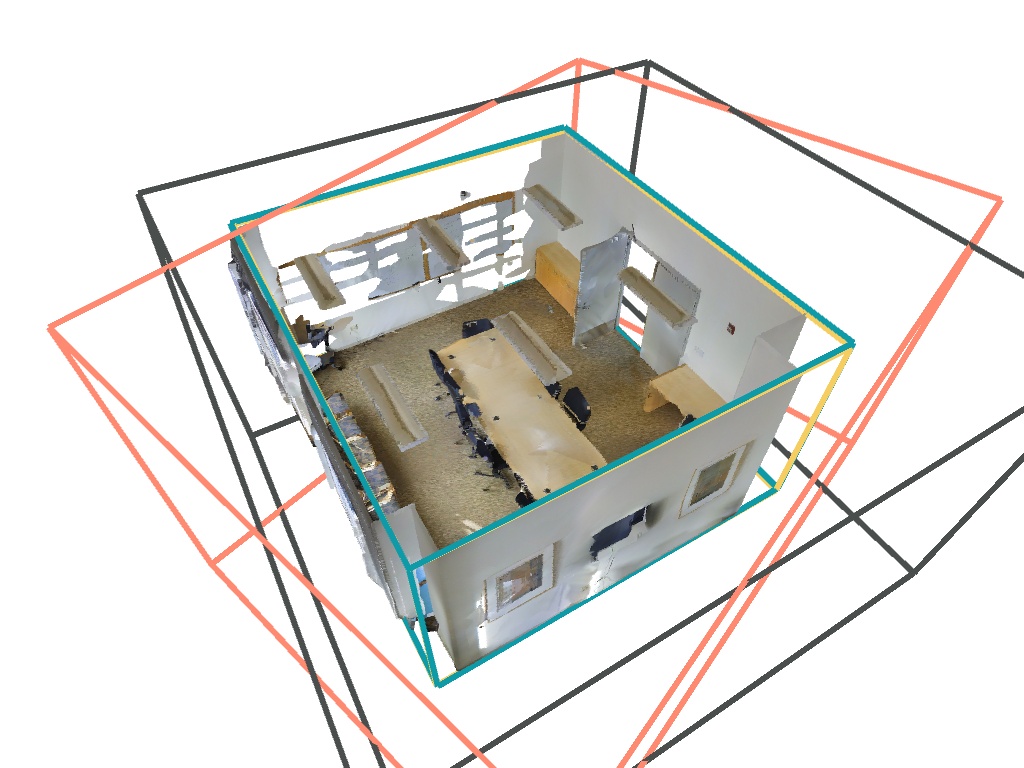}};
    \end{tikzpicture}
    \caption{Cuboid initialization examples on 2D-3D-Semantics. Even from poor initial estimates (\textcolor{InitColor}{\bf red}), \methodName{} can align the cuboids with the help of vanishing points (\textcolor{InitVPColor}{\bf gray}) and is able to converge to accurate layouts (\textcolor{PredColor}{\bf blue}). The ground truth is shown in \textcolor{GTColor}{\bf yellow}.}
    \label{fig:cuboid-init-extra}
\end{figure*}

%% file: tables/ablations.tex
\begin{table*}
\centering
\begin{tabular}{c l c c c c c c}
  \toprule
  && IoU $\uparrow$ & Chamfer $\downarrow$ & Rot. $\downarrow$ & Depth $\downarrow$ & Normal $\uparrow$ & Success $\uparrow$ \\
  \midrule
  \multirow{3}{*}{\rotatebox{90}{Feat.}}
  & RGB & 24.2 & 2.51 m & 36.4\degree & 1.02 m & 7.6 & 1.4\% \\
  & PixLoc (CMU) & 25.3 & 2.30 m & 33.4\degree & 1.02 m & 10.9 & 1.1\% \\
  & \methodName{} {\footnotesize ($E_{feat}$ only)} & \textbf{35.2} & \textbf{1.77 m} & \textbf{23.0\degree} & \textbf{0.75 m} & \textbf{30.8} & \textbf{5.8\%} \\
  \midrule
  \multirow{6}{*}{\rotatebox{90}{Cost}}
  & $E_{feat}$ & 35.2 & 1.77 m & 23.0\degree & 0.75 m & 30.8 & 5.8\% \\
  & $E_{edge}$ & 76.1 & 0.51 m & 4.7\degree & 0.23 m & 86.5 & 43.6\% \\
  & $E_{feat}$ + $E_{edge}$ & 81.9 & 0.39 m & 3.8\degree & 0.17 m & 88.9 & 61.4\% \\
  & $E_{feat}$ + $E_{VP}$ & 44.6 & 1.24 m & 1.5\degree & 0.57 m & 79.4 & 21.7\% \\
  & $E_{edge}$ + $E_{VP}$ & 83.1 & 0.29 m & \textbf{1.3\degree} & 0.13 m & 95.5 & 51.9\% \\
  & $E_{feat}$ + $E_{edge}$ + $E_{VP}$ & \textbf{87.2} & \textbf{0.22 m} & \textbf{1.3\degree} & \textbf{0.09 m} & \textbf{96.1} & \textbf{67.2\%} \\
  \midrule
  \multirow{3}{*}{\rotatebox{90}{Sampl.}}
  & Random & 86.9 & 0.23 m & 1.4\degree & 0.10 m & 95.8 & 65.8\% \\
  & Floor/wall/ceiling & 86.6 & 0.25 m & 1.4\degree & 0.10 m & 95.7 & 66.4\% \\
  & Guided & \textbf{87.2} & \textbf{0.22 m} & \textbf{1.3\degree} & \textbf{0.09 m} & \textbf{96.1} & \textbf{67.2\%} \\
  \midrule
  \multirow{3}{*}{\rotatebox{90}{Res.}}
  & Low (256 px) & 84.2 & 0.28 m & \textbf{1.3\degree} & 0.12 m & 95.6 & 63.3\% \\
  & Medium (512 px) & \textbf{87.2} & \textbf{0.22 m} & \textbf{1.3\degree} & \textbf{0.09 m} & \textbf{96.1} & \textbf{67.2\%} \\
  & High (768 px) & 85.7 & 0.26 m & \textbf{1.3\degree} & 0.12 m & 95.6 & 66.1\% \\
  \midrule
  \multirow{4}{*}{\rotatebox{90}{Init.}}
  & Random & 60.8 & 1.28 m & 18.0\degree & 0.52 m & 60.4 & 40.0\% \\
  & Random + VP & 72.2 & 0.79 m & 10.9\degree & 0.32 m & 74.6 & 51.9\% \\
  & Y down & 84.9 & 0.27 m & 2.1\degree & 0.12 m & 93.6 & 64.7\% \\
  & Y down + VP & \textbf{87.2} & \textbf{0.22 m} & \textbf{1.3\degree} & \textbf{0.09 m} & \textbf{96.1} & \textbf{67.2\%} \\
  \midrule
  \multirow{3}{*}{\rotatebox{90}{Scales}}
  & Coarse & 81.0 & 0.32 m & 1.4\degree & 0.15 m & 94.6 & 36.7\% \\
  & Coarse + medium & 86.5 & 0.23 m & 1.4\degree & 0.10 m & 95.8 & 65.6\% \\
  & Coarse + medium + fine & \textbf{87.2} & \textbf{0.22 m} & \textbf{1.3\degree} & \textbf{0.09 m} & \textbf{96.1} & \textbf{67.2\%} \\
  \bottomrule
\end{tabular}
\caption{Ablation experiments on our ScanNet++ v2 test set.}
\label{tab:ablation-study-full}
\end{table*}

%% file: figs/example_feat_edge_maps.tex
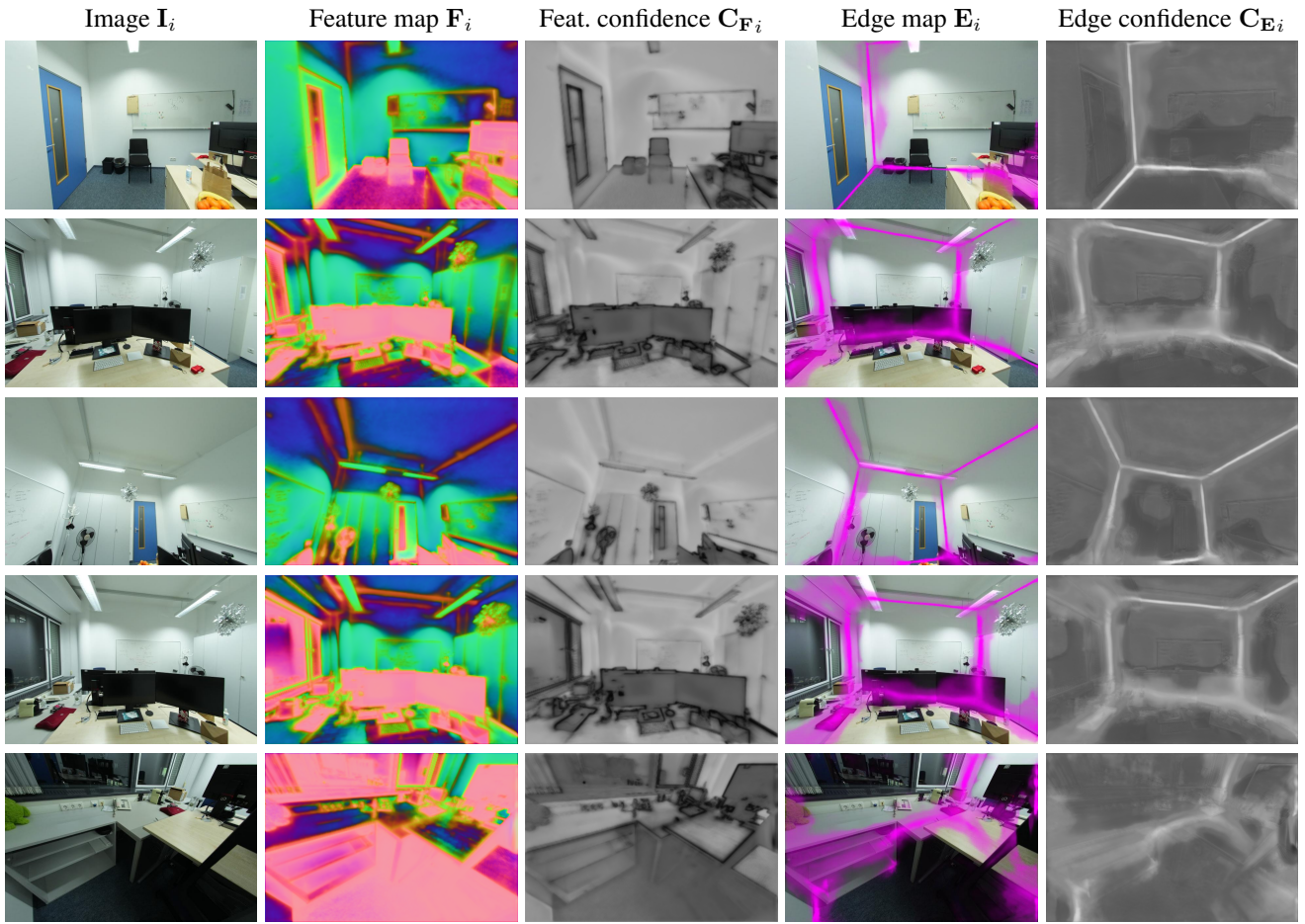
\begin{figure*}[tp]
    \centering
    \begin{tikzpicture}[scale=1.0]
        \node at (0.0,9.6) {\includegraphics[scale=0.125]{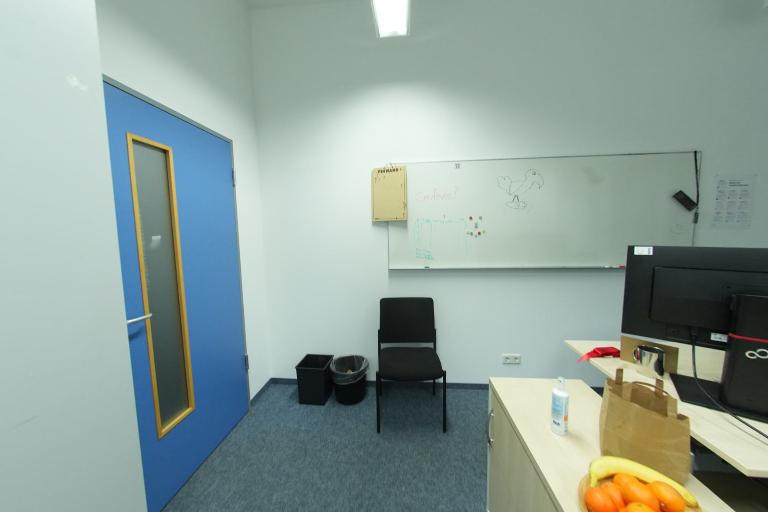}};
        \node at (0.0,7.2) {\includegraphics[scale=0.125]{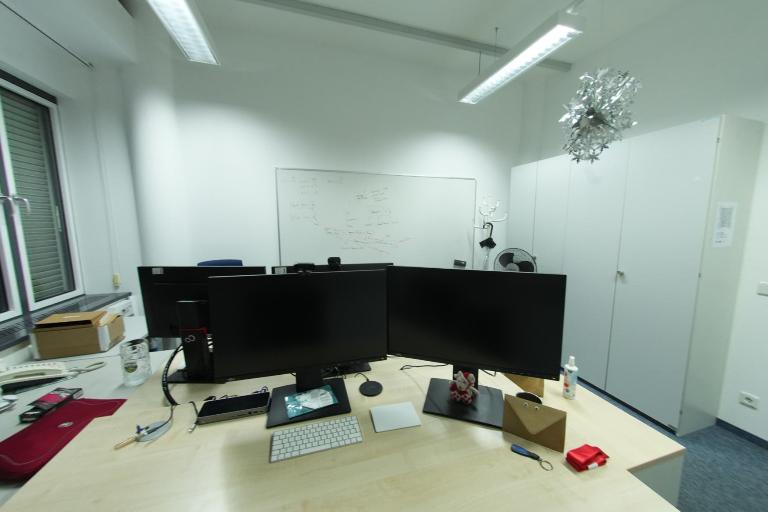}};
        \node at (0.0,4.8) {\includegraphics[scale=0.125]{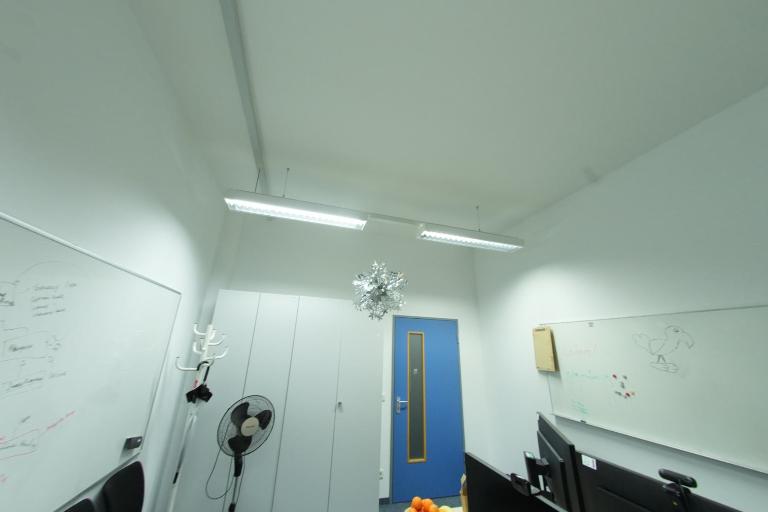}};
        \node at (0.0,2.4) {\includegraphics[scale=0.125]{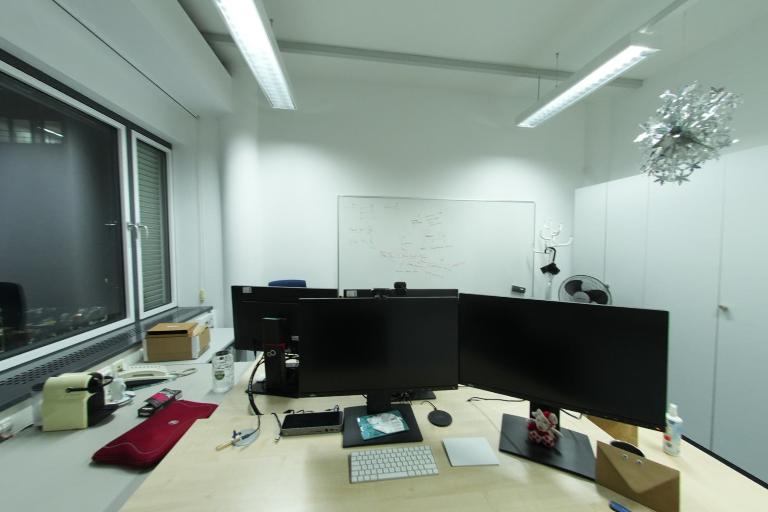}};
        \node at (0.0,0.0) {\includegraphics[scale=0.125]{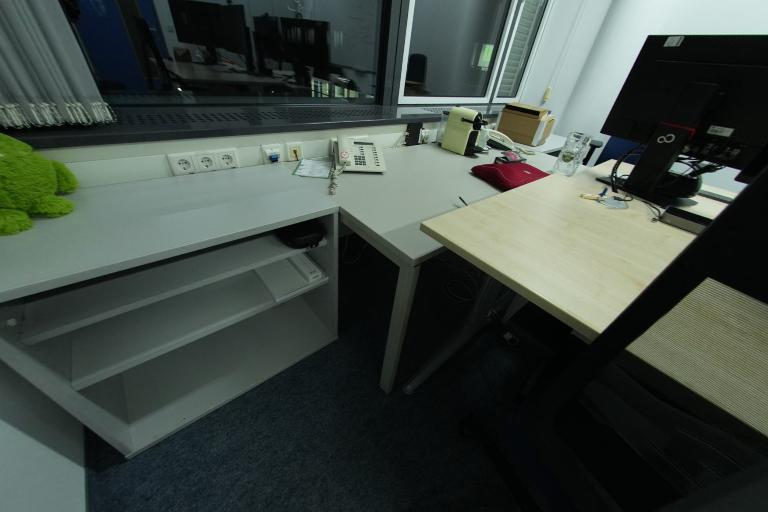}};

        \node at (3.5,9.6) {\includegraphics[scale=0.125]{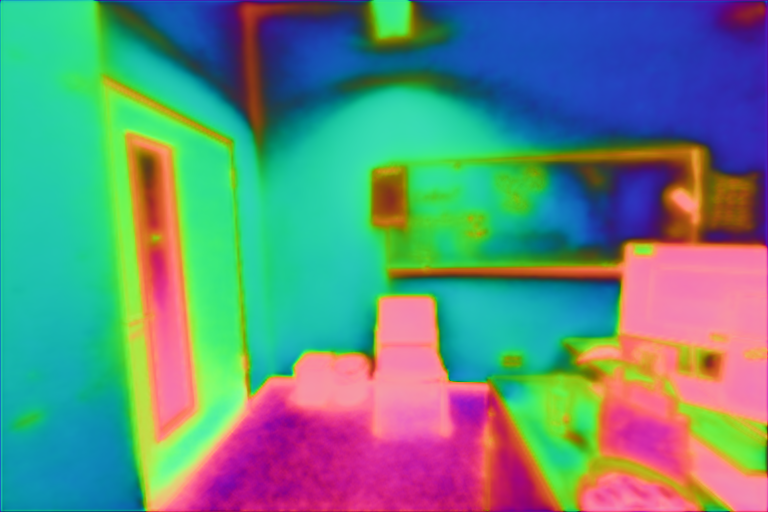}};
        \node at (3.5,7.2) {\includegraphics[scale=0.125]{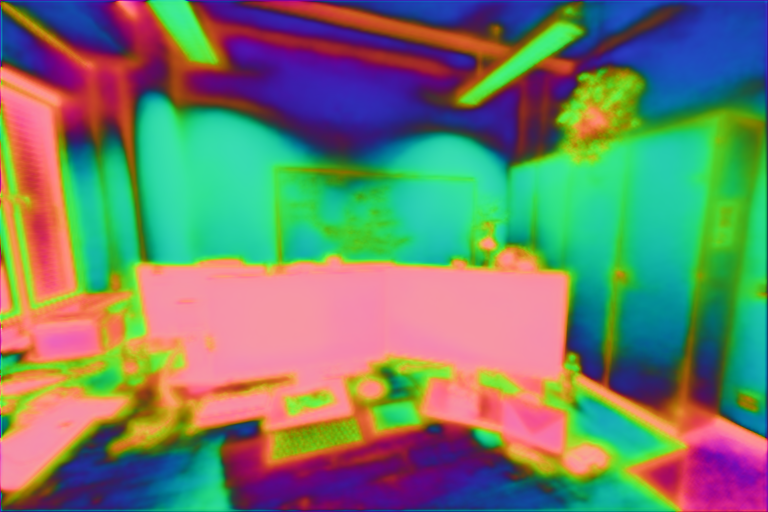}};
        \node at (3.5,4.8) {\includegraphics[scale=0.125]{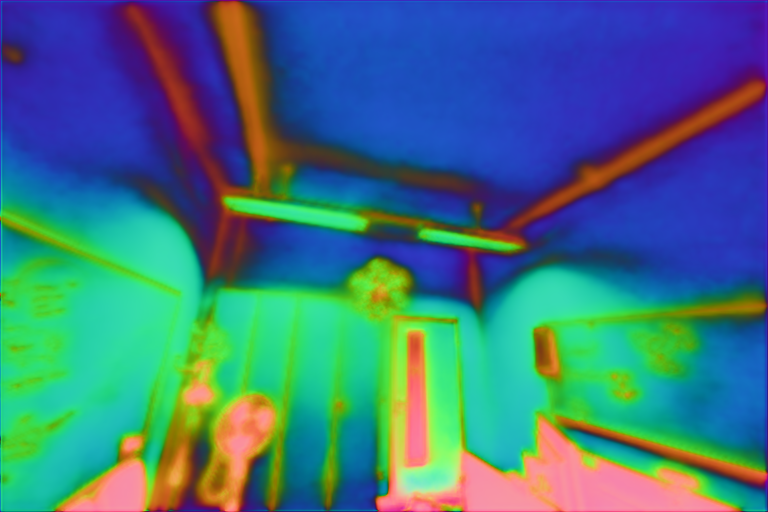}};
        \node at (3.5,2.4) {\includegraphics[scale=0.125]{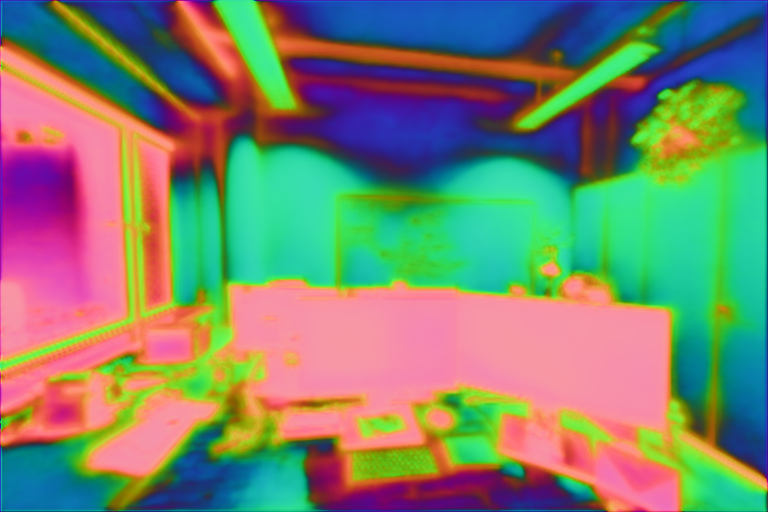}};
        \node at (3.5,0.0) {\includegraphics[scale=0.125]{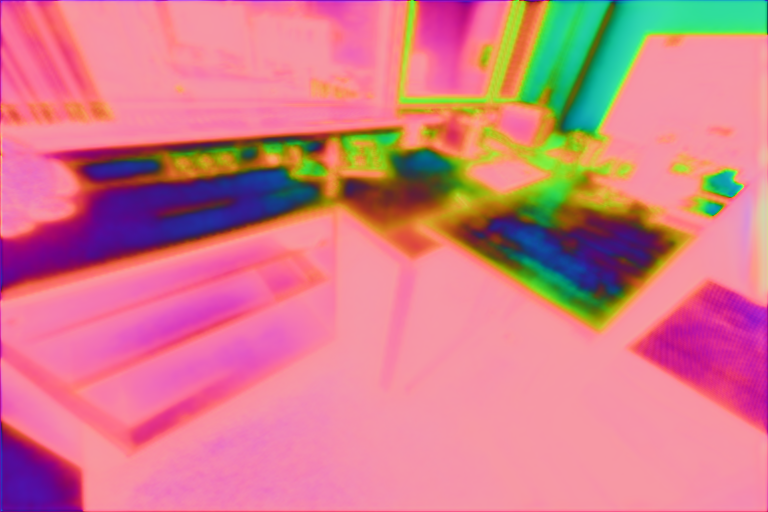}};

        \node at (7.0,9.6) {\includegraphics[scale=0.125]{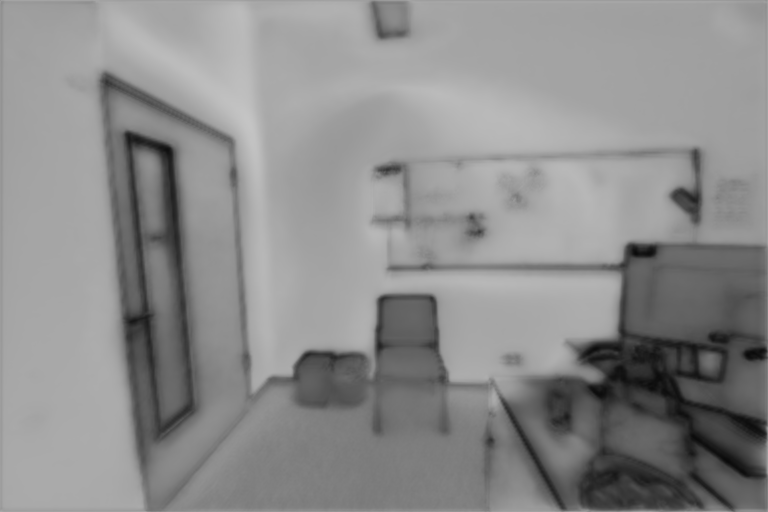}};
        \node at (7.0,7.2) {\includegraphics[scale=0.125]{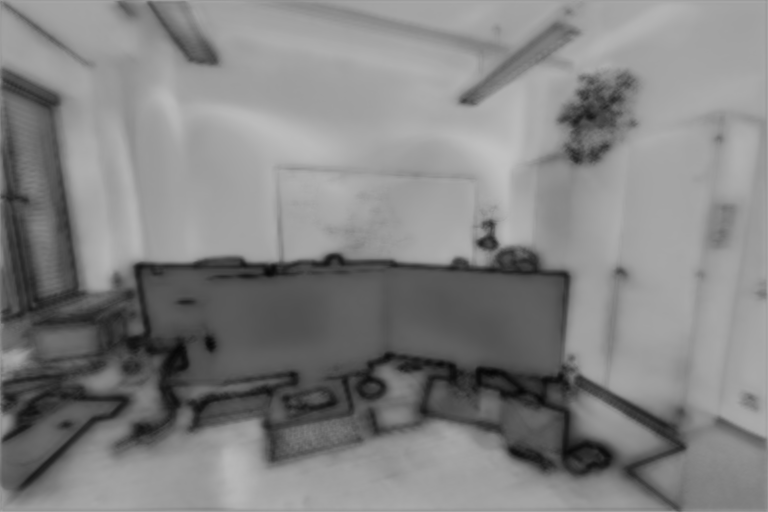}};
        \node at (7.0,4.8) {\includegraphics[scale=0.125]{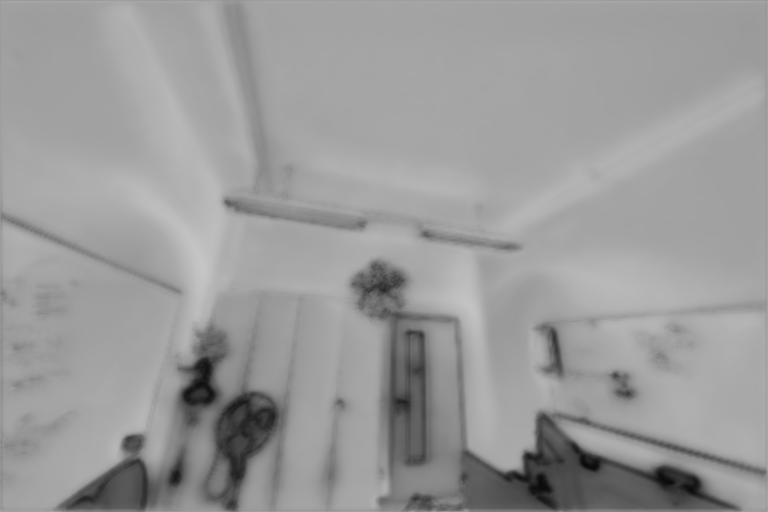}};
        \node at (7.0,2.4) {\includegraphics[scale=0.125]{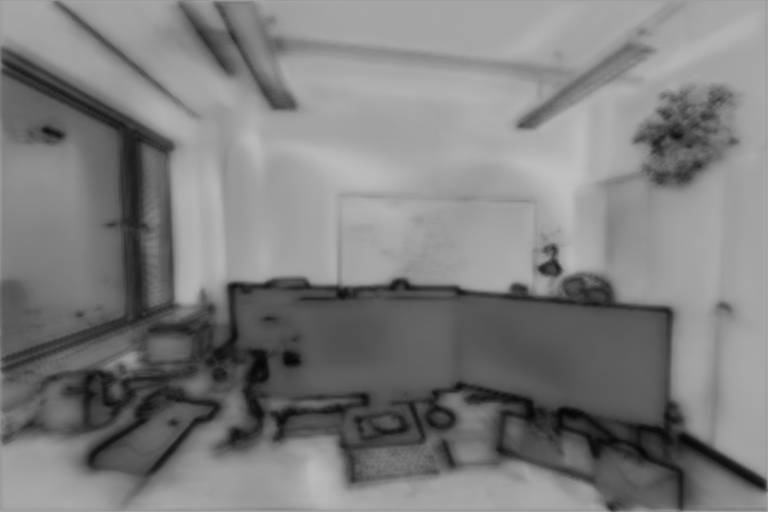}};
        \node at (7.0,0.0) {\includegraphics[scale=0.125]{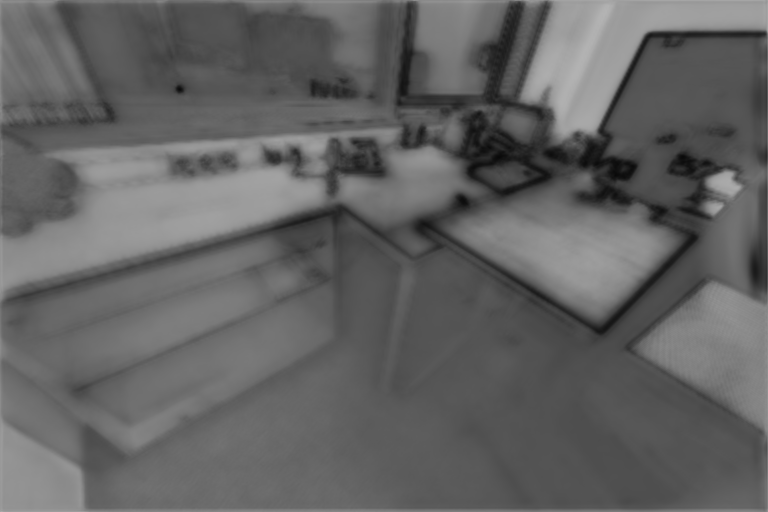}};

        \node at (10.5,9.6) {\includegraphics[scale=0.125]{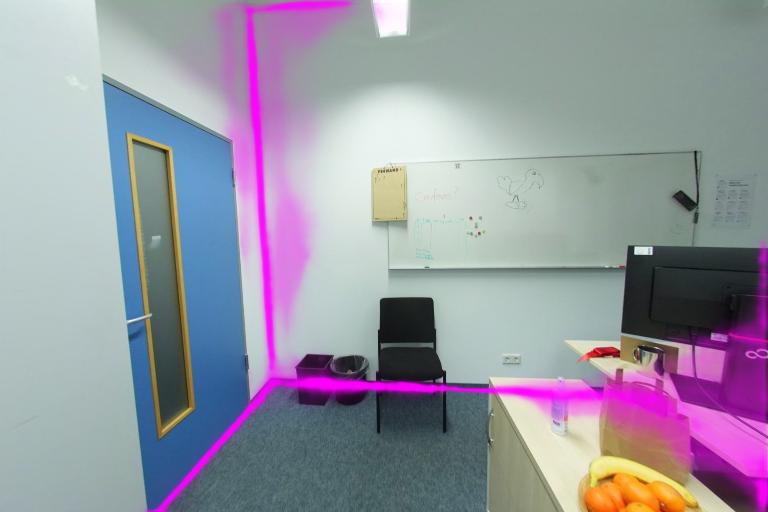}};
        \node at (10.5,7.2) {\includegraphics[scale=0.125]{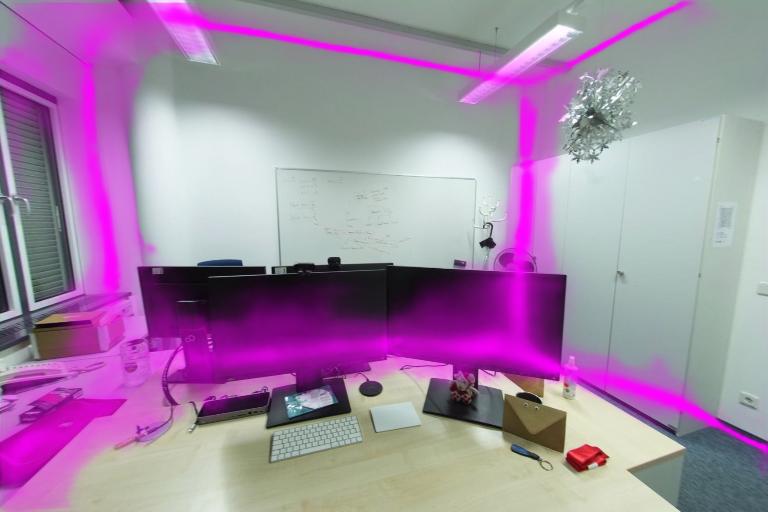}};
        \node at (10.5,4.8) {\includegraphics[scale=0.125]{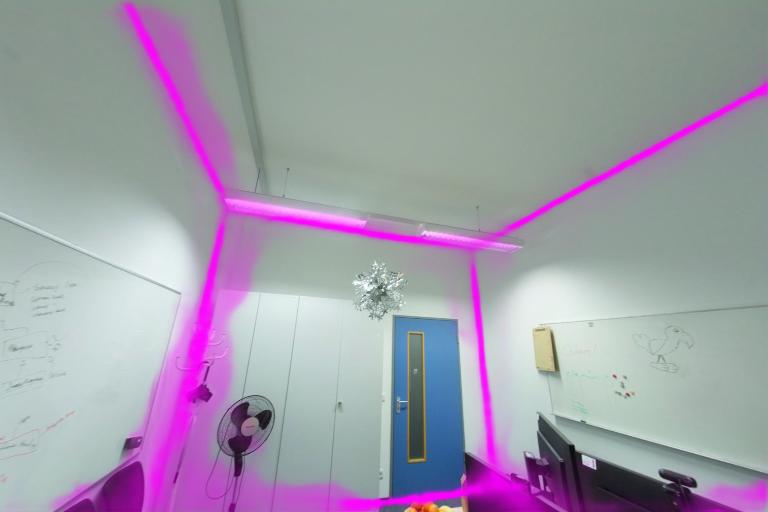}};
        \node at (10.5,2.4) {\includegraphics[scale=0.125]{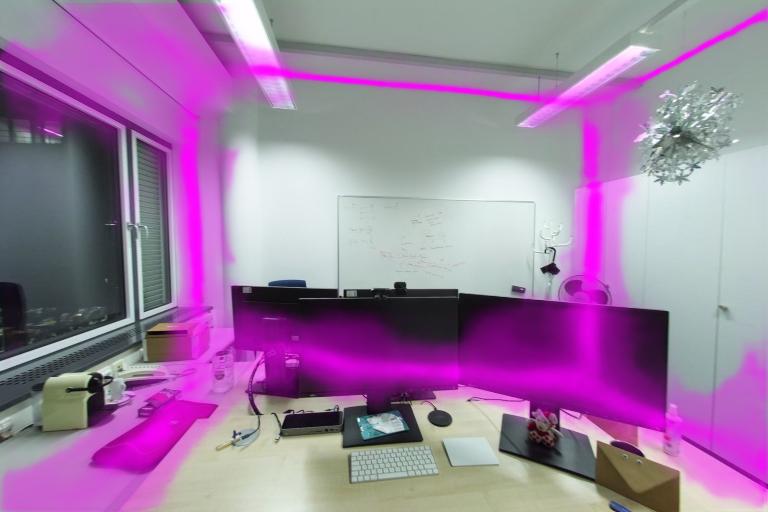}};
        \node at (10.5,0.0) {\includegraphics[scale=0.125]{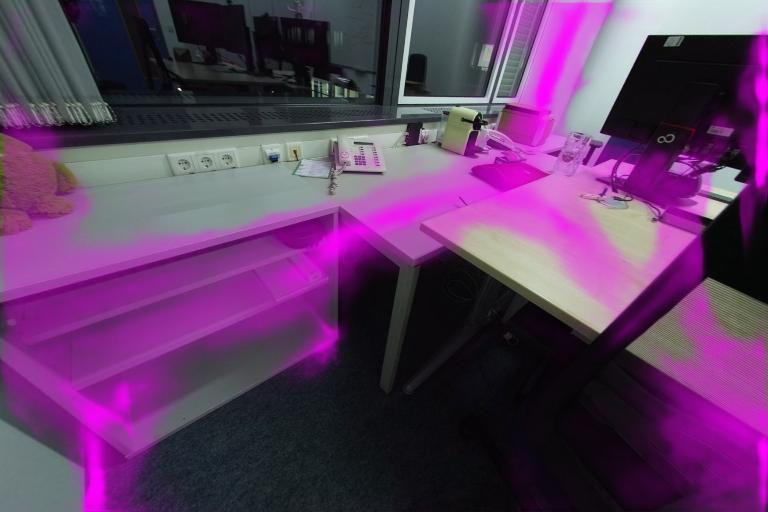}};

        \node at (14.0,9.6) {\includegraphics[scale=0.125]{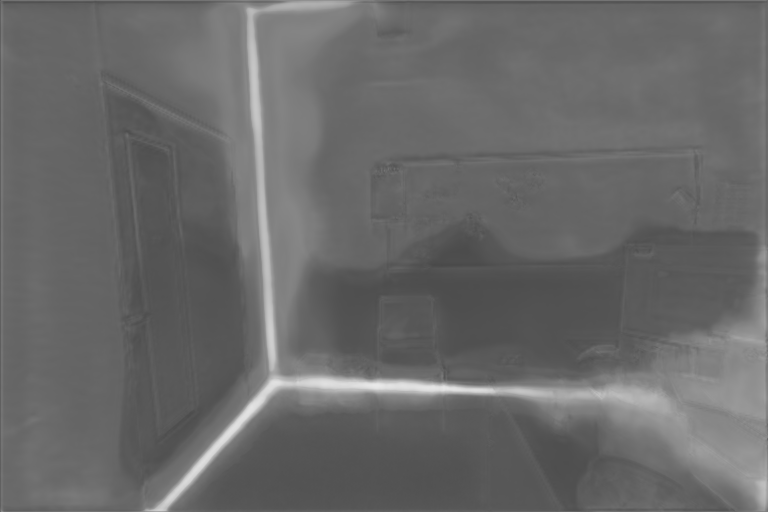}};
        \node at (14.0,7.2) {\includegraphics[scale=0.125]{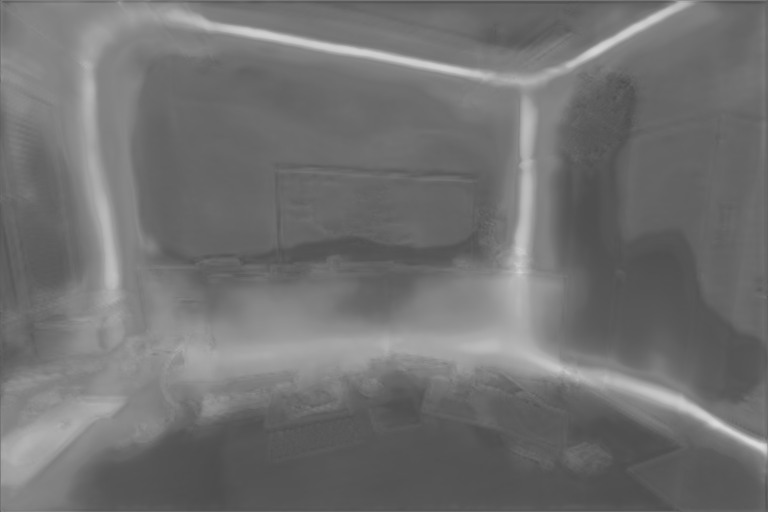}};
        \node at (14.0,4.8) {\includegraphics[scale=0.125]{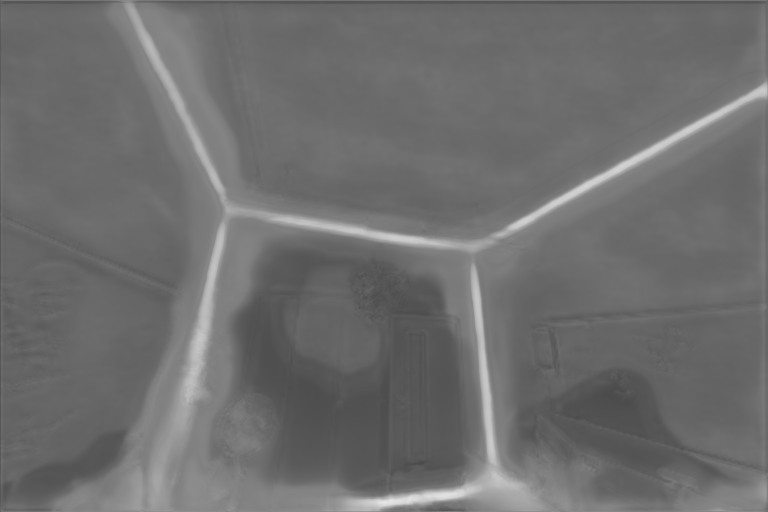}};
        \node at (14.0,2.4) {\includegraphics[scale=0.125]{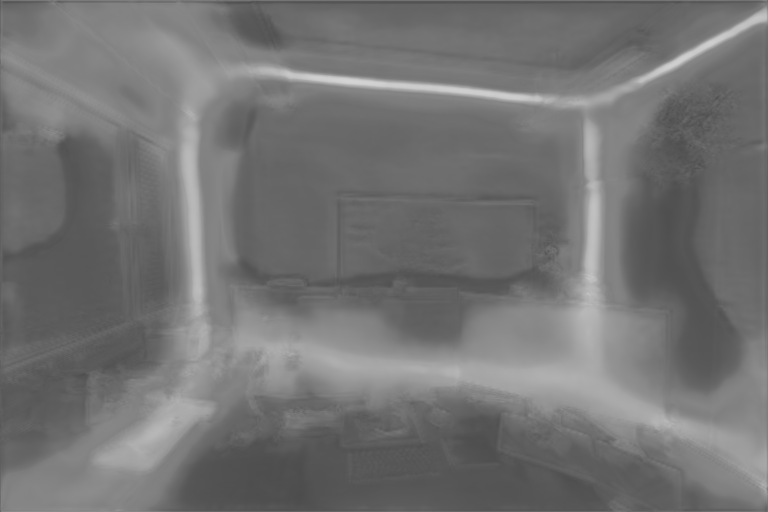}};
        \node at (14.0,0.0) {\includegraphics[scale=0.125]{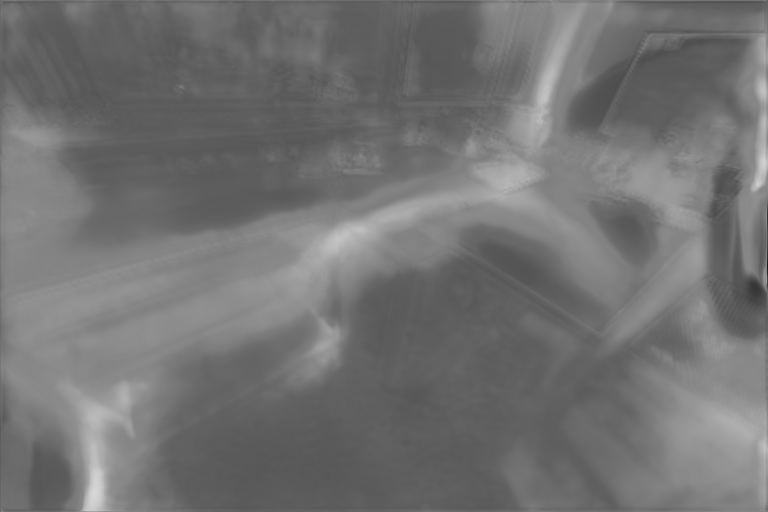}};

        \node[align=center] at (0.0,11.0) {Image $\II_i$};
        \node[align=center] at (3.5,11.0) {Feature map $\FF_i$};
        \node[align=center] at (7.0,11.0) {Feat. confidence $\CF_i$};
        \node[align=center] at (10.5,11.0) {Edge map $\EE_i$};
        \node[align=center] at (14.0,11.0) {Edge confidence $\CE_i$};
    \end{tikzpicture}
    \caption{Example feature- and edge maps with corresponding confidence maps for one image tuple in our ScanNet++ v2 test set. Results are shown only on the finest level. Feature maps have been mapped to RGB using PCA. Confidence ranges from low (black) to high (white). }
    \label{fig:example-maps}
\end{figure*}